\documentclass[manuscript]{acmart}
\usepackage{multirow}
\usepackage{bbding}
\usepackage{tabularx}
\usepackage{booktabs}
\usepackage{array}
\usepackage{threeparttable}
\usepackage{graphicx}
\usepackage{utfsym}

\usepackage{appendix}
\usepackage{amsthm,amsmath,amssymb}
\usepackage{mathrsfs}
\AtBeginDocument{%
  \providecommand\BibTeX{{%
    \normalfont B\kern-0.5em{\scshape i\kern-0.25em b}\kern-0.8em\TeX}}}

\setcopyright{acmlicensed}
\copyrightyear{2018}
\acmYear{2018}
\acmDOI{XXXXXXX.XXXXXXX}

\begin{document}

\title{Natural Language Understanding and Inference with MLLM in Visual Question Answering: A Survey}

\author{Jiayi Kuang}
\authornote{Both authors contributed equally to this research.}
\email{kuangjy6@mail2.sysu.edu.cn}
\author{Jingyou Xie}
\authornotemark[1]
\email{xiejy73@mail2.sysu.edu.cn}
\affiliation{%
  \institution{Sun Yat-sen University}
  \city{Shenzhen}
  \country{China}
}

\author{Haohao Luo}
\authornotemark[2]
\email{luohh5@mail2.sysu.edu.cn}
\affiliation{%
  \institution{Sun Yat-sen University}
  \city{Shenzhen}
  \country{China}
}

\author{Ronghao Li}
\authornotemark[2]
\email{lirh56@mail2.sysu.edu.cn}
\affiliation{%
  \institution{Sun Yat-sen University}
  \city{Shenzhen}
  \country{China}
}

\author{Zhe Xu}
\authornote{Equal contribution.}
\email{xuzh226@mail2.sysu.edu.cn}
\affiliation{%
  \institution{Sun Yat-sen University}
  \city{Shenzhen}
  \country{China}
}

\author{Xianfeng Cheng}
\email{chengxf6@mail2.sysu.edu.cn}
\affiliation{%
  \institution{Sun Yat-sen University}
  \city{Shenzhen}
  \country{China}
}

  \author{Yinghui Li}
  \email{liyinghu20@mails.tsinghua.edu.cn}
  \affiliation{%
      \institution{Tsinghua University}
      \city{Shenzhen}
      \country{China}
  }

  \author{Xika Lin}
  \email{xikalin@gmail.com}
  \affiliation{%
      \institution{Department of Computer Science, Worcester Polytechnic Institute}
      \city{Worcester, MA}
      \country{USA}
  }


\author{Ying Shen}
\authornote{Corresponding author.}
\email{sheny76@mail.sysu.edu.cn}
\affiliation{%
  \institution{Sun Yat-Sen University}
  \city{Shenzhen}
  \country{China}
}
\renewcommand{\shortauthors}{Kuang and Xie, et al.}

\begin{abstract}
Visual Question Answering (VQA) is a challenge task that combines natural language processing and computer vision techniques and gradually becomes a benchmark test task in multimodal large language models (MLLMs).
The goal of our survey is to provide an overview of the development of VQA and a detailed description of the latest models with high timeliness.
This survey gives an up-to-date synthesis of natural language understanding of images and text, as well as the knowledge reasoning module based on image-question information on the core VQA tasks. In addition, we elaborate on recent advances in extracting and fusing modal information with vision-language pretraining models and multimodal large language models in VQA. We also exhaustively review the progress of knowledge reasoning in VQA by detailing the extraction of internal knowledge and the introduction of external knowledge. Finally, we present the datasets of VQA and different evaluation metrics and discuss possible directions for future work.
\end{abstract}



\received{13 January 2023}
\received[revised]{24 June 2024}
\received[revised]{18 October 2024}

\maketitle

\section{INTRODUCTION}
\subsection{What is Visual Question Answering?}
Visual Question Answering (VQA) has emerged as a significant task in the intersection of computer vision and natural language processing (NLP). The goal of VQA is to predict an answer $A$ to a question $Q$ based on visual information $V$, formalized as $A = f(Q,V)$, where $f$ represents the model function \cite{antol2015vqa}. Visual inputs may include images of landscapes, people, or videos, and questions can range from multiple-choice to open-ended formats. 

\begin{figure}[t]
    \centering
    \includegraphics[width=0.85\linewidth]{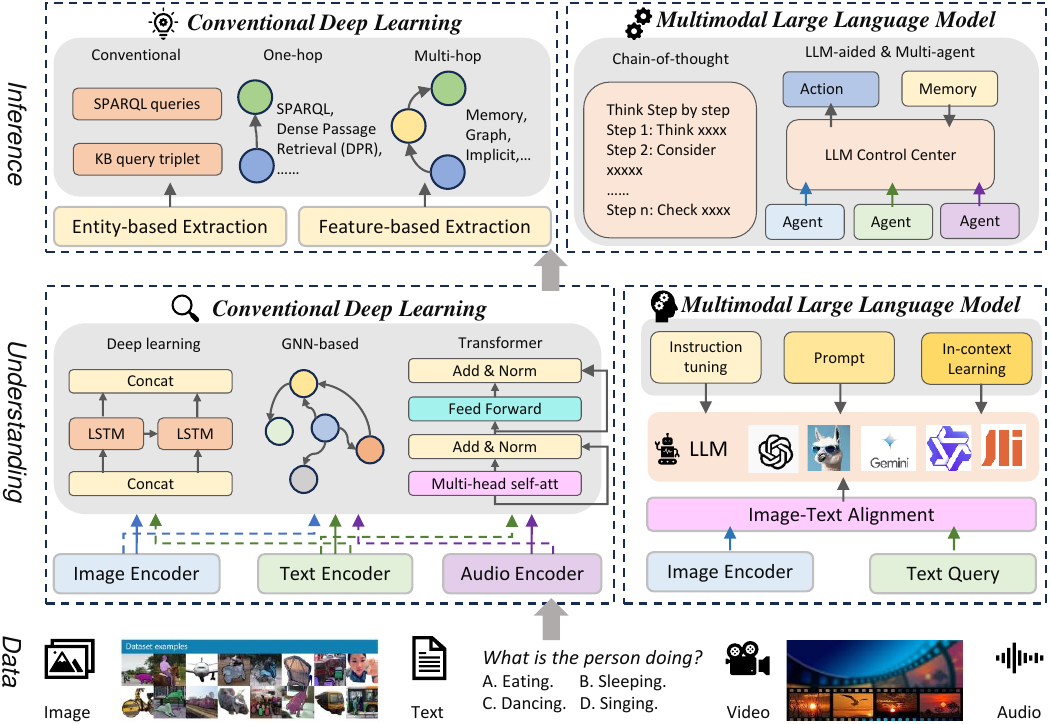}
    \caption{The data used in VQA tasks, with the conclusion of the understanding and inference methods from conventional models to multimodal large language models.}
    \label{fig:intro}
\end{figure}

A VQA model typically consists of the following steps. First, features are extracted from visual and textual information respectively. Then, intra-modal and inter-modal representations are learned by aligning and fusing the features. Finally, the obtained image-question knowledge representation is predicted to complete the question answering task. We summarize these natural language and image understanding and inference in Figure \ref{fig:intro}.

The VQA task was first introduced by Agrawal et al. \cite{antol2015vqa} alongside the classic VQA v1 dataset, which contains 614,163 manually curated question-answer pairs based on MS COCO images.
The earliest VQA tasks focus on the direct understanding of visual and textual information, mainly using deep learning methods, in which VGG-Net \cite{graves2012long}, Faster-RCNN \cite{ren2015faster} are used for visual feature extraction, and LSTM \cite{antol2015vqa}, GRU are used for textual feature extraction.
The introduction of attention mechanisms marked a pivotal advancement in the field. Stacked attention \cite{yang2016stacked} and co-attention \cite{lu2016hierarchical} frameworks significantly improved the fusion of visual and textual features by learning complementary information between modalities. Additionally, Graph Neural Networks (GNNs) have been increasingly utilized to capture the complex relationships between visual and textual elements through multimodal heterogeneous graphs \cite{ding2022mukea}.

In recent years, the Transformer architecture \cite{vaswani2017attention} has further revolutionized VQA. Visual-language pre-training models such as BLIP \cite{li2022blip} leverage self-attention and cross-attention mechanisms to enhance multimodal fusion, enabling significant progress in zero-shot VQA. Emerging multimodal large language models (MLLMs), including Flamingo \cite{alayrac2022flamingo}, BLIP-2 \cite{li2023blip}, InternVL \cite{chen2024internvl} have demonstrated exceptional performance, particularly in open-ended and zero-shot question answering scenarios.

While early VQA models focused on direct feature extraction and alignment, recent advances have shifted toward knowledge-based reasoning, which goes beyond perception to deeper cognitive understanding. Some approaches explore how to transform fragmented internal knowledge into structured knowledge. Knowledge can be represented as the form of knowledge triplets and applied to one-hop or multi-hop reasoning \cite{ding2022mukea}. Other approaches form a joint image-question-knowledge representation by introducing an external knowledge base \cite{heo2022hypergraph} or performing passage retrieval \cite{gao2022transform}. Then this joint representation is utilized for reasoning. Compared with the previous work, the understanding of the knowledge representation coming from images and texts at this stage is more in-depth. In addition, Multimodal large language models now integrate sophisticated reasoning methods such as instruction tuning and Chain-of-Thought prompting to improve answer accuracy. In this respect, VQA is a comprehensive task that bridges computer vision and natural language processing (NLP). On the one hand, computer vision aims to teach machines how to see, working on ways to acquire, process, and understand images. NLP, on the other hand, is a field concerned with enabling interactions between computers and humans in natural language, which not only aims to teach machines how to read but also pays attention to the thought process of question answering. It is worth noting that natural language generation methods play an important role in VQA, since it has a non-negligible role in the answer generation process, especially contributing to assisting the model to achieve better results in open-ended question answering.

VQA's impact extends across diverse applications, from aiding visually impaired users in navigating digital and real-world environments to improving image retrieval, autonomous driving, medical diagnostics, and conversational AI systems \cite{10.1016/j.patrec.2021.09.008,10104870}. The task can also evolve into visual dialogue, where systems must answer questions based on an entire conversation history, further highlighting its potential in real-world problem-solving.

\subsection{Comparison with Other Survey Works}
Over the years, Visual Question Answering (VQA) has garnered significant attention in the research community, leading to the publication of numerous high-quality surveys. These surveys provide valuable insights for both beginners and experienced researchers by outlining challenges, recent trends, and open problems. Table \ref{tab:survey} highlights some of the prominent generalized surveys in the field.

\begin{table}[t]
\caption{Some of the prominent generalized long surveys along with the published year, topics, challenges and key contributions.}
\scalebox{0.7}{
\begin{tabular}{l|l|l|ll}
\midrule[1.3pt]
Surveys                                     & Year & Topic                                                                                                & Challenges                                                                                                                                                                        & Contributions                                                                                                                                                                                                                                 \\ \midrule
Wu et al. \cite{wu2017visual}                                   & 2017 & \begin{tabular}[c]{@{}l@{}}Visual Question Answering,\\ Knowledge Base\end{tabular}                  & \begin{tabular}[c]{@{}l@{}}Question form is unknown,\\      Visual and textual comprehension,\\      Lack of knowledge\end{tabular}                                               & \begin{tabular}[c]{@{}l@{}}First comprehensive overview of   the field,\\      Definition and classification of the task,\\      In-depth analysis of the question/answer pairs\end{tabular}                                                  \\ \midrule
Kafle et al. \cite{kafle2017visual}                               & 2017 & \begin{tabular}[c]{@{}l@{}}Image Understanding,\\ VQA datasets\end{tabular}                          & \begin{tabular}[c]{@{}l@{}}Solving a wide range of CV   tasks,\\      Dataset bias,\\      VQA algorithm\end{tabular}                                                             & \begin{tabular}[c]{@{}l@{}}Compare VQA with other computer   vision tasks,\\      Exploring whether current VQA benchmarks \\ are suitable for evaluating\end{tabular}                                                                           \\ \midrule
Zhang et al. \cite{zhang2019information}                               & 2019 & \begin{tabular}[c]{@{}l@{}}ImageQA and VideoQA,\\      Information Fusion\end{tabular}               & \begin{tabular}[c]{@{}l@{}}Fusion of semantic information of \\ text and vision, \\     temporal relationship in VideoQA\end{tabular} & \begin{tabular}[c]{@{}l@{}}Abstract fusion framework that   can fit the \\ majority of existing VQA models,\\       Two-channel fusion and multi-channel   fusion,\\      Clear organization on the proposed fusion techniques\end{tabular} \\ \midrule
Srivastava et al. \cite{srivastava2021visual}                           & 2021 & \begin{tabular}[c]{@{}l@{}}Deep learning in VQA,\\      Robust datasets\end{tabular}                 & \begin{tabular}[c]{@{}l@{}}Real-life Datasets, \\      Prior knowledge bias\end{tabular}                                                                                          & \begin{tabular}[c]{@{}l@{}}Cover major datasets published   for validating the \\ Visual Question Answering task,\\      Discuss state-of-the-art architectures and compare results\end{tabular}                                    \\ \midrule
Patel et al. \cite{patel2021recent}                                & 2021 & \begin{tabular}[c]{@{}l@{}}Video Question Answering,\\      Temporal reasoning\end{tabular}          & \begin{tabular}[c]{@{}l@{}}Collection of Video-based QA   dataset,\\      Video content must have varied\\      actions\end{tabular}                                              & Review a number of methods and   datasets for VideoQA                                                                                                                                                                                         \\ \midrule
Faria et al. \cite{Faria2023VQA} & 2023 & \begin{tabular}[c]{@{}l@{}}Language bias for VQA,\\ Scene TextVQA,\\ OOD data reasoning\end{tabular} & \begin{tabular}[c]{@{}l@{}}Generalization in VQA,\\      Zero-shot VQA,\\      Consolidated VQA benchmark\end{tabular}                                                            & \begin{tabular}[c]{@{}l@{}}Discuss the steps involving VQA task,\\     Introduce the most recent and significant works comprising \\ strategies for   VQA pipeline\end{tabular}                                 \\ \midrule
Md.F. Ishmam et al. \cite{ISHMAM2024102270}                        & 2024 & vision language pre-training                                                            & \begin{tabular}[c]{@{}l@{}} encompass traditional VQA and \\ VLP-based methods \end{tabular}                                                                                               & \begin{tabular}[c]{@{}l@{}}Introduces a detailed taxonomy to categorize VQA, \\      Highlights the recent trends, challenges, and scopes for \\ improvement for more domain such as multimodalQA\end{tabular}                 \\ \midrule
Ma et al. \cite{10438044}                                  & 2024 & \begin{tabular}[c]{@{}l@{}}Robust VQA,\\      dataset bias\end{tabular}                              & \begin{tabular}[c]{@{}l@{}}Poor performance in out-of- \\distribution dataset of VQA    \end{tabular}                                                                                                                                 & \begin{tabular}[c]{@{}l@{}}Overview of in-distribution and out-of-distribution datasets,\\      Typology that presents existing debiasing methods\end{tabular}      \\ \midrule[1.3pt]                                  
\end{tabular}}\label{tab:survey}
\end{table}



In 2017, Wu et al. \cite{wu2017visual} provide a foundational review, offering an overall definition of VQA task types and a comprehensive overview of existing models. However, their survey did not cover more advanced VQA approaches. That same year, Kafle et al. \cite{kafle2017visual} review VQA tasks with a focus on datasets and evaluation metrics, discussing the challenges and limitations of existing methods.
Zhang et al. \cite{zhang2019information} later review information fusion methods in VQA, categorizing them into two-channel and multi-channel fusion strategies. While this work advanced the understanding of multimodal fusion, it provides limited insights into other critical components of the task. By 2020, Srivastava et al. \cite{srivastava2021visual} highlight the latest applications of deep learning methods in VQA, offering a detailed analysis of model performance. Patel et al. \cite{patel2021recent} further contribute by focusing on video-based question answering, emphasizing temporal information processing.
More recently, Faria et al. \cite{Faria2023VQA} explore the language bias problem in VQA, providing detailed analyses of scene-text datasets and strategies to address bias. Barra et al. \cite{10.1016/j.patrec.2021.09.008} and Singh et al. \cite{10104870} offered shorter surveys that summarized recent advances in neural network-based models, pre-trained language models, and their applications. Additionally, Ishmam et al. \cite{ISHMAM2024102270} presented a taxonomy of vision-language pretraining strategies and their relevance to VQA, while Ma et al. \cite{10438044} addressed dataset bias and proposed debiasing methods to enhance VQA robustness.

Compared to previous work, our survey provides an up-to-date overview of VQA development, with a particular focus on the latest models and their timeliness. We address knowledge reasoning techniques applied in recent years and the multimodal large language models in few-shot VQA, which have been underexplored in prior surveys.

\subsection{Contribution of this Survey}

In this paper, we give details of the processing models, relevant datasets and evaluation methods for the VQA task. We define the framework paradigm and taxonomy of VQA task in Fig. \ref{fig:taxonomy}, including natural language understanding of image and text and natural language inference. The main contributions of our paper are as follows:

\begin{figure}
    \centering
    \includegraphics[width=0.9\linewidth]{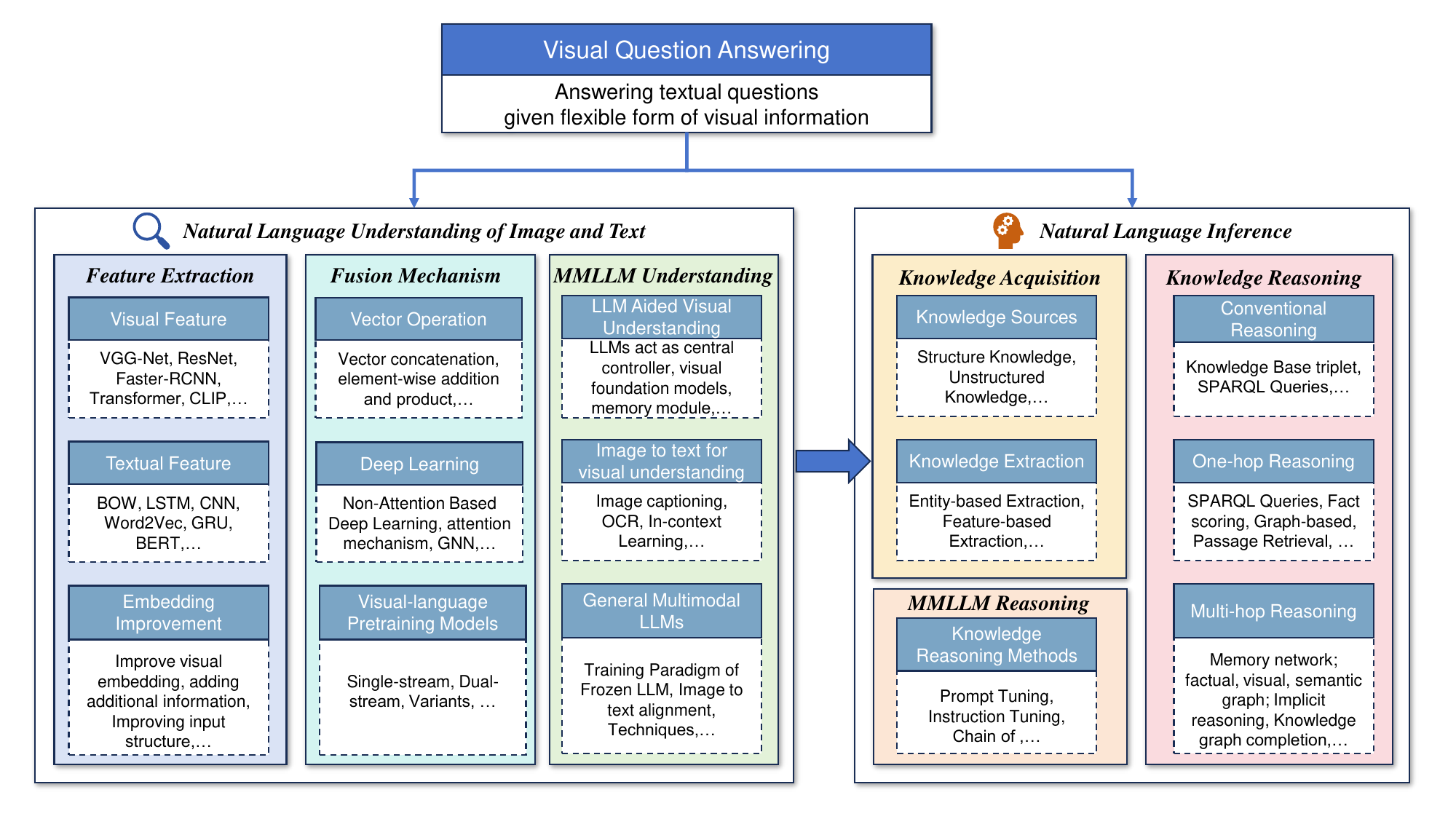}
    \caption{Taxonomy Graph of VQA Task.}
    \label{fig:taxonomy}
\end{figure}

\begin{enumerate}

    \item To give an up-to-date synthesis of VQA paradigm, including natural language understanding of images and text (perceptual ability), as well as the natural language inference module based on image-question information (cognitive ability) on the core VQA tasks, as shown in Fig. \ref{fig:taxonomy};
    \item To elaborate on the latest progress of visual-language pre-training models, graph neural network models, and multimodal large language models in VQA for natural language understanding of images and text;
    \item To highlight the progress of natural language inference in VQA by detailing the knowledge reasoning and multimodal large language model reasoning methods.
\end{enumerate}

The rest of this survey is organized as follows: In Sec. \ref{sec4}, we discuss natural language understanding for images and text, focusing on improved extraction, embedding, and multimodal fusion methods. Sec. \ref{MLLM} reviews large language models in zero-shot VQA, introducing three paradigms for their application. In Sec. \ref{sec5}, we explore natural language inference, including knowledge sources, acquisition, and reasoning processes, along with multimodal large language model reasoning methods in VQA. Finally, we discuss VQA challenges and propose future research directions in Sec. \ref{sec7}, concluding in Sec. \ref{sec8}.

\section{COMPREHENSION OF IMAGE AND TEXT IN VQA}\label{sec4}
\subsection{Feature Extraction}\label{featureextraction}
The majority of VQA models require modal feature extraction prior to answering questions, which can be used for subsequent multimodal feature fusion to eliminate the gap between modals. 

\subsubsection{Visual Feature Extraction}
For visual feature extraction, on the one hand, convolutional neural networks (CNNs) (e.g., VGG-Net \cite{simonyan2014very}, GoogleNet \cite{szegedy2015going} and ResNet \cite{he2016deep}) are often used to extract global image features in early VQA tasks. 
VGG-Net \cite{simonyan2014very} increases the convolutional layers to 19 and replaces $5 \times 5$ with $3 \times 3$ convolutions, reducing parameters and enhancing nonlinear mapping for improved expressive ability. ResNet \cite{he2016deep} introduces residual blocks to mitigate gradient issues in deeper networks, weakening strong connections. 
On the other hand, Visual Transformer (ViT) \cite{dosovitskiy2020image} applies the Transformer \cite{vaswani2017attention} framework to extract image features and compute attention map of the image by attending each pixel to every other pixel. Based on this, CLIP \cite{radford2021learning} shows the strong performance to extract features on pixel-level, and further becomes a widely-used image encoder to assistant the vision-language pretraining models and multimodal large language models with pixel-level image understanding ability.

Using global image features to perform VQA task weakens the relationship between tasks and objects in the image, so numerous models prominent the task-relevant regions by extracting region-based image features. Actually, it is a spatial-attention mechanism to extract finer-grained features. They first select the regions of interest to the task and input them into CNNs to extract region features. Specifically, there are two ways to extract region-based features. One is based on the uniform grid \cite{jiang2020defense}. By dividing image into uniformly sampled grids, the region features corresponding to each grids can be achieved after inputting them into CNNs. And the relevance weight of each grid feature is determined by the task. Another way is based on region proposal, which applys object recognition techniques to generate bounding boxs for the image, then inputs them with their corresponding size and position information into CNNs to extract region features. Compared with global-based features, this can better identify objects attribute, quantity and location relationship \cite{anderson2018bottom}. The commonly used object recognition techniques is Faster RCNN \cite{ren2015faster}.

\begin{table*}[]
\centering
\caption{The VQA models with different published year, architecture, and dataset.}
\label{tab2}
\scalebox{0.85}{
\begin{tabular}{c|c|ccc|c}
\midrule[1.3pt]
\multirow{2}{*}{Models} & \multirow{2}{*}{Year} & \multicolumn{3}{c|}{Architecture}                                    & \multirow{2}{*}{Datasets} \\ 
                        &                       & Visual Feature & Textual Feature & Fusion Strategy                  &                           \\ \midrule
IBOWIMG                 & 2015                  & GoogLeNet      & BoW             & Vector Concatenation             & DAQUAR, VQA1.0, MS COCO   \\
ABC-CNN                 & 2015                  & VGG -Net       & LSTM            & Element-Wise Addition, CNN       & DAQUAR, VQA1.0, COCO-QA   \\
SAN                     & 2016                  & VGG -Net       & LSTM            & Element-Wise Addition, Attention & VQA1.0, COCO-QA           \\
Full-CNN                & 2016                  & VGG -Net       & CNN             & CNN                              & DAQUAR, COCO-QA           \\
AMN                     & 2020                  & VGG -Net       & Word2Vec        & Attention                        & MovieQA                   \\
Marioqa                 & 2017                  & C3D            & GRU             & Attention                        & CLEVER, MovieQA           \\
MLB                     & 2016                  & ResNet         & GRU             & Bilinear Pooling Fusion          & VQA1.0                    \\
MCB                     & 2016                  & ResNet         & LSTM            & Bilinear Pooling Fusion          & VQA1.0, VQA2.0, Visual7W  \\
Pixel-BERT              & 2020                  & ResNet         & BERT            & Transformer                      & VQA1.0, VQA2.0            \\
SOHO                    & 2021                  & ResNet         & BERT            & Transformer                      & VQA1.0, VQA2.0            \\
LXMERT                  & 2019                  & Faster-RCNN    & BERT            & Cross-Modal Transformer          & A-OKVQA,   GQA, VizWiz    \\
ViLBERT                 & 2019                  & Faster-RCNN    & BERT            & Cross-Modal Transformer          & A-OKVQA,   VQA2.0         \\
Oscar                   & 2020                  & Faster-RCNN    & BERT            & BERT                             & VQA2.0                    \\
ConceptBert             & 2020                  & Faster-RCNN    & BERT            & Transformer                      & A-OKVQA, VQA1.0           \\
MuKEA                   & 2022                  & Faster-RCNN    & BERT            & LXMERT                           & VQA2.0, A-OKVQA, KRVQA    \\
ViLT                    & 2021                  & Transformer    & BERT            & ViT                              & VQA2.0                    \\
ALBEF                   & 2021                  & Transformer    & BERT            & Cross-Modal Transformer          & VQA2.0             \\ \midrule[1.3pt]      
\end{tabular}}
\end{table*}

\begin{figure}
    \centering
    \includegraphics[width=0.65\linewidth]{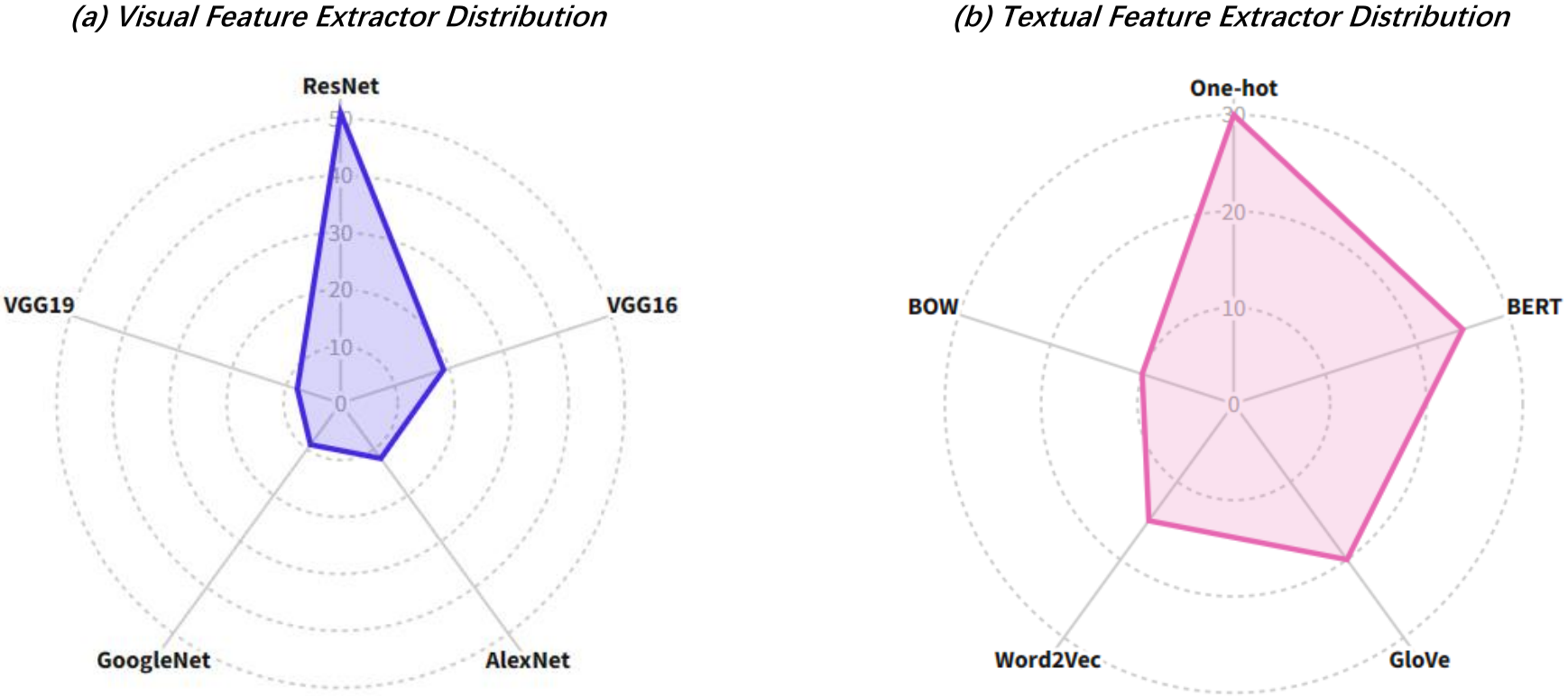}
    \caption{Percentage distribution of the usage of visual and textual feature extractors.}
    \label{fig:feature}
\end{figure}

For VQA tasks with video input, temporal channels are typical employed for visual feature extraction. Two common approaches are utilized for this purpose. One is C3D \cite{tran2015learning}, which scales convolutional networks to three dimensions to capture temporal information effectively. Alternatively, optical flow \cite{farneback2003two} can be used to extract video features. This method analyzes pixel movement and frame-to-frame correlation to capture temporal dynamics. Please refer to Table \ref{tab2} and Fig. \ref{fig:feature} for a summary of various methods for extracting visual features.

\subsubsection{Textual Feature Extraction}
Another crucial aspect is the extraction of texture features. Classic models for text encoding include Bag-of-Words (BoW) and Word2Vec \cite{mikolov2013distributed}. 
With the emergence of Recurrent Neural Networks (RNNs), models like LSTM, Bi-LSTM, and GRU have been extensively utilized for handling sequential data. 
In addition, CNNs, originally prominent in computer vision, have also been applied for text feature extraction.
Word2Vec, RNN and CNN are all common models for extracting textual features until the advent of Bert \cite{devlin2018bert}.
Bert, namely Bi-directional Encoder Representation from Transformers, is a pre-training model proposed by Google AI in 2018. Bert's architecture is based on bidirectional Transformer components, and remarkably, fine-tuning with an additional output layer achieves outstanding performance across various downstream tasks. Bert stands as a milestone work in the history of NLP. 

We give a summary of textual feature extraction models in Table \ref{tab2}, and conclude the visual and textual feature extrctor distribution in Fig \ref{fig:feature} to better show what the most used feature extraction methods are. 

\subsubsection{Improvement of Embedding}
Numerous methods have been introduced for extracting embeddings of different modalities. To adapt to more complex questions and generate more accurate answers, significant improvements have been made in feature embedding, including Improving Visual Embedding \cite{huang2020pixel,huang2021seeing}, Adding Additional Information \cite{li2020oscar,garderes2020conceptbert}, and Improving Input Structure \cite{yang2021auto,lafferty2001conditional}. We give the detailed improvement methods in Appendix. A.

\subsection{Fusion Mechanism}
\subsubsection{Simple Vector Operations}
Traditional fusion methods typically utilize simple vector operations like element-wise addition, element-wise product, and vector concatenation to directly operate on visual features $v_I$ and text features $v_Q$ for generating joint representations.
For element-wise addition and product, $v_I$ and $v_Q$ need to be related in the same dimension. In cases where they are not, a linear projection \cite{nam2017dual} can be employed to embed these vectors into the same space using transformation matrices $W_v$ and $W_q$. Vector connection splices $v_I$ and $v_Q$ together as fusion vector $v_F$. 


In general, the use of element-wise addition does not increase extra computation, and is therefore a common feature fusion method. In contrast, vector concatenation increases computational complexity significantly.

\subsubsection{Deep Learning Method}
The fusion of visual and textual features using deep learning can be categorized into three main approaches: non-attentional deep learning, attention-based deep learning, and graph neural networks.

\paragraph{\textbf{Non-Attention Based Deep Learning}} 
CNNs and RNNs are commonly utilized for non-attention-based multimodal fusion. Ren et al. \cite{ren2015exploring} propose aligning image features $v_I$ with word embeddings, feeding them into an LSTM model alongside question words $v_{Q_i}$. Ma et al. \cite{ma2018visual} leverage memory-augmented neural networks for visual question answering (VQA), using memory modules to maintain longer-term information.

CNN-based approaches have also been explored. Ma et al. \cite{ma2016learning} proposed an end-to-end framework incorporating image, sentence, and multimodal CNNs to enhance image-question interplay. Noh et al. \cite{noh2016image} use a modified VGG-16 for image feature extraction and GRU for text, enabling adaptive parameter prediction. To preserve spatial information, Gao et al. \cite{gao2018question} introduce Question-Guided Hybrid Convolution, using question-guided kernels to learn discriminative multimodal feature representations.

Bilinear pooling is widely employed for fine-grained visual recognition. Fukui et al. \cite{fukui2016multimodal} introduce Multimodal Compact Bilinear Pooling (MCB) to compress bilinear models, efficiently combining multimodal features. Schwartz et al. \cite{schwartz2017high} use MCB with attention mechanisms to capture high-order correlations, while Yu et al. \cite{yu2017multi} propose Multimodal Factorization Bilinear (MFB) pooling for effective fusion, employing sum pooling to reduce dimensionality.

\begin{figure}[t]
    \centering
    \includegraphics[width=0.6\linewidth]{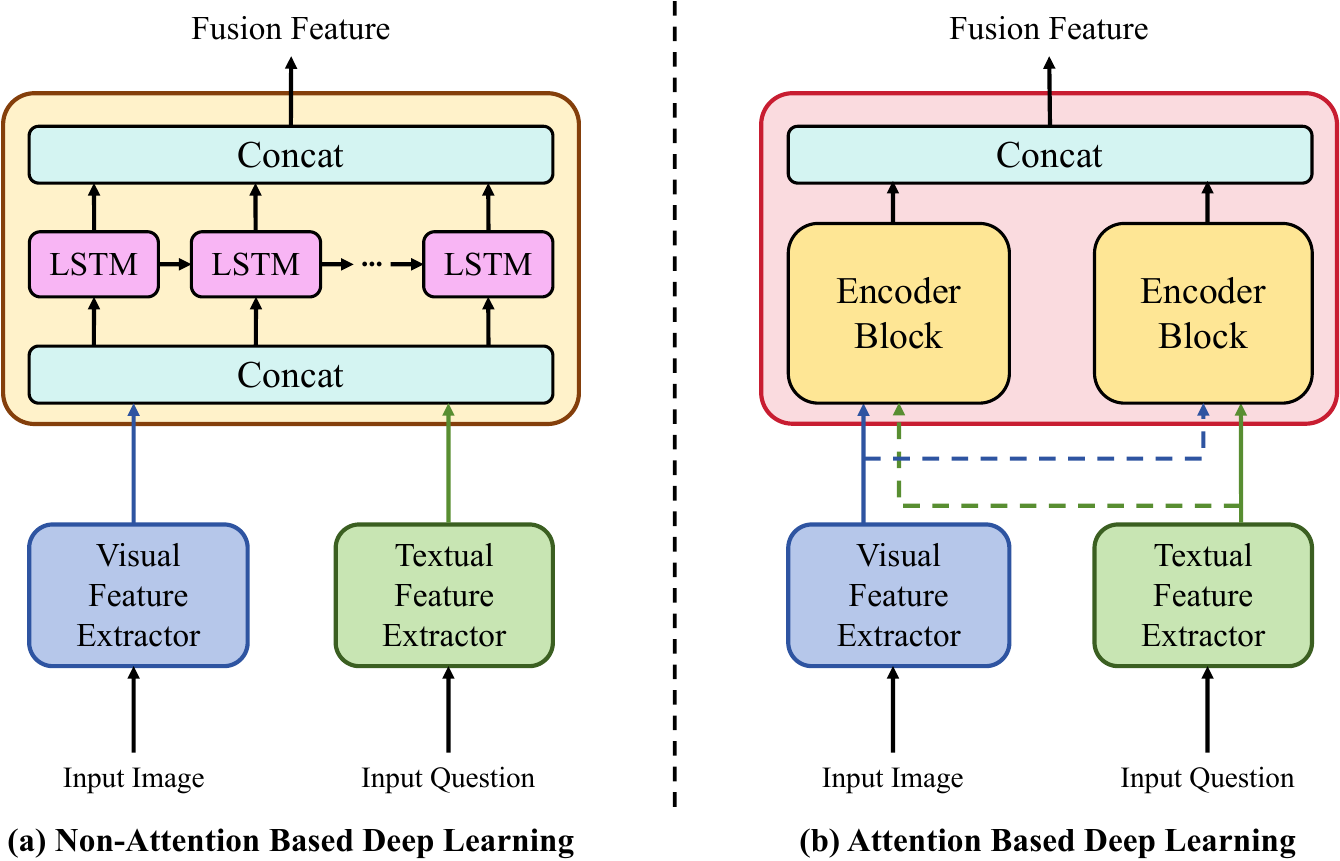}
    \caption{Comparison of fusion mechanism utilizing (a) non-attention based deep learning and (b) attention based deep learning.}
    \label{fig:attention}
\end{figure}

\paragraph{\textbf{Attention Based Deep Learning}}
Based on the information in the question, there are parts of the image that are
more relevant to the question, which is also the part of the model that needs more attention. Therefore, the attention mechanism is introduced into the multimodal fusion mechanism of VQA, and Figure \ref{fig:attention} compares the non-attention and attention-based deep learning methods. Li et al. \cite{li2016visual} propose QRU, iteratively updating question representations based on relevant image regions. Shih et al. \cite{shih2016look} introduce edge boxes to obtain image regions, and selectively combine image region features with text features, marking an early instance of attention-based deep learning methods in VQA. It maps the region image features $V=(v_1,v_2,...,v_m)$ and text features $q$ to a common $n$-dimensional space, then calculates the inner product between regions and question answers to determine relative weights. 

Researches establish loose, global associations between questions and images. Zhu et al. \cite{zhu2016visual7w} propose LSTM-Att to establish semantic associations between image regions and text descriptions, capturing specific associations between image-related questions and regions. Attention models have evolved to capture finer-grained associations. 
However, it is not enough to only focus on local regions in visual features, and it is equally important to determine which words need to be focused on in the problem. Lu et al. \cite{lu2016hierarchical} introduce co-attention mechanisms to jointly perform visual and question-guided attention. The joint attention is calculated as follows:
\begin{equation}
    C=tanh(Q^TW_bV),
\end{equation}
\begin{equation}
    H^v=tanh(W_vV+(W_qQ)C),
\end{equation}
\begin{equation}
    H^q=tanh(W_qQ+(W_vV)C^T),
\end{equation}
where $V$ is a visual feature, $Q$ is a text feature, $C$ is co-attention, and $W$ is the weight parameter.

Previous approaches mainly exploit low-level features while ignoring the rich semantics contained in high-level features. Yu et al. \cite{yu2017multi} present a multi-level attention network to reduce the semantic gap and visual attention for fine-grained spatial reasoning, aligning image regions and questions through feature multiplication for fine-grained spatial reasoning. To further enhance multimodal fusion, Nguyen et al. \cite{nguyen2018improved} propose a symmetrical co-attention mechanism, where each word in the question attends to image regions and vice versa. Wu et al. \cite{wu2018chain} addressed complex reasoning tasks by introducing a chain of reasoning model, enabling dynamic relational reasoning between objects.

\paragraph{\textbf{GNN-based Fusion Approaches}}
Traditional VQA methods often ignore the structural information in images or questions. Recent approaches leverage Graph Neural Networks (GNNs) for better feature fusion. Norcliffe et al. \cite{norcliffe2018learning} construct scene graphs conditioned on the question, using a $K$-kernel Graph Convolutional Network (GCN) to capture object interactions. Li et al. \cite{li2019relation} propose Relation-Aware Graph Attention Networks (ReGAT), encoding semantic, spatial, and implicit relations through attention mechanisms sensitive to node features and positional similarity:
\begin{equation}
    \alpha_{ij}=\frac{\alpha^b_{ij}\exp((\mathbf{U}v_i')^{T}\mathbf{V}v_j')}{\sum_{j\in\mathcal{N}(i)}\alpha^b_{ij}\exp((\mathbf{U}v_i')^{T}\mathbf{V}v_j')}.
\end{equation}
For explicit relations, the attention mechanism accounts for edge labels and directions, allowing the model to capture richer semantic dependencies.

In addition to image graphs, Teney et al. \cite{teney2017graph} construct a question graph based on the token syntactic relations. Subsequently, a GRU-based GNN is deployed to aggregate first-order information, and a cross-modal attention mechanism aligns textual tokens $\{x^{\prime Q}_{i}\}^{N^{Q}}_{i=1}$ from the question graph with visual objects $\{x^{\prime S}_{i}\}^{N^{S}}_{i=1}$ of the scene graph as follows:
\begin{equation}
    \alpha _{ij}=\sigma \left ( W_{5}\left ( \frac{x^{\prime Q}_{i}}{|| x^{\prime Q}_{i} ||}\circ \frac{x^{\prime S}_{i}}{|| x^{\prime S}_{i} ||}  \right ) +b_{5} \right ).
\end{equation}

Instead of using fully-connected graphs to represent images and questions, Huang et al. \cite{huang2020aligned} prune edges in the visual and question graphs based on the object overlapping region and syntactic dependencies respectively. In the aggregation stage, a dual-channel graph convolutional network simultaneously captures relations between objects in images and relations among textual tokens in questions.

Transforming questions into instructions to guide scene graph learning has garnered recent attention. Shi et al. \cite{shi2019explainable} propose using NLP tools to parse a given problem into a series of programs, and select different neural modules to infer the scene graph according to the corresponding programs. Since the instructions represented by the program set are finite and discontinuous, Hu et al. \cite{hu2019language} parse questions into several textual instruction embeddings $\{c_{t}\}^{T}_{t=1}$ which are used to guide the process of message passing.
In the procedure of scene graph learning, a GCN conditioned on instruction embeddings dynamically predicts edge weights $w^{(t)}_{j,i}$ to focus on different connections and aggregates information from neighboring nodes $\tilde{x}_{j,t}$ to the target $\tilde{x}_{i,t}$ in each iteration.
Liang et al. \cite{liang2020lrta} regard VQA as an answer generation task and propose a model LRTA consisting of four stages (Look, Read, Think, Answer). LRTA parses the problem into instructions using a transformer-based framework and traverses the scene graph using a recursive neural symbolic execution module that executes one instruction at each inference step.

\subsubsection{Vision-Language Pre-training Models}

\begin{figure}
    \centering
    \includegraphics[width=0.65\linewidth]{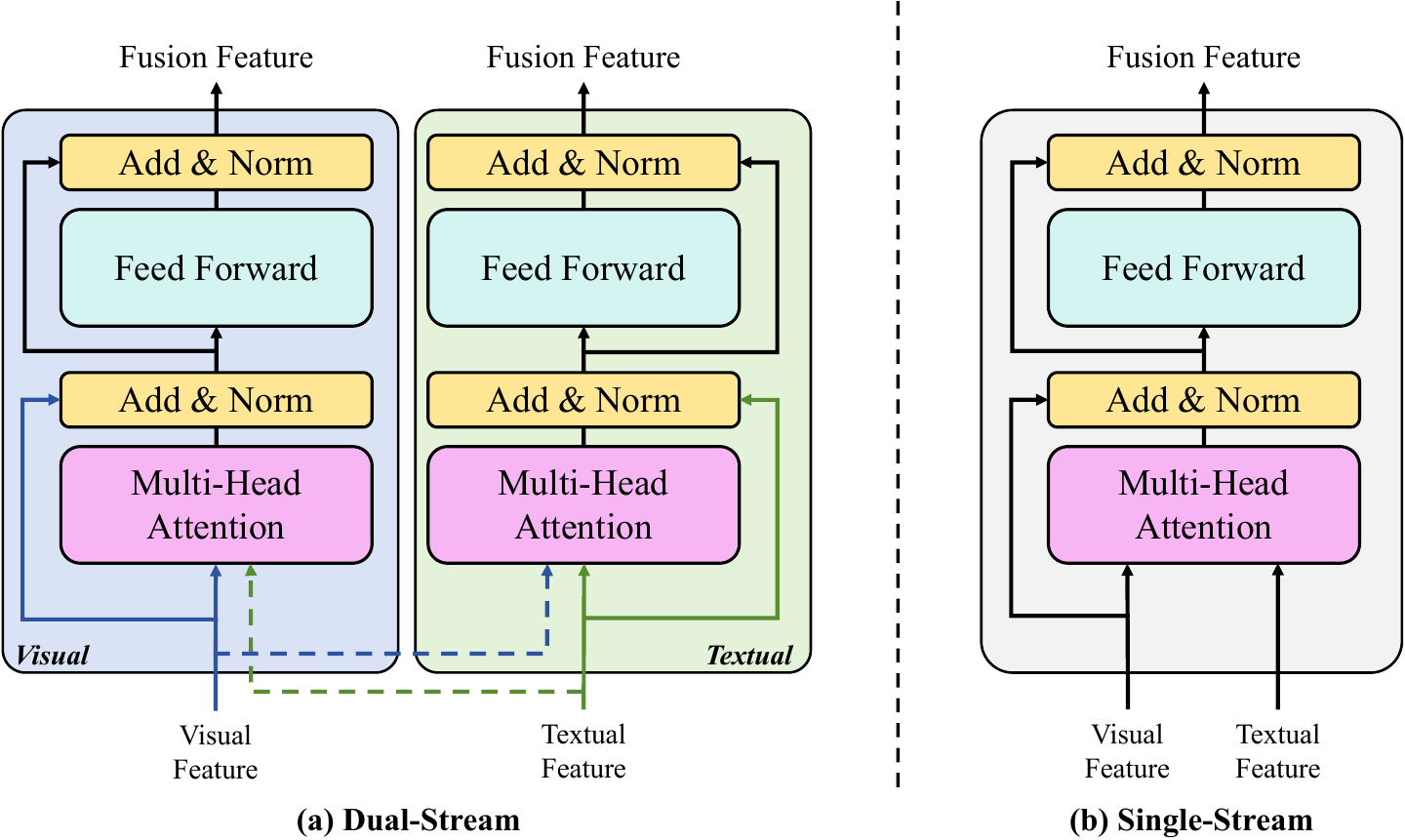}
    \caption{Comparison of fusion mechanism utilizing (a) dual-stream pre-trained models and (b) single-stream pre-trained models.}
    \label{fig:stream}
\end{figure}

Vision-Language Pre-training models are trained on large-scale unlabeled data via self-supervision and fine-tuned for specific tasks, allowing for knowledge transfer and improved performance with minimal labeled data. In multimodal research, methods fall into three categories: Dual-Stream, Single-Stream, and other variants. Fig. \ref{fig:stream} provide a comparison of Dual-Stream and Single-Stream.


\paragraph{\textbf{Dual-Stream}}

The dual-stream paradigm in multimodal research involves processing visual and textual inputs separately before integrating them through a cross-modal fusion module. Exemplifying this approach are ViLBERT \cite{lu2019vilbert} and LXMERT \cite{tan2019lxmert}. ViLBERT extends the BERT architecture to a multimodal cross-stream model, employing separate Transformers for visual and textual inputs and integrating them through a co-attention module. Similarly, LXMERT introduces an architecture to learn the language-vision connection, utilizing object relationship encoders and pre-training tasks to capture intra-modal and inter-modal relations.

In recent advancements, ERNIE-ViL \cite{yu2021ernie} proposes integrating structured knowledge from scene graphs to enhance semantic alignment. This model leverages the structured knowledge obtained from scene graphs to facilitate fine-grained semantic understanding. Building upon previous research, Dou et al. \cite{dou2022empirical} present the METER model, which refines the dual-stream approach by stacking transformer layers with self-attention, co-attention, and feedforward networks. METER conducts comprehensive experiments across various aspects of general pre-training models, providing insights into the efficacy of different architectural components.

\paragraph{\textbf{Single-Stream}}

Different from Dual-Stream approaches, Single-Stream models integrate textual and visual inputs for semantic learning. Li et al. \cite{li2020unicoder} propose Unicoder-VL, which matches specific text phrases with image features and inputs them jointly into a multi-layer Transformer for cross-modal representation learning. They introduce a formulation to combine text and image features, leveraging both region and location information. To mitigate overfitting with limited target tasks, Su et al. \cite{su2019vl} train VL-BERT on large-scale image description and plain text datasets simultaneously, enabling the learning of more general feature representations. Their model employs stacked multi-layer Transformer modules to adaptively aggregate information from different modalities.

Chen et al. \cite{chen2020uniter} propose UNITER, a general image-text representation learning model that conducts fine-grained semantic alignment between words and image regions. They introduce a novel pre-training task with conditional masking, enhancing the alignment process. Meanwhile, Kim et al. \cite{kim2021vilt} present ViLT, a lightweight model focused on modal interactions without relying on region or deep convolutional features. Instead, they use a pre-trained Vision Transformer (ViT) to extract visual features. In Single-Stream models, it's impractical to encapsulate intra-modal and vision-language learning in the same Transformer. Xue et al. \cite{xue2021probing} introduce self-attention to the visual domain to facilitate learning visual modalities. They utilize Swin Transformer \cite{liu2021swin} to obtain visual feature embeddings and perform visual masking based on internal attention scores from the inter-modal Transformer.

Previous pre-trained models rely heavily on the image feature extraction process (such as Faster R-CNN \cite{ren2015faster} and ResNet \cite{he2016deep}), requiring high hardware equipment and time-consuming training. Kim et al. \cite{kim2021vilt} propose the lightweight model ViLT, the main computation of which is concentrated on the modal interactions. ViLT does not use region features or deep convolutional features but uses a pre-trained ViT model \cite{dosovitskiy2020image} to extract visual features.
\begin{equation}
    \hat{z}^d=MSA(LN(z^{d-1}))+z^{d-1},
\end{equation}
\begin{equation}
    z^d=MLP(LN(\hat{z}^d))+\hat{z}^d,
\end{equation}
where $z$ indicates the vision-language embedding, $LN$ means LayerNorm, and $MSA$ indicates multiheaded self-attention.


\paragraph{\textbf{Variants and Improvements in Vision-Language Pre-training Approaches}}
There are some variants and improvements in training techniques, with the detailed introduction in Appendix. B.

Early methods such as ViLBERT \cite{lu202012} employ task-specific networks for vision-language multi-task learning. Lu et al. introduced a shared backbone with task-specific layers for each task, optimizing model parameters across different vision-language tasks.To enhance model robustness, adversarial learning was applied to vision-language pre-training by Gan et al. \cite{gan2020large}, who proposed the VILLA framework. This method introduces adversarial noise into the image and word embedding spaces, improving model generalization.
Contrastive learning is introduced in the multimodal domain by Li et al. \cite{li2020unimo}. UNIMO leverages a three-stream architecture to process language, visual, and cross-modal fusion independently, optimizing representations for both single- and multi-modal tasks. The objective function involves maximizing the similarity between positive pairs of image and text embeddings:
\begin{equation}
    \mathbb{E}_{V,W}[-\log\frac{\exp(d(V^+,W^+)/\tau)}{\sum_{\mathcal{X}} \exp(d(V',W')/\tau)}].
\end{equation}

Radford et al. \cite{radford2021learning} introduced CLIP, a significant step forward in learning image-text relationships through contrastive learning. CLIP normalizes word and region embeddings and computes their similarity via dot product, learning robust representations for zero-shot vision-language tasks, where $I_e$ and $T_e$ represent the joint multimodal features:
\begin{equation}
    I_e = L2Norm(I_f, W_i), \quad T_e = L2Norm(T_f, W_t).
\end{equation}


Li et al. \cite{li2021align} further improved vision-language models with ALBEF, which leverages momentum distillation to align image-text pairs and combat noisy data, introducing image-text contrastive loss.
Recent models like BLIP \cite{li2022blip} and PNP-VQA \cite{DBLP:conf/emnlp/Tiong0LSH22} have further extended this line of work by integrating multimodal generation and understanding within a unified framework, demonstrating strong performance in zero-shot VQA tasks.

These vision-language pre-training methods have not only bridged the gap between vision and language understanding in VQA but have also become essential for aligning visual and textual information. This alignment serves as a foundational element for transitioning from purely text-based Large Language Models to multimodal Large Language Models.

\section{COMPREHENSION OF IMAGE AND TEXT WITH LLMs IN ZERO-SHOT VQA}\label{MLLM}
As text large language models (LLMs) have shown amazing performance in multiple textual tasks and attracted great attention from the whole community, more and more research attempts have been made to explore the introduction of LLMS into other domains \cite{alayrac2022flamingo}. However, LLMs cannot directly process image information, so there is a rising need for generalized multimodal large language models (MLLMs) to accomplish various multimodal tasks \cite{koh2024generating}.

\subsection{LLM Aided Visual Understanding}

\begin{figure}[t]
    \centering
    \includegraphics[width=0.85\linewidth]{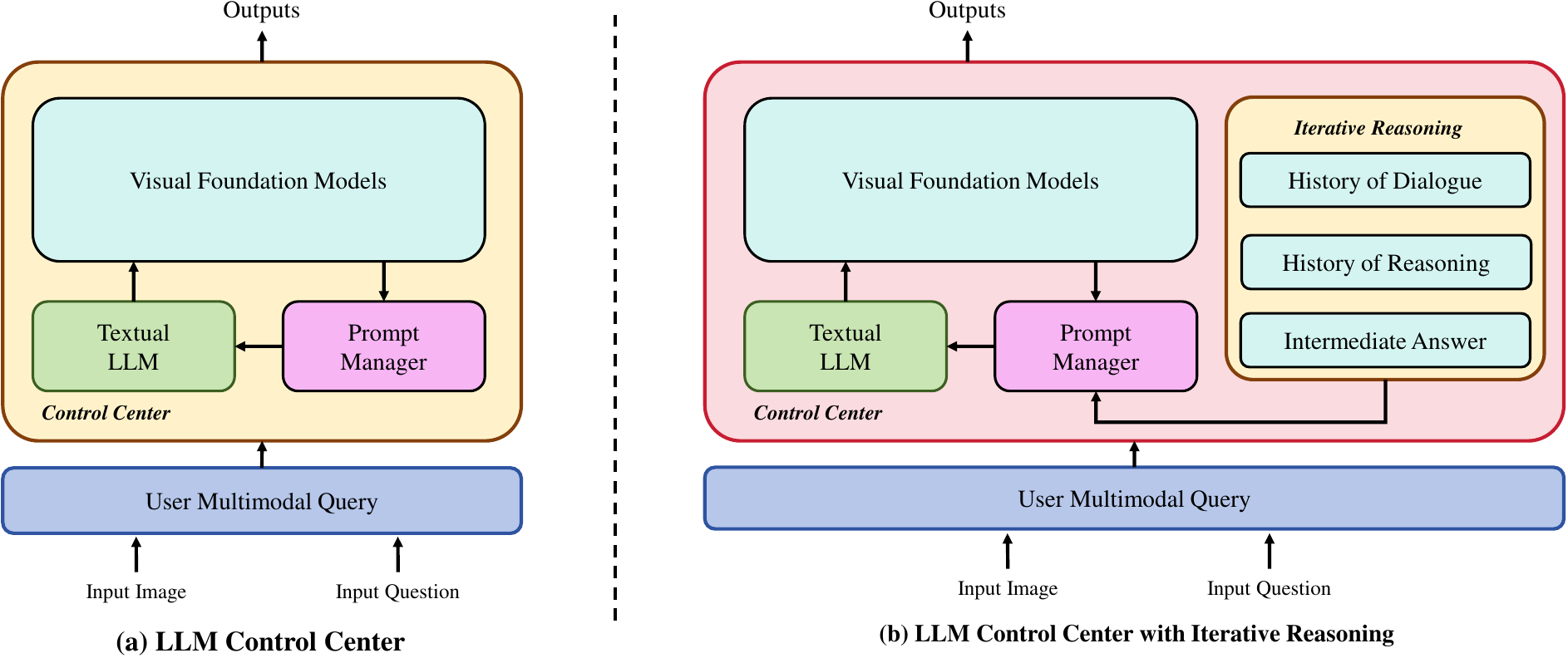}
    \caption{Two architectures of the LLM aided visual understanding models.}
    \label{fig:3-1}
\end{figure}

Since Large Language Models (LLMs) are text-based, early Multimodal Large Language Models (MLLMs) explore leveraging LLMs as central controllers for multimodal tasks \cite{gupta2023visual}. LLMs act as a central controller that (1) analyze the prompt and history of dialogues, (2) split a complex task into simpler sub-tasks and (3) assign these tasks to appropriate models. Fig.\ref{fig:3-1} shows the LLM aided visual understanding models. For instance, Microsoft Visual ChatGPT \cite{wu2023visual} integrates LLMs with various Visual Foundation Models (VFMs) (e.g., BLIP \cite{li2022blip}, Stable Diffusion, ControlNet), using a Prompt Manager to manage input/output formats. The LLM analyzes prompts, divides tasks, and invokes VFMs to generate outputs. Additionally, Visual ChatGPT utilizes dialogue history management and iterative reasoning to invoke further VFMs for more accurate results.

MM-REACT \cite{yang2023mm} focuses on broader visual interpretation by incorporating Azure APIs for tasks like celebrity recognition and Bing search. This enhances the LLM's role as a controller for visual understanding and interaction. As tasks grow more complex, LLMs evolve into decision-makers, as in IdealGPT \cite{you2023idealgpt}, where autonomous agents handle complex tasks. These agents consist of modules for profiling, memory, planning, and action \cite{shen2024hugginggpt}, allowing the LLM to understand dynamic environments and organize responses effectively.

\subsection{Image to Text Generation for Visual Understanding}

For the issue that textual large language models cannot process images, another inspiration is to transform images directly into corresponding textual descriptions \cite{yang2022empirical}, as shown in Fig.\ref{fig:3-2}. This typically involves models like Image Captioning and Optical Character Recognition (OCR) \cite{stefanini2022show}. For example, PICA \cite{yang2022empirical} uses image captioning to generate textual descriptions, which are concatenated with questions and fed into the LLM for question answering. To improve In-context Learning, PICA selects 16 training examples closest to the current test image-question pair using CLIP \cite{radford2021learning}.

However, converting images into text can lead to inaccuracies or loss of essential visual details. To address this, IMG2LLM \cite{guo2023images} generates more relevant captions and question-answer examples directly from the image. This model focuses on selecting image regions pertinent to the question and refining the captions for accuracy. It also synthesizes question-answer pairs to provide more representative in-context prompts. Prophet \cite{shao2023prompting} enhances the selection of in-context examples and the generation of answer heuristics for VQA tasks. Prophet uses a Vanilla VQA \cite{jiang2020defense} model to generate answer candidates, which serve as examples. These answer-aware heuristics, along with the testing samples, are input into the LLM as prompts, improving VQA performance.

\subsection{General Multimodal LLM}
\begin{figure}
    \centering
    \includegraphics[width=0.85\linewidth]{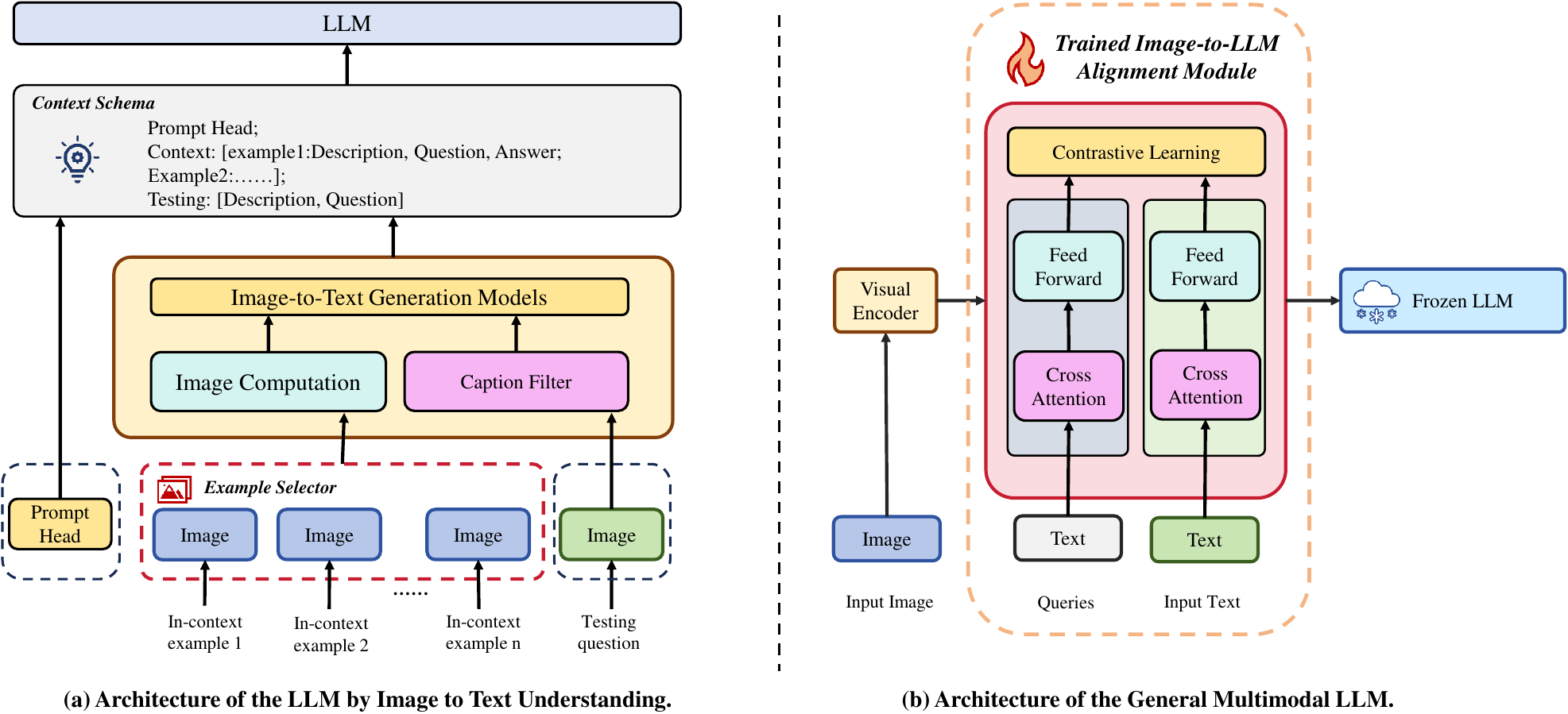}
    \caption{Architecture of the LLM by Image to Text Understanding and General MLLM.}
    \label{fig:3-2}
\end{figure}
LLM-aided visual understanding and image-to-text generation methods demonstrate the ability to leverage pure-text LLMs for multimodal tasks. However, these approaches do not inherently grant LLMs image comprehension capabilities. Instead, LLMs either rely on visual foundation models for image processing or convert images into textual representations. To address this, there is growing interest in developing general-purpose MLLMs with direct image comprehension abilities by integrating additional modalities into a unified framework, as shown in Fig. \ref{fig:3-2}.

\subsubsection{The Training Paradigm of General Multimodal LLM}
The training architecture for general multimodal LLMs, typically using frozen LLMs, follows a two-stage process, as shown in Fig \ref{fig:3-2}. First, the LLM is frozen while external components, such as visual encoders and alignment modules, are trained to map image features into the textual space of the LLM \cite{wang2024visionllm}. In the second stage, the visual components are frozen, and the LLM is fine-tuned with multimodal data, often using techniques like LoRA or Q-LoRA \cite{qlora}. This method significantly reduces the cost of extending LLMs to multimodal tasks, with a focus on designing effective image-to-text alignment modules, instead of training a new multimodal large model. Thus, it plays a pivotal role in the advancement of MLLMs, with the core challenge of designing more effective Image-to-Text alignment modules.

\subsubsection{Image-to-Text Alignment in Multimodal LLM}

Flamingo \cite{alayrac2022flamingo} introduces a vision encoder and a Perceiver Resampler to generate a fixed-length feature sequence, integrated with cross-attention layers for improved visual-textual alignment. PaLM-E \cite{driess2023palme} integrates visual features into the pre-trained PaLM model, leading to robust performance across real-world tasks, a method adopted by models like LLaVA \cite{liu2023llava} and Shikra \cite{chen2023shikra}. However, freezing the LLM during training can limit alignment effectiveness. BLIP-2 \cite{li2023blip} addresses this by proposing the Q-Former, using a two-stage pre-training approach where the visual encoder learns critical visual features through contrastive learning and image-text matching. This method enhances zero-shot capabilities, though it shows limited in-context learning improvements.

To reduce computational demands, smaller and faster models like PaLI-3 \cite{PaLI-3}, based on SigLIP \cite{DBLP:conf/iccv/ZhaiM0B23}, offer competitive performance across multimodal benchmarks while requiring fewer resources.

\subsubsection{Advanced Closed and Open-source General MLLMs}

As MLLM continues to develop, researchers continue to expand the data and parameter scale for MLLM pre-training, and continue pre-training, hoping to teach MLLM more and more new general knowledge and make it more powerful. In this context, the development of MLLM is no longer based on the improvement of VQA, a downstream task, but more focused on its general capabilities and performance in multiple tasks in the entire multimodal field. Many commercial models have emerged, such as OpenAI's GPT series, including GPT-4v with image and text interleaving processing capabilities \cite{chen2024far}, and GPT-4o frameworks with more powerful multimodal general capabilities and support for more multimodal inputs \cite{islam2024gpt}, such as Google's Genimi and Gemini Pro series \cite{team2023gemini}. 

In parallel, the open-source community has developed several high-performing models that rival, and in some cases surpass, these commercial counterparts. The \textbf{mPLUG series} \cite{mplug-owl2}, for example, has been praised for its versatile cross-modal capabilities. mPLUG-OWL excels in open-world vision-language tasks, benefiting from its ability to handle a wide range of image-text pairs and unstructured data sources. 
The \textbf{InternVL series}, such as InternVL-2 or InternLM-XComposer2.5-VL\cite{chen2024internvl}, distinguishes itself by focusing on improving vision-language alignment through sophisticated pre-training techniques, allowing for highly effective cross-modal understanding. which has proven especially effective in sophisticated tasks like math reasoning. 
Similarly, the \textbf{LLaVA series} \cite{liu2024visual} has garnered attention for its innovative approach to integrating large-scale vision-language models. Through a carefully curated combination of instruction tuning and multimodal dialogue datasets, the LLaVa series has achieved strong performance in both visual comprehension and interactive reasoning.


\begin{table}[]
\caption{MLLM models with their base LLM models, MLLM types, Techniques and Performances}
\scalebox{0.73}{
\begin{tabular}{c|c|c|ccc|ccc|c|cc}
\midrule[1.3pt]
\multirow{2}{*}{Models}       & \multirow{2}{*}{Year} & \multirow{2}{*}{Base}    & \multicolumn{3}{c|}{Type}       & \multicolumn{3}{c|}{Technique}             & \multirow{2}{*}{Setup}        & \multicolumn{2}{l}{Performance on VQA} \\
                              &                       &                          & LLM-aid & img2txt & general mm & SFT & ICL & CoT &              & OK-VQA          & VQAv2          \\ \midrule[1.3pt]
Viusal ChatGPT \cite{wu2023visual}               & 2023                  & ChatGPT text-davinci-003 & \Checkmark       &         &            & \Checkmark      &     & \Checkmark    &       -       & \multicolumn{2}{c}{Case study}         \\ \midrule
MM-REACT \cite{yang2023mm}                     & 2023                  & gpt-3.5-turbo            & \Checkmark       &         &            & \Checkmark      &     & \Checkmark        &   -           & \multicolumn{2}{c}{Case study}         \\ \midrule
\multirow{2}{*}{PICa \cite{yang2022empirical}}          & \multirow{2}{*}{2022} & GPT-3 (175B)             &         & \Checkmark       &            &        & \Checkmark   &                   & few-shot     & 46.9            &   -                   \\
                              &                       & GPT-3 (175B)             &         & \Checkmark       &            &        & \Checkmark   &              & few-shot     & 48.0            & -                     \\ \midrule
\multirow{2}{*}{Img2LLM \cite{guo2023images}}      & \multirow{2}{*}{2023} & OPT-3 (66B)              &         & \Checkmark       &            &        & \Checkmark   &                   & zero-shot    & 43.2            & 60.3                 \\
                              &                       & OPT-3 (175B)             &         & \Checkmark       &            &        & \Checkmark   &                   & zero-shot    & 45.6            & 61.9                 \\ \midrule
Prophet \cite{shao2023prompting}                      & 2023                  & GPT-3 API                &         & \Checkmark       &            &        & \Checkmark   &                   & few-shot     & 61.1            & -                      \\ \midrule
\multirow{3}{*}{Flamingo \cite{alayrac2022flamingo}} & \multirow{3}{*}{2022} & Chinchilla(70B)          &         &         & \Checkmark          &        & \Checkmark   &                          & zero-shot    & 50.6            & 56.3                 \\
                              &                       & Chinchilla(71B)          &         &         & \Checkmark          &        & \Checkmark   &                           & few-shot(4)  & 57.4            & 63.1                 \\
                              &                       & Chinchilla(72B)          &         &         & \Checkmark          &        & \Checkmark   &                           & few-shot(32) & 57.8            & 67.6                 \\ \midrule
\multirow{2}{*}{BLIP-2 \cite{li2023blip}}       & \multirow{2}{*}{2023} & OPT(6.7B)                &         &         & \Checkmark          & \Checkmark      &     &                           & zero-shot    & 36.4            & 52.6                 \\
                              &                       & FlanT5(XXL)              &         &         & \Checkmark          & \Checkmark      &     &                           & zero-shot    & 45.9            & 65.0                 \\ \midrule
LLaVA-1.5 \cite{liu2023llava}               & 2023                  & Vicuna(7B)               &         &         & \Checkmark          & \Checkmark      &     &                           & few-shot     &   -              & 78.5                 \\ \midrule
Qwen-VL-Chat \cite{Qwen-VL}            & 2023                  & Qwen(7B)                 &         &         & \Checkmark          & \Checkmark      &     &     &                       few-shot     & 56.6            & 78.2                 \\ \midrule
mOLUG-owl2 \cite{mplug-owl2}              & 2023                  & LLaMA-2(7B)              &         &         & \Checkmark          & \Checkmark      &     &     &                       few-shot     & 57.7            & 79.4     \\ 
InternVL2 \cite{chen2024internvl}              & 2024                  & InternLM2-Chat              &         &         & \Checkmark          & \Checkmark      &     &     &                       few-shot     & -            & -     \\ \midrule[1.3pt]           
\end{tabular}}\label{tab:mm}
\end{table}

\subsection{Techniques of the MLLMs}
We summarize the recent MLLM models with their base LLM models, MLLM types, Techniques and Performances in Table \ref{tab:mm}. There are various techniques that have helped the researchers in generalizing LLM to MLLM, which equip the LLM with the image understanding capability.

\paragraph{Fine-tuning and Instruction-tuning.} 
Fine-tuning enhances MLLM's ability to interpret input images and questions \cite{wei2021finetuned,iyer2022opt,sanh2022multitask}. Typically, a VQA prompt includes: (1) task background description, (2) input context (image, image description, OCR), (3) the target question and answer candidates, and (4) reference answers \cite{guo2023images,zhu2023minigpt,chen2023x}. The prompt design varies across models. For instance, Visual ChatGPT \cite{wu2023visual} incorporates multiple visual model formats, while domain-specific VQA may include specialized background knowledge or pre-labeled visual prompts for object detection and segmentation tasks \cite{yang2023setofmark}.

Fine-tuning with specific datasets helps MLLM perform specialized tasks. For example, general MLLMs struggle with mathematical reasoning \cite{lu2023mathvista}, so tailored datasets for diagram understanding \cite{zhang2024mathverse}, geometry reasoning \cite{li2024eagle}, and handwritten equation recognition \cite{pereira2024can} have been created to improve performance. To avoid losing general capabilities, fine-tuning often involves both task-specific and general datasets.

\paragraph{In-context Learning.} 
To improve few-shot or zero-shot learning in VQA, researchers leverage in-context learning (ICL), which uses example contexts to enhance LLM performance \cite{ICL74,shen2024hugginggpt,Gupta_2023_CVPR}. Early ICL prompts were textual, selecting image-question pairs similar to the target \cite{dou2022empirical,lu2024chameleon,shao2023prompting}, while more recent approaches generate examples directly from the target image itself \cite{guo2023images}. Multimodal ICL integrates additional visual context, beyond text, to aid learning \cite{lu2022fantastically,mmicl}. Current research explores two key aspects: (1) improving example selection for better task understanding, with methods like RAG utilizing retrieval and generation to create more representative examples \cite{yang2022empirical,shao2023prompting,tsimpoukelli2021multimodal}; and (2) designing richer multimodal context schemas, such as MMICL, which generates subgraphs of target images with symbolic correspondences to textual elements \cite{lu2024chameleon,mmicl}.

\paragraph{Visual Perception Capability.} Early studies directly extend LLM to equip with visual functions and choose to call basic visual models, such as Object Detection, Image Captioning, and OCR, to provide usable visual information for plain text LLM \cite{yang2022empirical,shao2023prompting}. Further, researchers explored directly giving LLM visual understanding capabilities. By designing a visual encoder adapted to LLM, the ability to match images and texts is pre-trained in a large amount of image and text data \cite{li2023blip,li2024blip,instructblip}. The general MLLM achieves pixel-level image understanding based on image encoders such as visual Transformer. Compared with the previous MLLM based on object-level image understanding, it has better image-text alignment and fine-grained understanding capabilities \cite{liu2024visual}. Based on this training, MLLM enhances the semantic segmentation capability of images through better visual annotation \cite{wu2023visual}, and combines multiple and larger image data to train MLLM to understand more diverse images \cite{mplug-owl2}. However, since a large amount of training data comes from natural images in real scenes, MLLM exposes its defects in understanding specific images, such as understanding document images such as PDF (especially formulas in images) \cite{ding2023vqa}, understanding related text and symbols in real scenes \cite{li2023towards}, and understanding complex charts \cite{zeng2024advancing}. 
More real-scene OCR technologies are being explored.

\section{KNOWLEDGE REASONING IN VQA}\label{sec5}
\subsection{Knowldege Sources}
External knowledge is indispensable for knowledge-based VQA task whose answers cannot be readily inferred from images but requires common sense knowledge. We summarize the most widely used knowledge sources for recent knowledge-based VQA methods, which can be divided into two categories based on the data format, structured knowledge such as DBpedia \cite{auer2007dbpedia} and ConceptNet \cite{liu2004conceptnet}, and unstructured knowledge \cite{li_inner_2022,ding2022mukea}. We give detailed introduction of the knowledge sources in Appendix. C., and conclude the different knowledge source in models in Table \ref{knowledge}.

\begin{table}[]
\caption{Knowledge reasoning models}
\scalebox{0.8}{
\begin{tabular}{l|l|lll}
\midrule[1.3pt]
\multirow{2}{*}{Models}              & \multirow{2}{*}{Knowledge Source} & \multicolumn{3}{c}{Knowledge Reasoning}                                                             \\
                                     &                                   & Methods                 & \multicolumn{2}{c}{description}                                                                    \\ \midrule[1.3pt]
Explicit Knowledge-based\cite{wang2015explicit}   & DBpedia                           & Conventional             & \multicolumn{2}{l}{SPARQL queries}                                                  \\
FVQA \cite{wang2017fvqa}                                & WebChild, ConceptNet, DBpedia     & Conventional             & \multicolumn{2}{l}{KB query triplet}                                \\
Ask me anything \cite{wu2016ask}                     & DBpedia                           & One-hop                 & \multicolumn{2}{l}{SPARQL queries}                                             \\
Out of the box \cite{narasimhan2018out}                      & ConceptNet,  DBpedia              & One\&Multi-hop      & \multicolumn{2}{l}{fact scoring and source scoring, Graph-based}                               \\
Passage Retrieval \cite{qu2021passage}                   & Wikipedia passage                 & One-hop                 & \multicolumn{2}{l}{Dense Passage Retrieval}                                          \\
Transform-Retrieve-Generate \cite{gao2022transform}         & Wikipedia passage                 & One-hop                 & \multicolumn{2}{l}{Dense Passage Retrieval, generative model}                                \\
Incorporating external knowledge \cite{li2017incorporating}    & ConceptNet                        & Multi-hop               & \multicolumn{2}{l}{Memory-based: dynamic memory network}                  \\
visual knowledge memory \cite{su2018learning}    & Visual Genome                     & Multi-hop               & \multicolumn{2}{l}{Memory-based: key-value structural memory}                      \\
Inner Knowledge-Based Img2Doc\cite{li2022dynamic}  & Inner Knowledge                   & Multi-hop               & \multicolumn{2}{l}{Memory-based}                                          \\
Mucko \cite{zhu2020mucko}                               & Inner Knowledge                   & Multi-hop               & \multicolumn{2}{l}{Graph-based: factual,visual,semantic graph}                                 \\
context-aware knowledge aggr. \cite{li2020boosting}  & Wikipedia                         & Multi-hop               & \multicolumn{2}{l}{Graph-based: semantic graph}                                              \\
See is Knowing \cite{ramnath2020seeing}                      & Others                            & Multi-hop               & \multicolumn{2}{l}{Implicit Reasoning: ERMLP mode}                                             \\
MuKEA \cite{ding2022mukea}                               & Inner Knowledge VQA2.0            & Multi-hop               & \multicolumn{2}{l}{Implicit Reasoning: knowledge graph completion}       \\ \midrule[1.3pt]            
\end{tabular}}\label{knowledge}
\end{table}

\subsection{Knowledge Extraction}

\subsubsection{Entity-based Extraction}
Entity-based methods extract visual or textual concepts from image-question pairs, using them as anchors for knowledge extraction. For instance, Wu et al. \cite{wu2016ask} extract key attributes from images and generate SPARQL queries for DBpedia retrieval. Similarly, Wang et al. \cite{wang2015explicit} detect visual concepts (objects, attributes, scenes) and link them to synonymous entities in DBpedia to construct RDF graphs. To enhance extraction, Su et al. \cite{su2018learning} apply subgraph hashing to match knowledge triplets with question phrases and expand relevant connections. Recently, Li et al. \cite{li2020boosting} introduced an approach that treats visual and textual concepts as KG anchors, expanding to include first-order neighbors and scoring nodes to select relevant knowledge.

\subsubsection{Feature-based Extraction}
Feature-based methods focus on converting knowledge into continuous representations. Narasimhan et al. \cite{narasimhan2018straight} use LSTMs to extract key relations from questions and calculate affinity scores between the image-question features and knowledge base facts, selecting the most relevant knowledge. Ziaeefard et al. \cite{ziaeefard2020towards} further develop this by transforming facts into semi-phrases, which are represented using BERT. Ding et al. \cite{ding2022mukea} merge knowledge extraction with image-question feature extraction through a pre-trained visual-linguistic model. Li et al. \cite{li2022inner} view knowledge extraction as graph representation learning, linking images to captions and KGs, and using Deepwalk to generate knowledge-aware embeddings. More details about the knowledge extraction methods can be found in Appendix. D.

\subsection{Knowledge Reasoning}
Knowledge reasoning in knowledge-based VQA involves deriving answers from extracted knowledge \cite{shen2021knowledge}. 

\subsubsection{Conventional Reasoning Methods}
Conventional methods often employ rule-based or template-based approaches. Wang et al. \cite{wang2015explicit} extract visual concepts from images and parse questions using templates to generate SPARQL queries for answer reasoning. Wang et al. \cite{wang2017fvqa} take this further by using LSTMs to parse questions into KB query triplets, filtering concepts and relations from both the image and the knowledge base, and applying distinct rules based on the source of the knowledge.

\subsubsection{One-Hop Reasoning Methods}
Methods based on conventional reasoning are usually unstable under complex conditions and unscalable to different domains. One-hop reasoning methods utilize deep learning to solve these problems from first-order knowledge, where latent rules are learned for fact selection in training.

As for structured knowledge, Wu et al. \cite{wu2016ask} proposed an encoder-decoder answer generation architecture. They first generate SPARQL queries based on an image for KB searching. Then, collected knowledge comment paragraphs are combined together and sent to Doc2Vec \cite{le2014distributed} to get the knowledge feature. Finally, knowledge and image features are combined with question tokens, which are sequentially fed into an encoder-decoder LSTM framework to get the answer. Narasimhan et al. \cite{narasimhan2018out} treat answer reasoning as the combination of fact scoring and source scoring. They first extract the image feature, visual caption feature, and question feature, and utilize a Multi-Layer Perception (MLP) to fuse these features into an image-question representation $g^{MLP}_{w_{1}}(x,Q)$. Then, they leverage $g^{MLP}_{w_{1}}(x,Q)$ to score vectorized external knowledge $g^{F}(f_{i})$, and choose the most relevant fact as the candidate answer $\hat{f}$:
\begin{equation}
\begin{aligned}
    S_{w_{1}}(g^{F}(f_{i}),g^{MLP}_{w_{1}}(x,Q)) &=cos(g^{F}(f_{i}),g^{MLP}_{w_{1}}(x,Q)) \\
    &= \frac{g^{F}(f_{i})\cdot g^{MLP}_{w_{1}}(x,Q)}{\left \| g^{F}(f_{i}) \right \| \cdot  \left \|g^{MLP}_{w_{1}}(x,Q)) \right \|}.
\end{aligned}
\end{equation}

Finally, the problem $Q$ is sent into an LSTM to predict the answer source $\hat{s}=h^{s}_{w_{2}}(Q)$, where $\hat{s}\in \{Image, KB\}$. If $\hat{s}=Image$, the head entity of $\hat{f}$ is taken as the answer, and vice versa.


Qu et al. \cite{qu2021passage} take the idea of Dense Passage Retrieval (DPR) \cite{karpukhin2020dense} and propose a coarse-grained approach for outside-knowledge-based VQA, that is, finding several passages containing answers as output. They use a BERT to extract the passage representations of candidate collection before the training process and store them in the memory slot to reduce redundant computations. To further merge the visual problem information, the given image and question are simultaneously fed into a pre-trained visual language model LXMERT \cite{tan2019lxmert} to obtain a query representation. The dot products of query representation and passage representations are calculated to obtain the top-$k$ passages.

Since coarse-grained passages only provide texts that may contain an answer but not the answer itself, Gao et al. \cite{gao2022transform} take candidate passages as their external knowledge, and deploy a generative model for answer reasoning. To enhance multimodal compatibility, they first utilize the combination of caption text $C_{i}$, attribute text $L_{i}$, and optical character recognition (OCR) text $O_{i}$ to represent the given image $v_{i}=(C_{i},L_{i},O_{i})$. Then, each candidate passages $p_{i,k}$ of top-$k$ collections are concatenated with question context $Q_{i}$ and image context $v_{i}$ to a Transformer-based encoder to get $k$ hidden layer representations:
\begin{equation}
\mathbf{z}^{Q_{i}}=(\mathbf{z}^{Q_{i}}_{1},\mathbf{z}^{Q_{i}}_{2},\cdots,\mathbf{z}^{Q_{i}}_{k}),
\end{equation}
\begin{equation}
\mathbf{z}^{Q_{i}}_{k}=E_{SelfAttn}(Q_{i},v_{i},p_{i,k}).
\end{equation}

Finally, these representations are sent to a decoder to generate the answer, and an auto-regressive cross-entropy loss is used to train the entire model.
\begin{equation}
P(a_{1}),\dots ,P(a_{1})=\sigma (D_{SelfAttn}(\mathbf{z}^{Q_{i}})).
\end{equation}

\subsubsection{Multi-Hop Reasoning Methods}
One-hop reasoning methods typically extract shallow knowledge from the KB and are incapable of exploiting implicit knowledge and handling inter-fact relations. To solve these problems, multi-hop reasoning has been widely used in recent methods, which refers to performing multi-step inference on the KB to explore the logical and semantic relations between facts. There are three main branches of multi-hop reasoning: memory-based reasoning, graph-based reasoning, and implicit reasoning.

\paragraph{\textbf{Memory-based Reasoning}}
Memory-based methods treat reasoning as the process of knowledge memory. These methods iteratively memorize candidate facts to extract relevant knowledge and ignore extraneous knowledge, which brings the capacity of multi-hop reasoning.

Li et al. \cite{li2017incorporating} first filter the extracted knowledge, where facts are scored based on the knowledge graph topology and the top-$N$ facts are selected as candidates. Then, the representations of candidate facts are extracted and stored in memory slots for reading and writing. In the next process, they deploy a dynamic memory network with $T$ iterations for knowledge accumulation, which consists of an attention component and a memory updating component. The attention component assigns weights for each knowledge representation $M_{i}$:
\begin{equation}
\begin{aligned}
    \mathbf{\alpha }^{(t)} &=softmax(\mathbf{w}tanh(\mathbf{W}_{2}\mathbf{z}^{(t)}_{i}+\mathbf{b}_{2})),\\
  \ \mathbf{z}^{(t)}_{i} &=[M_{i};m^{(t-1)};\mathbf{q}], \\
  \ \mathbf{q} &=tanh(\mathbf{W}_{1}[\mathbf{f}^{(I)};\mathbf{f}^{(Q)};\mathbf{f}^{(A)}]+\mathbf{b}_{1}),
\end{aligned}
\end{equation}
in which $\mathbf{f}^{(I)}, \mathbf{f}^{(Q)}, \mathbf{f}^{(A)}$ are features of images, questions, and multi-choice answers respectively. The memory updating component accumulates knowledge based on attention weights to update the memory vector $m^{(t)}$:
\begin{equation}
\begin{aligned}
     m^{(t)} &=RELU(\mathbf{W}_{3}[m^{(t-1)};\mathbf{c}^{(t)};\mathbf{q}]+\mathbf{b}_{3}), \\ 
     \ \mathbf{c}^{(t)} &=\sum_{i=1}^{N}\mathbf{\alpha }^{(t)}M_{i}, t=1,\dots ,T.
\end{aligned}
\end{equation}

Finally, $\mathbf{f}^{(I)}, \mathbf{f}^{(Q)}, \mathbf{f}^{(A)}$ and $m^{(T)}$ are fused together to obtain the confidence score for each candidate answer.

Instead of directly considering each fact as an entirety, Su et al. \cite{su2018learning} use key-value structural memory slots. They first decompose each knowledge triplet $(s,r,t)$ into three key-value pairs (i.e., $(s,r)$-$t$, $(s,t)$-$r$, $(r,t)$-$s$), which are passed to the joint embedding module to get the key representation $\mathbf{k}_{i}$ and value representation $\mathbf{v}_{i}$:
\begin{equation}
    \mathbf{k}_{i}=\Psi (e_{1},u_{i})+\Psi (e_{2},u_{i}),\ \mathbf{v}_{i}=\Psi (e_{3},u_{i}),
\end{equation}
where $e_{1},e_{2},e_{3}$ are different entries of $i$-th triplet, and $u_{i}$ is the image feature. Then, a memory network iteratively refines the memory vector by performing key addressing and value addressing to obtain the answer. Li et al. \cite{li2022dynamic} store all entries of each knowledge triplet in a value slot (i.e., $[F_{s},F_{r},F_{t}]$) and take their average representation as the key embedding. Then, a memory reading module captures the correlation between query embedding $\hat{q}$ and key-value pairs ($\hat{k}_{i}$-$\hat{v}_{i}$) and obtains the question-aware knowledge representation $m^{t}$ which is further used to guide the graph learning:
\begin{equation}
\begin{aligned}
    m^{t} &=\sum_{i=1}^{N}p_{i}\hat{v}_{i},\\
    p_{i} &=softmax(\hat{q}\cdot \hat{k}^{T}_{i}),\\ 
    \hat{v}_{i} &= \sum_{\hat{v}_{ij}\in[F_{s},F_{r},F_{t}]} (1-softmax(\hat{q}\cdot \hat{v}_{ij}^{T}))\hat{v}_{ij}/2.
\end{aligned}
\end{equation}

\paragraph{\textbf{Graph-based Reasoning}}
Graph-based methods tend to represent extracted facts as graphs, and then use the message passing paradigm to aggregate knowledge from multi-hop neighborhoods to the target nodes. This process enables the model to explicitly exploit the attribute information and relational information embedded in the KB, which is the most commonly used multi-hop reasoning method for the knowledge-based VQA task.

Narasimhan et al. \cite{narasimhan2018out} propose a Graph Convolutional Network (GCN) based model for graph-based reasoning, which mainly consists of two components: the factual graph construction and answer scoring. For the factual graph construction, facts fetched from the KB are composed into a homogeneous graph, where each node represents an entity, and each edge represents a relation. For incorporating images and questions information, the representation of each node is formed by the concatenation of image feature $g_{w}^{V}(I)$, question feature $g_{w}^{Q}(Q)$ and entity feature $g_{w}^{C}(e)$:
\begin{equation}
    H_{i}^{0}=(g_{w}^{V}(I);g_{w}^{Q}(Q);g_{w}^{C}(e)),\ e_{i}\in E,
\end{equation}
where $E$ is the entity set. In the process of answer scoring, a GCN integrates node features based on graph topology and outputs the $L$-th layer features $\hat{g}(e_{i})=H_{i}^{L}$ which are fed into an MLP to predict answer $\hat{A}$:
\begin{equation}
    \hat{A}=\mathrm{arg} \max_{e_{i}\in E}MLP(\hat{g}(e_{i})).
\end{equation}

In addition to using graph structure to represent the structural relationships between facts, Zhu et al. \cite{zhu2020mucko} further introduce visual graphs and semantic graphs to comprehensively depict image information. The visual graph is a scene graph, which is composed of objects in the given image and their positional relations. As for the semantic graph, an image is first sent to the DenseCap \cite{johnson2016densecap} to generate several captions, which are parsed into a semantic graph. The knowledge reasoning of this model comprises two parts: intra-modal knowledge selection and inter-modal knowledge reasoning. In the first stage, the visual, semantic, and factual graphs are aggregated separately using different GCNs to obtain the updated node features: $\{\hat{v}_{i}^{V}\}_{i=1}^{N^{V}}, \{\hat{v}_{i}^{S}\}_{i=1}^{N^{S}}, \{\hat{v}_{i}^{F}\}_{i=1}^{N^{F}}$ respectively. In the second stage, the information on the visual and semantic graphs is mapped to the factual graph (i.e., visual-to-factual and semantic-to-factual). Taking the semantic-to-factual process as an example: 
\begin{equation}
\begin{aligned}
        m_{i}^{S\rightarrow F} &=\sum_{j\in N^{V}} \Upsilon _{ji}^{S\rightarrow F}\hat{v}_{j}^{S},\\ 
  \Upsilon _{ji}^{S\rightarrow F} &=softmax(w_{c}tanh(W_{8}\hat{v}_{j}^{S}+W_{9}[\hat{v}_{i}^{F},q])),
\end{aligned}
\end{equation}
in which $\Upsilon _{ji}^{S\rightarrow F}$ is the attention vector between $\hat{v}_{i}^{F}$ and each node of semantic graph, and $m_{i}^{S\rightarrow F}$ is the weighted sum of nodes on semantic graph for $\hat{v}_{i}^{F}$. Finally, $m_{i}^{S\rightarrow F}, m_{i}^{V\rightarrow F}, \hat{v}_{i}^{F}$ are fed into an element-wise gate network to get the final representation $\tilde{v}_{i}^{F}$:
\begin{equation}
\begin{aligned}
    \tilde{v}_{i}^{F} &=W_{11}(gate_{i}\circ [m_{i}^{S\rightarrow F}; m_{i}^{V\rightarrow F}; \hat{v}_{i}^{F}]),\\
    gate_{i} &=\sigma (W_{10}[m_{i}^{S\rightarrow F}; m_{i}^{V\rightarrow F}; \hat{v}_{i}^{F}]).
\end{aligned}
\end{equation}

Li et al. \cite{li2020boosting} treat knowledge reasoning as a process of anchor entity feature learning, where anchor entities could continuously acquire knowledge from the KB and facilitate answer prediction. They first use an object detector and a natural language parsing tool to extract entities from images, questions and candidate answers as anchor entities. In the second process, the first-order neighbors of anchor entities are obtained from the KB to form a global knowledge graph, which is fed into the attention-based GNN for feature aggregation. Then, extracted knowledge is distilled into three auxiliary features $\vec{e}^{(ctx)},\vec{u},\vec{e}^{(ans)}$:
\begin{equation}
  \vec{e}^{(ctx)}=\sum_{e_{i}\in \mathcal{C}}\alpha_{i}\vec{e}_{i},\ \vec{u}=RELU(\beta \cdot W_{3}),\ \vec{e}^{(ans)}=\sum_{e_{j}\in \mathcal{A}} \beta_{j}\vec{e}_{j},
\end{equation}
\begin{equation}
    \textrm{where}\ \alpha_{i}\propto exp(h^{q} \cdot \vec{e}_{i}),\ \beta_{j}\propto exp(\vec{e}^{(ctx)} \cdot \vec{e}_{j}),
\end{equation}
in which $h^{q}$ is the query embedding, $\mathcal{C}$ is the anchor entity set extracted from the question and image, and $\mathcal{A}$ is the anchor entity set extracted from candidate answers. Finally, auxiliary features are fused with image feature $\tilde{v}_{k}$ and $h^{q}$:
\begin{equation}
\vec{f}=BaseFusion(\{\tilde{v}_{k}\};\{\vec{e}^{(ctx)},\vec{e}^{(ans)},\vec{u}\}\cup \{h_{m}^{q}\}).
\end{equation}

\paragraph{\textbf{Implicit Reasoning}}
Different from memory-based or graph-based methods, implicit reasoning treats the multi-hop reasoning task as an entity feature space learning issue, which aims to map the head, relation, and tail entities into a common feature space such that they can establish some statistical associations.

Ramnath et al. \cite{ramnath2020seeing} proposed the ``See is Knowing'' framework, where the visual information is represented as several knowledge vectors to facilitate answer prediction. They first deploy an ERMLP model \cite{dong2014knowledge} on the large-scale KG, which captures the intrinsic connections between entities by learning to identify whether given facts exist. ERMLP can implicitly perform multi-hop reasoning in learning and generate a dense embedding for each entity. Then, scene, object, and action concepts are detected in given images and passed to ERMLP to get their knowledge-aware embedding $e_{i}^{j}, j\in[1,m]$. Finally, these visual knowledge embeddings and the query embedding $A(q_{i})$ are passed into an attention module:
\begin{equation}
 A(I_{i})=\sum^{m}_{j=1} \alpha_{I}^{j}e_{i}^{j}, \alpha_{I}^{j}=\frac{exp(w^{T}_{\alpha_{I}}[A(q_{i});e_{i}^{j}])}{\sum_{k=1}^{m}exp(w^{T}_{\alpha_{I}}[A(q_{i});e_{i}^{k}])},
\end{equation}
in which $A(I_{i}), A(q_{i})$ are further used to query the answer from the KG.

As facts on the KG are typically rigid and incapable of complex scene understanding, Ding et al. \cite{ding2022mukea} propose MuKEA, which extracts and accumulates complex knowledge from VQA scenarios directly. They consider the model learning process as a multimodal knowledge graph completion problem, i.e., extracting multimodal knowledge triplets from image-question pairs and training the model taking the idea of TransE \cite{bordes2013translating}. During the head entity extraction, a pre-trained visual-language model LXMERT \cite{tan2019lxmert} is used to obtain the image embeddings and question embeddings jointly. 
These embeddings are fed into a hard attention mechanism, which captures the correlations between each object and question token to get the head entity. During tail entity extraction, the answer representation is used as the tail embedding. 
This implicit reasoning allows the model to continuously acquire multimodal knowledge from the VQA dataset and fully exploit potential clues between facts.

\subsection{MULTIMODAL REASONING WITH LLMs}
Recent years have witnessed the remarkable progress of MLLMs, especially in their surprising zero/few-shot reasoning abilities. In varieties of multimodal reasoning tasks (e.g., VQA), MLLMs often demonstrate impressive effectiveness. Therefore, MLLMs and VQA task run towards each other at the same time, forging a new direction for VQA research. Specifically, MLLMs are capable to make great comprehension and integration of information from image and textual questions after pretraining on large-scale multimodal data, enabling them to reason carefully and generate appropriate answers. Furthermore, several strategies have recently emerged to enhance the reasoning capabilities of MLLMs and improve the VQA performance, such as multimodal instruction tuning~\cite{instructblip}, multimodal in-context learning~\cite{alayrac2022flamingo}, multimodal chain-of-thought~\cite{lu2024chameleon}, and LLM-aided visual reasoning \cite{shen2024hugginggpt}.

\subsubsection{Multimodal Instruction Tuning}
Pretrained MLLMs often struggle with generalizing to novel tasks and aligning with users’ intentions, resulting in incorrect and dissatisfactory responses. To address these limitations, a strategy known as multimodal instruction tuning (M-IT) \cite{sanh2022multitask} has been introduced. Instruction refers to the task description. Multimodal instruction tuning is a technique that involves finetuning pretrained MLLMs on a collection of multimodal instruction-following data. Tuning in this way, MLLMs are guided to understand and adapt to the task of interest, thus boosting their zero-shot reasoning capabilities and task-specific performance. The success of some notable frameworks on VQA (e.g., BLIP-2~\cite{li2023blip}) validates the effectiveness of this idea. 
\subsubsection{Multimodal In-Context Learning}
As the demand for customized MLLMs for specific VQA task continues to grow, finetuning them by instruction tuning proves to be resource-intensive and may diminish the model’s generalization capabilities. Furthermore, state-of-the-art MLLMs like GPT-4V are primarily accessible only through API calls, with their parametric weights remaining proprietary and unavailable to the public. This scenario underscores the growing need for a new methodology, multimodal in-context learning (M-ICL), which allow learning from analogy~\cite{surveyicl} without requiring parametric updates. Specifically in M-ICL, MLLMs learn from a few examples noted as demonstration. The examples not only provide supplementary contextual knowledge for MLLMs but also exert flexible control over output, thereby solving complex and unseen tasks in a few-shot manner~\cite{mmreact}. Researches in M-ICL~\cite{alayrac2022flamingo,DBLP:conf/aaai/YangGW0L0W22} are shown empirically to enhance the reasoning ability of MLLMs on VQA tasks.

\subsubsection{Multimodal Chain-of-Thought}
In recent studies, chain-of-thought has gained widespread usage in eliciting the multi-step reasoning abilities of LLMs. Specifically, CoT aims to enable LLMs to imitate the step-by-step thinking process of humans. It encourages the LLMs to generate not only the final answer but also the intermediate reasoning chains that lead to the answer by adding a prompt like ``Let's think step by step''. Subsequently, to extend CoT reasoning to multimodality, several works~\cite{DBLP:journals/corr/abs-2305-02317,DBLP:conf/emnlp/HimakunthalaORH23} have been proposed to extend the unimodal CoT to Multimodal CoT (M-CoT). Given the inputs in different modalities, Multimodal CoT decomposes multi-step problems into intermediate reasoning steps (rationale) and then infers the answer. The success of multiple researches~\cite{DBLP:journals/corr/abs-2302-00923,scienceqa} validates the effectiveness of M-CoT in enhancing MLLMs reasoning ability in VQA tasks. 
\subsubsection{LLM-Aided Visual Reasoning}
Building on the achievements of tool-augmented LLMs~\cite{talm}, researchers have explored the potential of invoking external tools and modular approaches for visual reasoning tasks like VQA. Specifically, vision foundation models~\cite{visualgpt} like image captioning and optical character recognition are usually invoked for better visual information understanding. Invoking external tools~\cite{lu2024chameleon} such as knowledge retrieval and web search engines help LLMs access real-time information and leverage domain-specific knowledge from external resources. Multiple researches~\cite{shen2024hugginggpt,yang2024gpt4tools} indicate that effective utilization of external tools enables LLMs to accommodate various reasoning capabilities for accomplishing complex VQA tasks. 

\subsubsection{Math and Logical Reasoning}
MLLMs are increasingly being applied to mathematical and logical reasoning tasks. The core challenge is that solving math and logic problems requires not only linguistic understanding but also a deeper ability to perform structured reasoning, abstract thinking, and accurate computation, all of which pose limitations for traditional language models \cite{lu2023mathvista,xiao2024logicvista}.

To address this, various methods have been explored. One prevalent approach is symbolic manipulation, where MLLMs are trained to recognize symbolic representations of mathematical expressions (e.g., formulas, diagrams) and perform algebraic transformations \cite{wu2024symbol,li2024mllm}. These models integrate visual features from images or written equations with text-based inputs, allowing for more nuanced reasoning in both spatial and symbolic contexts. There are some attempts that try to improve the visual ability for math diagrams or logical charts \cite{zhang2024mathverse}, to explore the potential math and logical reasoning performance. Another key method is chain of thought reasoning \cite{DBLP:journals/corr/abs-2405-06705,DBLP:conf/jsai/LeNDNN24}. In this approach, models learn to extract symbolic rules from data and apply them within a structured reasoning framework. This enables them to handle tasks that require step-by-step deduction, such as proofs or multi-step logical arguments. Recent advancements also involve the multi-agent systems to integrate external symbolic solvers, where MLLMs collaborate with dedicated math engines (e.g., Wolfram Alpha) to perform complex calculations or proofs, improving accuracy for higher-level reasoning tasks.

\section{DATASETS AND METRICS}\label{sec6}
\subsection{Datasets}
\begin{figure}
    \centering
    \includegraphics[width=0.85\linewidth]{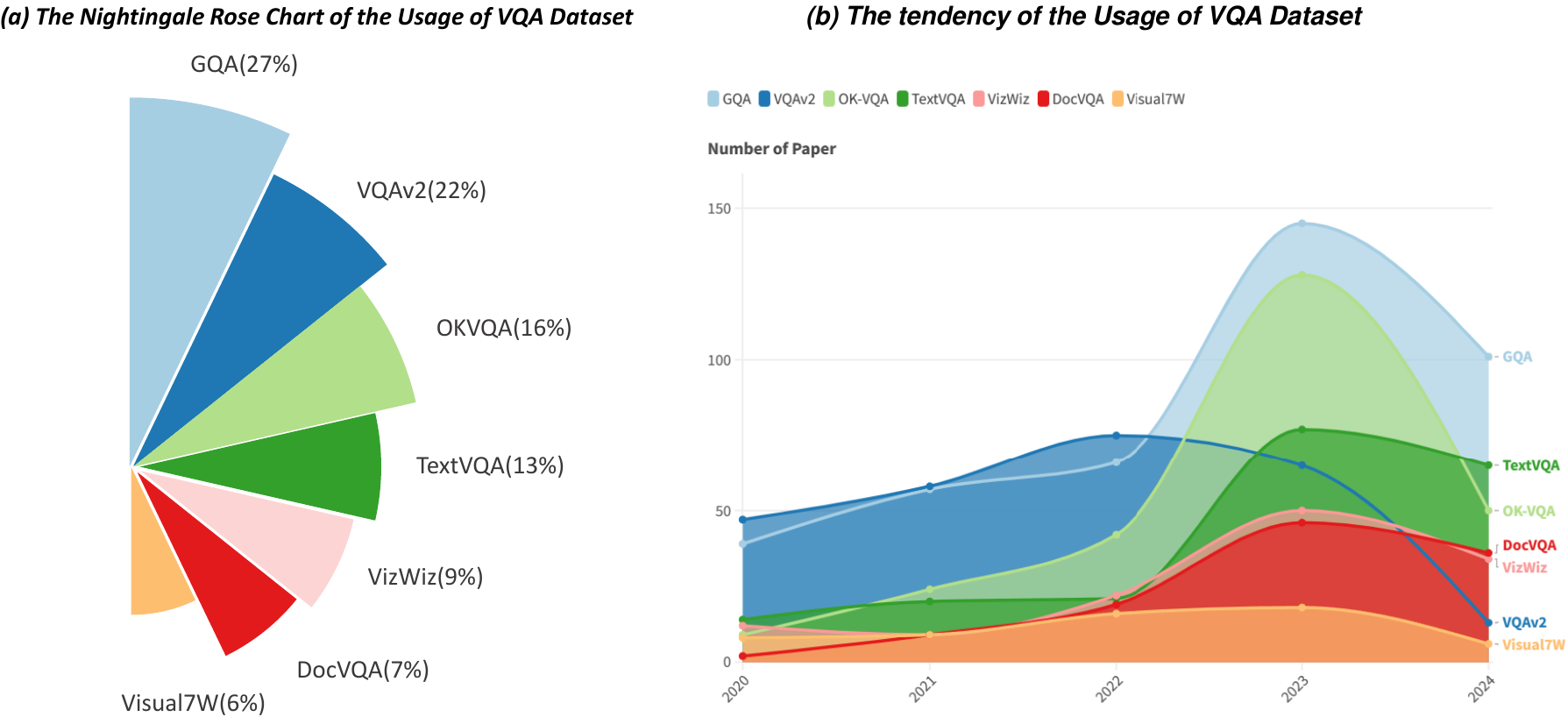}
    \caption{The statistics of the widely-used dataset from 2020 to 2024.}
    \label{fig:dataset-statis}
\end{figure}

In this part, we systematically summarize the most widely used VQA datasets that are divided into two categories: (1) datasets whose questions are typically based on common sense knowledge (Sec \ref{sec6.1.1}). (2) datasets based on external knowledge (Sec \ref{sec6.1.2}). The statistics of all datasets are listed in Table \ref{table4}, and we conclude the seven most widely-used dataset in Fig. \ref{fig:dataset-statis}. Some representative datasets are selected to be presented here, and the rest are shown in Appendix. E.

\begin{table*}[]
\centering
\caption{Datasets for VQA and their main characteristics. Avg.Q/Image indicates how many questions a image meanly corresponds to, and Avg.Q length and Avg.A length denotes the avarage length of question and answer respectively. }\label{table4}
\resizebox{\linewidth}{!}{
\begin{tabular}{l|ccc|ccccc|c}
\midrule[1.3pt]
\textbf{Dataset} & \textbf{\begin{tabular}[c]{@{}c@{}}Knowledge\\ based?\end{tabular}} & \textbf{\begin{tabular}[c]{@{}c@{}}Published\\ year\end{tabular}} & \textbf{Image source} & \textbf{Images} & \textbf{Q\&A pairs} & \textbf{\begin{tabular}[c]{@{}c@{}}Avg.Q/\\ Imgae\end{tabular}} & \textbf{\begin{tabular}[c]{@{}c@{}}Avg.Q\\ length\end{tabular}} & \textbf{\begin{tabular}[c]{@{}c@{}}Avg.A\\ length\end{tabular}} & \textbf{\begin{tabular}[c]{@{}c@{}}Questions \\ Generation\end{tabular}} \\ \midrule
DAQUAR \cite{malinowski2014multi}           & \usym{2717}                                                                   & 2015                                                              & NYU-Depth V2          & 1,449           & 12,468              & 8.6                                                             & 11.5                                                            & 1.2                                                             & Human                                                                    \\
COCO-QA \cite{ren2015image}         & \usym{2717}                                                                   & 2015                                                              & COCO                  & 117,684         & 117,684             & 1.0                                                             & 8.6                                                             & 1.0                                                             & Automatic                                                                \\
Visual Madlibs \cite{yu2015visual}  & \usym{2717}                                                                    & 2015                                                              & COCO                  & 10,738          & 360,001             & 33.5                                                            & 4.9                                                             & 2.8                                                             & Human                                                                    \\
FM-IQA \cite{gao2015you}           & \usym{2717}                                                                    & 2015                                                              & COCO                  & 158,392         & 316,193             & 2.0                                                             & 7.4                                                             & 3.8                                                             & Human                                                                    \\
VQAv1 \cite{antol2015vqa}           & \usym{2717}                                                                    & 2015                                                              & COCO                  & 204,721         & 614,163             & 3.0                                                             & 6.2                                                             & 1.1                                                             & Human                                                                    \\
Visual Genome \cite{krishna2017visual}    & \usym{2717}                                                                    & 2016                                                              & COCO \& YFCC100M      & 108,077         & 1,445,322           & 13.4                                                            & 5.7                                                             & 1.8                                                             & Human                                                                    \\
Visual7W \cite{krishna2017visual}        & \usym{2717}                                                                    & 2016                                                              & Visual Genome         & 47,300          & 327,939             & 6.9                                                             & 6.9                                                             & 2.0                                                             & Human                                                                    \\
VQAv2 \cite{goyal2017making}           & \usym{2717}                                                                    & 2017                                                              & COCO                  & 204,721         & 1,105,904           & 5.4                                                             & 6.1                                                             & 1.2                                                             & Human                                                                    \\
CLEVR \cite{johnson2017clevr}           & \usym{2717}                                                                    & 2017                                                              & Synthetic             & 100,000         & 999,968             & 10.0                                                            & 18.4                                                            & 1.0                                                             & Synthetic                                                                \\
CLEVR-CoGenT-A \cite{johnson2017clevr}  & \usym{2717}                                                                    & 2017                                                              & Synthetic             & 100,000         & 999,951             & 10.0                                                            & -                                                               & -                                                               & Synthetic                                                                \\
CLEVR-CoGenT-B \cite{johnson2017clevr}  & \usym{2717}                                                                    & 2017                                                              & Synthetic             & 30,000          & 299,972             & 10.0                                                            & -                                                               & -                                                               & Synthetic                                                                \\
VQA-CPv1 \cite{agrawal2018don}        & \usym{2717}                                                                    & 2018                                                              & COCO                  & 205,000         & 370,000             & 1.8                                                             & -                                                               & -                                                               & Human                                                                    \\
VQA-CPv2 \cite{agrawal2018don}        & \usym{2717}                                                                    & 2018                                                              & COCO                  & 219,000         & 658,000             & 3.0                                                             & -                                                               & -                                                               & Human                                                                    \\
VizWiz     & \usym{2717} &  2018   &  blind people by phone      &  72,205  &  72,205 &  1.0  & -  & -  &  Human  \\
VQA-Rephrasings \cite{shah2019cycle} & \usym{2717}                                                                    & 2019                                                              & VQAv2                 & 40,504          & 162,016             & 4.0                                                             & -                                                               & -                                                               & Human                                                                    \\
GQA \cite{hudson2019gqa}             & \usym{2717}                                                                    & 2019                                                              & COCO \& Flickr        & 113,018         & 22,669,678          & 200.6                                                           & -                                                               & -                                                               & Synthetic                                                                \\
DocVQA \cite{mathew2021docvqa}          & \usym{2717}                                                                    & 2021                                                              & UCSF Library          & 12,767          & 50,000              & 3.9                                                             & 9.5                                                             & 2.4                                                             & Human                                                                    \\
InfographicVQA \cite{mathew2022infographicvqa}  & \usym{2717}                                                                    & 2021                                                              & Internet              & 5,485           & 30,035              & 5.5                                                             & 11.5                                                            & 1.6                                                             & Human                                                                    \\
IconQA \cite{lu2021iconqa}          & \usym{2717}                                                                    & 2022                                                              & digital textbooks     & 96,817          & 107,439             & 1.1                                                             & 8.4                                                             & -                                                               & Human                                                                   
\\ 
PDFVQA \cite{ding2023vqa}          & \usym{2717}                                                                    & 2023                                                             &  PubMed PDF Doc     & 25,147          & 130,700             &  5.2                                                            & -                                                            & -                                                               & Automatic                                                                   
\\
E-VQA \cite{yang2023event}          & \usym{2717}                                                                    & 2023                                                             &  News article     & 2,690          & 9,088             &  3.4                                                            & -                                                            & -                                                               & Automatic                                                                   
\\

\midrule
KB-VQA \cite{wang2015explicit}          & \Checkmark                                                                   & 2015                                                              & COCO \& ImageNet      & 700             & 2,402               & 3.4                                                             & 6.8                                                             & 2.0                                                             & Human                                                                    \\
FVQA \cite{wang2017fvqa}            & \Checkmark                                                                   & 2017                                                              & COCO                  & 2,190           & 5,826               & 2.7                                                             & 9.5                                                             & 1.2                                                             & Human                                                                    \\
R-VQA \cite{lu2018r}           & \Checkmark                                                                   & 2018                                                              & Visual Genome         & 335,000         & 4,335,966           & 13.4                                                            & -                                                               & -                                                               & Human                                                                    \\
KVQA \cite{shah2019kvqa}            & \Checkmark                                                                   & 2019                                                              & Wikidata              & 24,602          & 183,007             & 7.4                                                             & 10.1                                                            & 1.6                                                             & Human                                                                    \\
OK-VQA \cite{marino2019ok}          & \Checkmark                                                                   & 2019                                                              & COCO                  & 14,031          & 14,055              & 1.0                                                             & 8.1                                                             & 1.3                                                             & Human                                                                    \\
ViQuAE \cite{lerner2022viquae}          & \Checkmark                                                                   & 2022                                                              & Wikidata              & 3,300           & 3,700               & 1.1                                                             & 12.4                                                            & -                                                               & Automatic                                                                \\
KRVQA \cite{cao2021knowledge}           & \Checkmark                                                                   & 2022                                                              & Visual Genome         & 32,910          & 157,201             & 4.8                                                             & 11.7                                                            & -                                                               & Automatic               \\
Lora \cite{gao2024lora}          & \Checkmark                                                                    & 2023                                                             & food-and-kitchen     & 100,000          & 200,000             & 2                                                             & -                                                            & -                                                               & Automatic                                                                   
\\ \midrule[1.3pt]
\end{tabular}
}
\end{table*}
\subsubsection{Datasets without External Knowledge}\label{sec6.1.1} 
\paragraph{\textbf{DAQUAR} \cite{malinowski2014multi}}
DAQUAR is the first proposed VQA challenge. It is built on the NYU-Depth v2 \cite{silberman2012indoor} dataset, which contains 1,449 images and 12,468 Q\&A pairs. Its annotation generation methods include synthetic and human. The synthetic annotation uses eight predefined templates and the original annotation of NYU-Depth v2. Human annotations come from 5 in-house participants. Although the proposal of DAQUAR is important to VQA, it also has some problems. For instance, the magnitude of images is too small; the image quality is poor and disorganized; the dataset is unbalanced; there are too many single-choice questions.
\paragraph{\textbf{VQAv1} \cite{antol2015vqa}}
VQAv1 is one of the most widely used datasets, which contains 204,721 real images from the COCO dataset (123,287 images for training and 81,434 images for testing). It covers 614,163 free-form questions and 7,984,119 answers, allowing yes/no, multiple-choice, and open-ended forms of questions. These questions are collected by humans, and each question is manually annotated by 10 different people. The annotations also include the answers given by humans without looking at the images.
\paragraph{\textbf{VQAv2} \cite{goyal2017making}}
VQAv2 is the enhanced version of the VQAv1 dataset, which contains 204,721 images sourced from the COCO dataset. It has 443,757, 214,354, and 447,793 question annotations on the training set, validation set, and test set, respectively. VQAv2 has a total of 1,105,904 free-form Q\&A pairs annotated by humans, twice as many as VQAv1, and provides a complementary image for each question so that the same question can be combined with two similar images to generate different answers. Compared with VQAv1, VQAv2 reduces the bias and imbalance of the dataset through the above improvements.
\paragraph{\textbf{CLEVR} \cite{johnson2017clevr}}
CLEVR is a synthetic dataset and contains 100,000 rendered images and about 1M synthetic Q\&A pairs where 853,000 questions are totally different. To make the task more challenging, questions are divided into five categories: querying attributes, comparing attributes, existence, counting, and integer comparison and each image is represented as a visual scene composed of simple geometric bodies, where a VQA model needs to handle novel combinations of unseen attributes during training and goes through a long reasoning process to answer the question.
\paragraph{\textbf{IconQA} \cite{lu2021iconqa}}
IconQA consists of 96,817 Icon images and 107,439 Q\&A pairs. These Icon images come from IXL Math Learning, an open-source mathematics textbook on the Internet, and Q\&A pairs are obtained by manual collection and filtering. The questions of this dataset are mainly divided into three subtasks: 57,672 multi-image-choice, 31,578 multi-text-choice, and 18,189 filling-in-the-blank. The questions of IconQA are derived from real-world mathematical questions, which require commonsense reasoning and arithmetic reasoning.

\subsubsection{Datasets with External Knowledge Base}\label{sec6.1.2} 
\paragraph{\textbf{KB-VQA} \cite{wang2015explicit}}
KB-VQA is the first VQA dataset requiring an external KB, which includes 700 images from the COCO dataset and 2,402 Q\&A pairs. KB-VQA has 23 templates for questions, and each question is proposed by five workers according to one of the appropriate templates. The proposers assign different labels to questions of different knowledge levels. Answering questions at the ``KB-knowledge'' level requires the use of a KB like DBpedia. The ``KB-knowledge'' level questions in KB-VQA are far more than that of other contemporaneous VQA datasets.
\paragraph{\textbf{FVQA} \cite{wang2017fvqa}}
FVQA has 2,190 images and 5,826 questions which are split into five train/test sets (1,100/1,090 images and 2,927/2,899 questions for training/testing per set). The questions can be divided into 32 categories in total. Its annotations include not only Q\&A pairs, but also extra knowledge. FVQA builds a KB by collecting knowledge triples from WebChild, ConceptNet, and DBpedia, which contains 193,449 sentences as supporting facts related to 580 visual concepts (234 objects, 205 scenes, and 141 attributes). This dataset contains a supporting fact in each Q\&A pair.
\paragraph{\textbf{OK-VQA} \cite{marino2019ok}}
The Outside Knowledge-VQA (OK-VQA) dataset consists of 14,055 questions (including 12,951 unique questions) and 14,031 real images from the COCO dataset. The labeling process of OK-VQA is divided into two steps: first, workers are asked to provide questions that require external knowledge to answer for a given image, and then five different workers are asked to label answers for each image-text pair. After the annotation is completed, further filtering is required. If the answer to a question have more than five Q\&A instances, the question will be deleted, thereby ensuring an even distribution of answers and eliminating potential bias. 

\subsection{VQA Datasets with MLLM Benchmark}


With the rapid advancement of multimodal large language models (MLLMs), increasingly sophisticated general-purpose benchmarks have emerged to evaluate their performance across various dimensions. These benchmarks are designed to assess a broad range of capabilities, often with a foundation in visual question answering annotation. The creation of such evaluation data typically involves filtering image data from extensive sources, followed by generating corresponding question-and-answer (QA) pairs, either through automated processes or manual annotation. Although these benchmarks are not always explicitly developed to evaluate methodologies specific to the VQA task, they can nonetheless be regarded as generalized VQA datasets due to their shared emphasis on image-based question answering.

One of the earliest unified MLLM benchmarks, MME \cite{fu2023mme}, compiles a substantial collection of images and generates corresponding QA pairs to evaluate MLLM performance, emphasizing consistency and objectivity. Similar efforts have been observed in subsequent benchmarks, such as SEEDBENCH \cite{li2023seed} and SEEDBENCH-2 \cite{li2023seed-2}, both of which aim to standardize MLLM evaluation. However, more recent benchmarks have shifted focus towards assessing specific capabilities, expanding the scope of evaluation to include more specialized dimensions. For example, recent benchmarks evaluate models on their ability to comprehend visual information \cite{fu2023challenger,li2024seed,cai2023benchlmm}, demonstrate reasoning skills \cite{zhang2024benchmarking,roberts2023charting}, and perform in-context learning \cite{shukor2023beyond,liu2023hallusionbench}. In addition, some benchmarks address challenges such as hallucination detection and mitigation \cite{cui2023holistic,liu2023hallusionbench}, where models are tested for their ability to provide accurate information without generating false or misleading content.

Moreover, recent developments in MLLM benchmarks have extended their evaluation scope beyond general comprehension tasks, venturing into highly specialized domains. These include mathematics, physics, music, medicine, and more, as seen in datasets such as MathVista \cite{lu2023mathvista}, MMMU \cite{yue2023mmmu}, and CMMMU \cite{zhang2024cmmmu}. These domain-specific benchmarks highlight the growing recognition that MLLMs must be versatile, capable of handling complex, domain-oriented tasks that require a deep understanding of specialized knowledge. This multidimensional evaluation framework ensures that MLLMs are not only assessed for their general performance but also their abilities to excel in high-stakes areas.

\subsection{Evaluation Metrics}
Common VQA evaluation metrics can be divided into two categories: objective evaluation metrics and subjective evaluation metrics, which are often used for two mainstream VQA tasks: open-ended and multiple-choice.

\subsubsection{Subjective Evaluation Metrics} 

The most common human evaluation is to ask human judges to directly evaluate the quality of the generated answers, either from an overall perspective or from a specific dimension. These specific dimensions need to be able to reflect the interrelated properties of the answer sentences. For example, Bai et al. \cite{bai2021infobox} evaluate from three dimensions: grammar, faithfulness, and coherence. If evaluating in terms of the entire answer, human judges are usually asked to select one of three levels of fine-grained evaluation as a score, including 0 (completely false), 1 (partially true), and 2 (exactly true).

On the FM-IQA dataset, Gao et al. \cite{gao2015you} propose a Visual Turing Test (VTT) based on human evaluation. In this test, a lot of sets of images and questions, along with their corresponding human-annotated answers or answers generated by a VQA model, are presented to human judges. Human judges need to discriminate whether the answers are given by humans or computers based on the given materials. If an answer of the VQA system is judged to be that of a human, the answer passes the VTT. Finally, the percentage of answers that pass the VTT is counted, and several additional VTTs are set to calculate the standard deviation.


\subsubsection{Objective Evaluation Metrics} 
\paragraph{\textbf{QA Evaluation Metrics}}
Simple Accuracy \cite{zou2020survey} based on string matching is first proposed. For the multiple-choice VQA task, the comparison between the predicted answer and the ground truth is straightforward. However, as generated answers are mostly phrases consisting of multiple words in the open-ended VQA task, Simple Accuracy is often difficult to evaluate. On the one hand, indiscriminate judgment is clearly flawed, because the severity of wrong answers varies. For example, compared with the completely irrelevant answers obtained from prediction, the severity of answers that contain temporal errors is significantly lower, where the punishment should be different for these situations. On the other hand, there may be multiple matching answers to the same question, while their appropriateness may vary. For instance, the correct answer to ``What's swimming in the water?'' is ``bluefish'', while ``scad'' means the same as ``bluefish'', and ``fish'' is also appropriate.

S. Antol et al. \cite{antol2015vqa} propose $Accuracy_{VQA}$, which is used for open-ended evaluation on the VQAv1 dataset:
\begin{equation}
Accuracy_{VQA} = \min (\frac{n}{3},1),
\end{equation}
where $n$ is the number of predicted answers that are the same as those given by annotators. This means that the answer predicted by the algorithm with the same answer as three or more annotations is 100\% accurate. However, this evaluation standard is also unreasonable. For example, the annotators of the COCO-VQA dataset only have consensus on a few questions, which limits the highest accuracy of the model.

In addition, $Accuracy_{VQA}$ is prone to errors on ``yes/no'' type questions. In this type of question, the answer ``yes'' or ``no'' may repeat more than three times in one question, which would result in a high score of both ``yes'' and ``no'' on the $Accuracy_{VQA}$. Other Metrics are shown in Appendix. F.

\paragraph{\textbf{Generation Evaluation Metrics}}
Some of the metrics originally used to evaluate answer generation can also be used in the open-ended VQA task. The mainstream evaluation metrics among them typically include Bilingual evaluation understudy (BLEU) \cite{papineni2002bleu}, Recall-Oriented Understudy for Gisting Evaluation (ROUGE) \cite{lin2004rouge} and Metric for Evaluation of Translation with Explicit Ordering (METEOR) \cite{banerjee2005meteor}. The effectiveness of using these generation metrics for VQA system evaluation has been confirmed in \cite{gurari2018vizwiz,abacha2019vqa}. We give a breif introduction of BLEU metric, and the ROUGE and METEOR are introduced in Appendix. F.

BLEU measures the quality of the answer by comparing the coincidence degree of n-gram phrases of different lengths in the predicted answer and the real answer. The higher the coincidence degree is, the higher the quality of the answer is. 
BLEU calculates the coincidence accuracy of the corresponding answer according to the following formula:
\begin{equation}
C{P_n}(C,S) = \frac{{\sum\nolimits_i {\sum\nolimits_k {min({h_k}({c_i}),ma{x_{j \in m}}{h_k}({s_{ij}}))} } }}{{\sum\nolimits_i {\sum\nolimits_k {{h_k}({c_i})} } }},
\end{equation}
where ${c_i}$ is the candidate answer and its corresponding group of reference answers is ${S_i} = \left\{ {{s_{i1}},{s_{i2}},...,{s_{im}}} \right\} \in S$, ${\omega _k}$ represents the possible n-grams of the $k$-th group, ${h_k}({c_i})$ represents the number of occurrences of ${\omega _k}$ in the candidate answer ${c_i}$, and ${h_k}({s_{ij}})$ represents the number of occurrences of ${\omega _k}$ in the reference answer ${s_{ij}}$. BLEU is the weighted geometric average of n-grams coincidence accuracy, which represents the ratio of the correct matching times of n-grams to the occurrence times of all n-grams. The calculation formula is as follows:
\begin{equation}
BLE{U_N}(C,S) = b(C,S)exp(\sum\nolimits_{n = 1}^N {{\omega _n}} logC{P_n}(C,S)),
\end{equation}
where N=1,2,3,4, ${\omega _k}$ is generally $\frac{1}{n}$ for every $n$.

\begin{figure}
    \centering
    \includegraphics[width=0.9\linewidth]{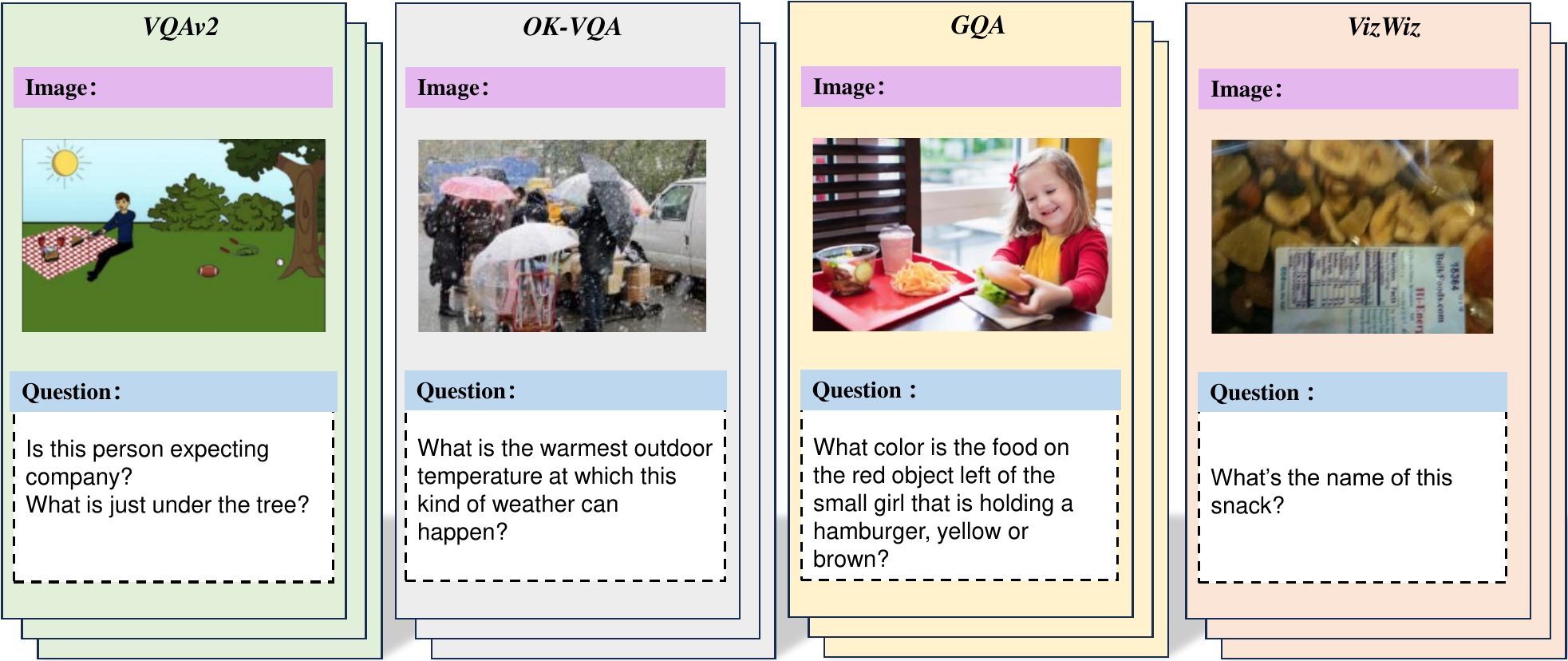}
    \caption{Examples of four widely-used datasets.}
    \label{fig:data}
\end{figure}

\begin{table}[]
\caption{Comparisons of performance of different models on four widely-used datasets.}
\scalebox{0.9}{
\begin{tabular}{cccc|cccc}
\midrule[1.3pt]
Model name & Year & Dataset        & VQA Accuruacy   & Model name   & Year & Dataset     & VQA Accuruacy    \\ \midrule[1.3pt]
mPLUG-Huge & 2022 & VQAv2 test-std & 83.62           & PaLI-X       & 2023 & OK-VQA      & 66.1             \\
Florence   & 2021 & VQAv2 test-std & 80.36           & Prophet      & 2023 & OK-VQA      & 62.5             \\
LLaVa      & 2023 & VQAv2 test-std & 78.5(few-shot)  & Flamingo     & 2022 & OK-VQA      & 50.6             \\
Flamingo   & 2022 & VQAv2 test-std & 67.6(few-shot)  & PICa         & 2021 & OK-VQA      & 48.0             \\
BLIP-2     & 2023 & VQAv2 test-std & 65(zero-shot)   & BLIP-2       & 2023 & OK-VQA      & 45.9             \\
IMG2LLM    & 2023 & VQAv2 test-std & 61.9(zero-shot) & MuKEA        & 2022 & OK-VQA      & 42.59            \\
UNITER     & 2019 & VQAv2 test-std & 73.4            & KRISP        & 2021 & OK-VQA      & 38.9             \\
LXMERT     & 2019 & VQAv2 test-std & 72.5            & ConceptBERT  & 2020 & OK-VQA      & 33.66            \\
VisualBERT & 2019 & VQAv2 test-std & 71.0            & LXMERT       & 2019 & OK-VQA      & 32.04            \\
Up-Down    & 2017 & VQAv2 test-std & 70.3            & ViLBERT      & 2019 & OK-VQA      & 31.35            \\
DMN        & 2018 & VQAv2 test-std & 68.4            & Mucko        & 2020 & OK-VQA      & 29.2             \\
MUTAN      & 2017 & VQAv2 test-std & 67.4            & MUTAN        & 2017 & OK-VQA      & 26.41            \\ \midrule[1.3pt]
PaLI-X     & 2023 & GQA test-dev   & 67.3            & mOLUG-owl2   & 2023 & VizWiz 2020 & 54.5(zero-shot)  \\
LXMERT     & 2019 & GQA test-dev   & 60.0            & LLaVA        & 2023 & VizWiz 2020 & 50.0(zero-shot)  \\
BLIP-2     & 2023 & GQA test-dev   & 44.7(zero-shot) & KOSMOS-1     & 2023 & VizWiz 2020 & 35.3(few-shot)   \\
TRRNet     & 2023 & GQA Test2019   & 74.03           & InstructBLIP & 2023 & VizWiz 2020 & 32.08(zero-shot) \\
VinVL      & 2023 & GQA Test2019   & 64.85           & Flamingo     & 2022 & VizWiz 2020 & 49.8(few-shot)   \\
BAN        & 2017 & GQA Test2020   & 57.1            & PaLI         & 2022 & VizWiz 2020 & 73.3             \\
Up-Down    & 2017 & GQA Test2021   & 49.74           & CLIP         & 2022 & VizWiz 2020 & 61.64           \\ \midrule[1.3pt]
\end{tabular}}\label{tab:performance}
\end{table}

\subsection{Examples and Comparisons of the Performance of Four Widely-used Datasets}
To enhance comprehension of the discrete contributions made by diverse datasets and the focal points within the evaluation framework, we elected to examine four datasets that are widely utilized. An overview of each dataset's attributes and illustrative instances of image-related challenges is provided in Fig. \ref{fig:data}.

For every dataset, we have synthesized an appraisal of the performances delivered by assorted models, encompassing the most advanced methodologies, shown in Table. \ref{tab:performance}. This comparative analysis encompasses outcomes from conventional deep learning architectures, attention mechanism-driven approaches, vision-language pre-training models, and multimodal large language models. A critical consideration in these evaluations is the variation in experimental configurations across different models, particularly in scenarios with limited or absent samples, which can lead to a substantial diminution in performance compared to standard settings. It is our aspiration that these exhaustive performance comparisons facilitate a deeper understanding among researchers of the disparities between models and inspire novel investigative endeavors.

\section{PROBLEMS AND CHALLENGES}\label{sec7}
We present problems and challenges that can be further improved with respect to the existing status of VQA.
\paragraph{\textbf{Visual Reasoning Techniques}} As the ability to capture information indirectly presented in images is still underdeveloped, the current VQA is more of a ``What you see is what you get'' task. Knowledge inference techniques can be applied to this problem to discover more latent knowledge. Also, more common-sense knowledge can be applied to help machines for image understanding.
The current VQA is more of a ``What you see is what you get'' task. Currently, the ability to capture information indirectly presented in images is at a lower level of development.
\paragraph{\textbf{Robustness (data bias)}} VQA models are often trained on datasets that contain inherent biases in both visual and language modalities. As a result, models tend to rely on statistical correlations rather than genuinely understanding the relationship between the question and the image content. For example, in some datasets, certain objects or scenes are more frequently associated with specific answers, leading the model to produce incorrect results when these patterns are not present. This undermines the model's ability to generalize effectively to unseen or real-world data.
\paragraph{\textbf{Explainability}}VQA models often behave as black boxes, providing answers without clear reasoning, which raises concerns about the trustworthiness of the system. Current methods, such as attention mechanisms, only partially reveal the decision-making process, often failing to offer human-interpretable explanations. Enhancing the explainability of VQA systems is crucial for their adoption in sensitive fields, like medical imaging or autonomous vehicles.
\paragraph{\textbf{Natural Language Generation}} Most of the methods choose from a given library of answers and answer questions through prediction-based methods. However, the answer generation capability has significant limitations. Although there are researches in open-ended VQA, the ability to generate answers still faces a big challenge.

\section{CONCLUSION}\label{sec8}
Since the VQA task was proposed, it has received great attention and gained rapid development and wide application. This survey presents a comprehensive review of the state-of-the-art on VQA task from two aspects. For the understanding of image-question pairs, feature extraction from individual modalities and information fusion methods between modalities are introduced, with particular attention to the application of visual-language pre-training models, and multimodal large language models. For the knowledge reasoning of graphical knowledge, we review knowledge sources and describe methods of acquisition and specific reasoning processes, highlighting differences in the type and difficulty of the models they include. Datasets of VQA and different evaluation metrics follow. We believe that the suggested possible future directions will benefit the specific task of VQA as well as the general objective of visual scene understanding.

\bibliographystyle{plainnat}
\bibliography{ref}

\begin{thebibliography}{526}
\providecommand{\natexlab}[1]{#1}
\providecommand{\url}[1]{\texttt{#1}}
\expandafter\ifx\csname urlstyle\endcsname\relax
  \providecommand{\doi}[1]{doi: #1}\else
  \providecommand{\doi}{doi: \begingroup \urlstyle{rm}\Url}\fi

\bibitem[{$\!\!$}()]{1984:1040142}
{$\!\!$}.
\newblock \emph{SIGCOMM Comput. Commun. Rev.}, 13-14\penalty0 (5-1), 1984.
\newblock ISSN 0146-4833.

\bibitem[Cze(2008)]{Czerwinski:2008:1358628}
\emph{CHI '08: CHI '08 extended abstracts on Human factors in computing systems}, New York, NY, USA, 2008. ACM.
\newblock ISBN 978-1-60558-012-X.
\newblock General Chair-Czerwinski, Mary and General Chair-Lund, Arnie and Program Chair-Tan, Desney.

\bibitem[Abacha et~al.(2019)Abacha, Hasan, Datla, Liu, Demner-Fushman, and M{\"u}ller]{abacha2019vqa}
Asma~Ben Abacha, Sadid~A Hasan, Vivek~V Datla, Joey Liu, Dina Demner-Fushman, and Henning M{\"u}ller.
\newblock Vqa-med: Overview of the medical visual question answering task at imageclef 2019.
\newblock \emph{CLEF (Working Notes)}, 2\penalty0 (6), 2019.

\bibitem[Abavisani et~al.(2020)Abavisani, Wu, Hu, Tetreault, and Jaimes]{abavisani2020multimodal}
Mahdi Abavisani, Liwei Wu, Shengli Hu, Joel Tetreault, and Alejandro Jaimes.
\newblock Multimodal categorization of crisis events in social media.
\newblock In \emph{Proceedings of the IEEE/CVF Conference on Computer Vision and Pattern Recognition}, pages 14679--14689, 2020.

\bibitem[Ablamowicz and Fauser(2007)]{Ablamowicz07}
Rafal Ablamowicz and Bertfried Fauser.
\newblock Clifford: a maple 11 package for clifford algebra computations, version 11, 2007.
\newblock URL \url{http://math.tntech.edu/rafal/cliff11/index.html}.

\bibitem[Abril and Plant(2007)]{Abril07}
Patricia~S. Abril and Robert Plant.
\newblock The patent holder's dilemma: Buy, sell, or troll?
\newblock \emph{Communications of the ACM}, 50\penalty0 (1):\penalty0 36--44, January 2007.
\newblock \doi{10.1145/1188913.1188915}.
\newblock URL \url{http://doi.acm.org/10.1145/1219092.1219093}.

\bibitem[Adya et~al.(2004)Adya, Bahl, Padhye, A.Wolman, and Zhou]{Adya-01}
A.~Adya, P.~Bahl, J.~Padhye, A.Wolman, and L.~Zhou.
\newblock A multi-radio unification protocol for {IEEE} 802.11 wireless networks.
\newblock In \emph{Proceedings of the IEEE 1st International Conference on Broadnets Networks (BroadNets'04)}, pages 210--217, Los Alamitos, CA, 2004. IEEE.

\bibitem[Afouras et~al.(2018)Afouras, Chung, and Zisserman]{afouras2018lrs3}
Triantafyllos Afouras, Joon~Son Chung, and Andrew Zisserman.
\newblock Lrs3-ted: a large-scale dataset for visual speech recognition.
\newblock \emph{arXiv preprint arXiv:1809.00496}, 2018.

\bibitem[Agrawal et~al.(2018)Agrawal, Batra, Parikh, and Kembhavi]{agrawal2018don}
Aishwarya Agrawal, Dhruv Batra, Devi Parikh, and Aniruddha Kembhavi.
\newblock Don't just assume; look and answer: Overcoming priors for visual question answering.
\newblock In \emph{Proceedings of the IEEE conference on computer vision and pattern recognition}, pages 4971--4980, 2018.

\bibitem[Akyildiz et~al.(2002)Akyildiz, Su, Sankarasubramaniam, and Cayirci]{Akyildiz-01}
I.~F. Akyildiz, W.~Su, Y.~Sankarasubramaniam, and E.~Cayirci.
\newblock Wireless sensor networks: A survey.
\newblock \emph{Comm. ACM}, 38\penalty0 (4):\penalty0 393--422, 2002.

\bibitem[Akyildiz et~al.(2007)Akyildiz, Melodia, and Chowdhury]{Akyildiz-02}
I.~F. Akyildiz, T.~Melodia, and K.~R. Chowdhury.
\newblock A survey on wireless multimedia sensor networks.
\newblock \emph{Computer Netw.}, 51\penalty0 (4):\penalty0 921--960, 2007.

\bibitem[Alam et~al.(2018)Alam, Ofli, and Imran]{alam2018crisismmd}
Firoj Alam, Ferda Ofli, and Muhammad Imran.
\newblock Crisismmd: Multimodal twitter datasets from natural disasters.
\newblock In \emph{Twelfth international AAAI conference on web and social media}, 2018.

\bibitem[Alani et~al.(2003)Alani, Kim, Millard, Weal, Hall, Lewis, and Shadbolt]{alani2003automatic}
Harith Alani, Sanghee Kim, David~E Millard, Mark~J Weal, Wendy Hall, Paul~H Lewis, and Nigel~R Shadbolt.
\newblock Automatic ontology-based knowledge extraction from web documents.
\newblock \emph{IEEE Intelligent Systems}, 18\penalty0 (1):\penalty0 14--21, 2003.

\bibitem[Alayrac et~al.(2022)Alayrac, Donahue, Luc, Miech, Barr, Hasson, Lenc, Mensch, Millican, Reynolds, et~al.]{alayrac2022flamingo}
Jean-Baptiste Alayrac, Jeff Donahue, Pauline Luc, Antoine Miech, Iain Barr, Yana Hasson, Karel Lenc, Arthur Mensch, Katherine Millican, Malcolm Reynolds, et~al.
\newblock Flamingo: a visual language model for few-shot learning.
\newblock \emph{Advances in neural information processing systems}, 35:\penalty0 23716--23736, 2022.

\bibitem[Alberti et~al.(2019)Alberti, Ling, Collins, and Reitter]{alberti2019fusion}
Chris Alberti, Jeffrey Ling, Michael Collins, and David Reitter.
\newblock Fusion of detected objects in text for visual question answering.
\newblock In \emph{Proceedings of the 2019 Conference on Empirical Methods in Natural Language Processing and the 9th International Joint Conference on Natural Language Processing (EMNLP-IJCNLP)}, pages 2131--2140, 2019.

\bibitem[Ame(2015)]{Amsthm15}
\emph{Using the amsthm Package}.
\newblock American Mathematical Society, April 2015.
\newblock \url{http://www.ctan.org/pkg/amsthm}.

\bibitem[Anderson et~al.(2018)Anderson, He, Buehler, Teney, Johnson, Gould, and Zhang]{anderson2018bottom}
Peter Anderson, Xiaodong He, Chris Buehler, Damien Teney, Mark Johnson, Stephen Gould, and Lei Zhang.
\newblock Bottom-up and top-down attention for image captioning and visual question answering.
\newblock In \emph{Proceedings of the IEEE conference on computer vision and pattern recognition}, pages 6077--6086, 2018.

\bibitem[Andler(1979)]{Andler79}
Sten Andler.
\newblock Predicate path expressions.
\newblock In \emph{Proceedings of the 6th. ACM SIGACT-SIGPLAN symposium on Principles of Programming Languages}, POPL '79, pages 226--236, New York, NY, 1979. ACM Press.
\newblock \doi{10.1145/567752.567774}.
\newblock URL \url{http://doi.acm.org/10.1145/567752.567774}.

\bibitem[Andreas et~al.(2015)Andreas, Rohrbach, Darrell, and Dan]{2015Deep}
J.~Andreas, M.~Rohrbach, T.~Darrell, and K.~Dan.
\newblock Deep compositional question answering with neural module networks.
\newblock \emph{Computer Science}, 27:\penalty0 55--56, 2015.

\bibitem[Andreas et~al.(2016{\natexlab{a}})Andreas, Rohrbach, Darrell, and Klein]{andreas2016learning}
Jacob Andreas, Marcus Rohrbach, Trevor Darrell, and Dan Klein.
\newblock Learning to compose neural networks for question answering.
\newblock In \emph{Proceedings of the 2016 Conference of the North {A}merican Chapter of the Association for Computational Linguistics: Human Language Technologies}, pages 1545--1554, San Diego, California, June 2016{\natexlab{a}}. Association for Computational Linguistics.
\newblock \doi{10.18653/v1/N16-1181}.
\newblock URL \url{https://aclanthology.org/N16-1181}.

\bibitem[Andreas et~al.(2016{\natexlab{b}})Andreas, Rohrbach, Darrell, and Klein]{andreas2016neural}
Jacob Andreas, Marcus Rohrbach, Trevor Darrell, and Dan Klein.
\newblock Neural module networks.
\newblock In \emph{Proceedings of the IEEE conference on computer vision and pattern recognition}, pages 39--48, 2016{\natexlab{b}}.

\bibitem[Anisi(2003)]{anisi03}
David~A. Anisi.
\newblock Optimal motion control of a ground vehicle.
\newblock Master's thesis, Royal Institute of Technology (KTH), Stockholm, Sweden, 2003.

\bibitem[Antol et~al.(2015)Antol, Agrawal, Lu, Mitchell, Batra, Zitnick, and Parikh]{antol2015vqa}
Stanislaw Antol, Aishwarya Agrawal, Jiasen Lu, Margaret Mitchell, Dhruv Batra, C~Lawrence Zitnick, and Devi Parikh.
\newblock Vqa: Visual question answering.
\newblock In \emph{Proceedings of the IEEE international conference on computer vision}, pages 2425--2433, 2015.

\bibitem[Anzaroot and McCallum(2013)]{UMassCitations}
Sam Anzaroot and Andrew McCallum.
\newblock {UMass} citation field extraction dataset, 2013.
\newblock URL \url{http://www.iesl.cs.umass.edu/data/data-umasscitationfield}.

\bibitem[Anzaroot et~al.(2014)Anzaroot, Passos, Belanger, and McCallum]{AnzarootPBM14}
Sam Anzaroot, Alexandre Passos, David Belanger, and Andrew McCallum.
\newblock Learning soft linear constraints with application to citation field extraction, 2014.

\bibitem[{Archer, Jr.} et~al.(1984){Archer, Jr.}, Conway, and Schneider]{6:1:1}
J.~E. {Archer, Jr.}, R.~Conway, and F.~B. Schneider.
\newblock User recovery and reversal in interactive systems.
\newblock \emph{ACM Trans. Program. Lang. Syst.}, 6\penalty0 (1):\penalty0 1--19, January 1984.

\bibitem[Auer et~al.(2007)Auer, Bizer, Kobilarov, Lehmann, Cyganiak, and Ives]{auer2007dbpedia}
S{\"o}ren Auer, Christian Bizer, Georgi Kobilarov, Jens Lehmann, Richard Cyganiak, and Zachary Ives.
\newblock Dbpedia: A nucleus for a web of open data.
\newblock In \emph{The semantic web}, pages 722--735. Springer, 2007.

\bibitem[Baevski et~al.(2021)Baevski, Hsu, Conneau, and Auli]{baevski2021unsupervised}
Alexei Baevski, Wei-Ning Hsu, Alexis Conneau, and Michael Auli.
\newblock Unsupervised speech recognition.
\newblock \emph{Advances in Neural Information Processing Systems}, 34:\penalty0 27826--27839, 2021.

\bibitem[Bagher~Zadeh et~al.(2018)Bagher~Zadeh, Liang, Poria, Cambria, and Morency]{bagher-zadeh-etal-2018-multimodal}
AmirAli Bagher~Zadeh, Paul~Pu Liang, Soujanya Poria, Erik Cambria, and Louis-Philippe Morency.
\newblock Multimodal language analysis in the wild: {CMU}-{MOSEI} dataset and interpretable dynamic fusion graph.
\newblock In \emph{Proceedings of the 56th Annual Meeting of the Association for Computational Linguistics (Volume 1: Long Papers)}, pages 2236--2246, Melbourne, Australia, July 2018. Association for Computational Linguistics.
\newblock \doi{10.18653/v1/P18-1208}.
\newblock URL \url{https://aclanthology.org/P18-1208}.

\bibitem[Bahl et~al.(2004)Bahl, Chancre, and Dungeon]{Bahl-02}
P.~Bahl, R.~Chancre, and J.~Dungeon.
\newblock {SSCH}: Slotted seeded channel hopping for capacity improvement in {IEEE} 802.11 ad-hoc wireless networks.
\newblock In \emph{Proceeding of the 10th International Conference on Mobile Computing and Networking (MobiCom'04)}, pages 112--117, New York, NY, 2004. ACM.

\bibitem[Bai et~al.(2023)Bai, Bai, Yang, Wang, Tan, Wang, Lin, Zhou, and Zhou]{Qwen-VL}
Jinze Bai, Shuai Bai, Shusheng Yang, Shijie Wang, Sinan Tan, Peng Wang, Junyang Lin, Chang Zhou, and Jingren Zhou.
\newblock Qwen-vl: A versatile vision-language model for understanding, localization, text reading, and beyond.
\newblock \emph{arXiv preprint arXiv:2308.12966}, 2023.

\bibitem[Bai et~al.(2021)Bai, Li, Ding, Shen, and Zheng]{bai2021infobox}
Yang Bai, Ziran Li, Ning Ding, Ying Shen, and Hai-Tao Zheng.
\newblock Infobox-to-text generation with tree-like planning based attention network.
\newblock In \emph{Proceedings of the Twenty-Ninth International Conference on International Joint Conferences on Artificial Intelligence}, pages 3773--3779, 2021.

\bibitem[Baltru{\v{s}}aitis et~al.(2018)Baltru{\v{s}}aitis, Ahuja, and Morency]{baltruvsaitis2018multimodal}
Tadas Baltru{\v{s}}aitis, Chaitanya Ahuja, and Louis-Philippe Morency.
\newblock Multimodal machine learning: A survey and taxonomy.
\newblock \emph{IEEE transactions on pattern analysis and machine intelligence}, 41\penalty0 (2):\penalty0 423--443, 2018.

\bibitem[Banerjee and Lavie(2005)]{banerjee2005meteor}
Satanjeev Banerjee and Alon Lavie.
\newblock Meteor: An automatic metric for mt evaluation with improved correlation with human judgments.
\newblock In \emph{Proceedings of the acl workshop on intrinsic and extrinsic evaluation measures for machine translation and/or summarization}, pages 65--72, 2005.

\bibitem[Barra et~al.(2021)Barra, Bisogni, De~Marsico, and Ricciardi]{10.1016/j.patrec.2021.09.008}
Silvio Barra, Carmen Bisogni, Maria De~Marsico, and Stefano Ricciardi.
\newblock Visual question answering: Which investigated applications?
\newblock \emph{Pattern Recogn. Lett.}, 151\penalty0 (C):\penalty0 325–331, nov 2021.
\newblock ISSN 0167-8655.

\bibitem[Bekkerman and Tabrikian(2006)]{bekkerman2006target}
Ilya Bekkerman and Joseph Tabrikian.
\newblock Target detection and localization using mimo radars and sonars.
\newblock \emph{IEEE Transactions on Signal Processing}, 54\penalty0 (10):\penalty0 3873--3883, 2006.

\bibitem[Ben-Younes et~al.(2017)Ben-Younes, Cadene, Cord, and Thome]{ben2017mutan}
Hedi Ben-Younes, R{\'e}mi Cadene, Matthieu Cord, and Nicolas Thome.
\newblock Mutan: Multimodal tucker fusion for visual question answering.
\newblock In \emph{Proceedings of the IEEE international conference on computer vision}, pages 2612--2620, 2017.

\bibitem[Benesty et~al.(2009)Benesty, Chen, Huang, and Cohen]{benesty2009noise}
Jacob Benesty, Jingdong Chen, Yiteng Huang, and Israel Cohen.
\newblock \emph{Noise reduction in speech processing}, volume~2.
\newblock Springer Science \& Business Media, 2009.

\bibitem[Bordes et~al.(2013)Bordes, Usunier, Garcia-Duran, Weston, and Yakhnenko]{bordes2013translating}
Antoine Bordes, Nicolas Usunier, Alberto Garcia-Duran, Jason Weston, and Oksana Yakhnenko.
\newblock Translating embeddings for modeling multi-relational data.
\newblock \emph{Advances in neural information processing systems}, 26, 2013.

\bibitem[Bornmann et~al.(2019)Bornmann, Wray, and Haunschild]{Bornmann2019}
Lutz Bornmann, K.~Brad Wray, and Robin Haunschild.
\newblock Citation concept analysis {(CCA)}---a new form of citation analysis revealing the usefulness of concepts for other researchers illustrated by two exemplary case studies including classic books by {Thomas S.~Kuhn} and {Karl R.~Popper}, May 2019.

\bibitem[Bowman et~al.(1993)Bowman, Debray, and Peterson]{bowman:reasoning}
Mic Bowman, Saumya~K. Debray, and Larry~L. Peterson.
\newblock Reasoning about naming systems.
\newblock \emph{ACM Trans. Program. Lang. Syst.}, 15\penalty0 (5):\penalty0 795--825, November 1993.
\newblock \doi{10.1145/161468.161471}.

\bibitem[Braams(1991)]{braams:babel}
Johannes Braams.
\newblock Babel, a multilingual style-option system for use with latex's standard document styles.
\newblock \emph{TUGboat}, 12\penalty0 (2):\penalty0 291--301, June 1991.

\bibitem[Brown et~al.(1990)Brown, Cocke, Della~Pietra, Della~Pietra, Jelinek, Lafferty, Mercer, and Roossin]{brown1990statistical}
Peter~F Brown, John Cocke, Stephen~A Della~Pietra, Vincent~J Della~Pietra, Frederick Jelinek, John Lafferty, Robert~L Mercer, and Paul~S Roossin.
\newblock A statistical approach to machine translation.
\newblock \emph{Computational linguistics}, 16\penalty0 (2):\penalty0 79--85, 1990.

\bibitem[Brown et~al.(2020)Brown, Mann, Ryder, Subbiah, Kaplan, Dhariwal, Neelakantan, Shyam, Sastry, Askell, Agarwal, Herbert{-}Voss, Krueger, Henighan, Child, Ramesh, Ziegler, Wu, Winter, Hesse, Chen, Sigler, Litwin, Gray, Chess, Clark, Berner, McCandlish, Radford, Sutskever, and Amodei]{mict}
Tom~B. Brown, Benjamin Mann, Nick Ryder, Melanie Subbiah, Jared Kaplan, Prafulla Dhariwal, Arvind Neelakantan, Pranav Shyam, Girish Sastry, Amanda Askell, Sandhini Agarwal, Ariel Herbert{-}Voss, Gretchen Krueger, Tom Henighan, Rewon Child, Aditya Ramesh, Daniel~M. Ziegler, Jeffrey Wu, Clemens Winter, Christopher Hesse, Mark Chen, Eric Sigler, Mateusz Litwin, Scott Gray, Benjamin Chess, Jack Clark, Christopher Berner, Sam McCandlish, Alec Radford, Ilya Sutskever, and Dario Amodei.
\newblock Language models are few-shot learners.
\newblock In Hugo Larochelle, Marc'Aurelio Ranzato, Raia Hadsell, Maria{-}Florina Balcan, and Hsuan{-}Tien Lin, editors, \emph{Advances in Neural Information Processing Systems 33: Annual Conference on Neural Information Processing Systems 2020, NeurIPS 2020, December 6-12, 2020, virtual}, 2020.

\bibitem[Buss et~al.(1987{\natexlab{a}})Buss, Rosenberg, and Knott]{897367}
Jonathan~F. Buss, Arnold~L. Rosenberg, and Judson~D. Knott.
\newblock Vertex types in book-embeddings.
\newblock Technical report, Amherst, MA, USA, 1987{\natexlab{a}}.

\bibitem[Buss et~al.(1987{\natexlab{b}})Buss, Rosenberg, and Knott]{Buss:1987:VTB:897367}
Jonathan~F. Buss, Arnold~L. Rosenberg, and Judson~D. Knott.
\newblock Vertex types in book-embeddings.
\newblock Technical report, Amherst, MA, USA, 1987{\natexlab{b}}.

\bibitem[Cai et~al.(2023)Cai, Song, Guan, Chen, Luo, Yi, and Kot]{cai2023benchlmm}
Rizhao Cai, Zirui Song, Dayan Guan, Zhenhao Chen, Xing Luo, Chenyu Yi, and Alex Kot.
\newblock Benchlmm: Benchmarking cross-style visual capability of large multimodal models.
\newblock \emph{arXiv preprint arXiv:2312.02896}, 2023.

\bibitem[Campbell and Bobick(1995)]{campbell1995recognition}
Lee~W Campbell and Aaron~F Bobick.
\newblock Recognition of human body motion using phase space constraints.
\newblock In \emph{Proceedings of IEEE international conference on computer vision}, pages 624--630. IEEE, 1995.

\bibitem[Cao et~al.(2022)Cao, Qin, Zhao, and Shen]{cao2022bilateral}
Jianjian Cao, Xiameng Qin, Sanyuan Zhao, and Jianbing Shen.
\newblock Bilateral cross-modality graph matching attention for feature fusion in visual question answering.
\newblock \emph{IEEE Transactions on Neural Networks and Learning Systems}, 2022.

\bibitem[Cao et~al.(2021)Cao, Li, Liang, Wang, and Lin]{cao2021knowledge}
Qingxing Cao, Bailin Li, Xiaodan Liang, Keze Wang, and Liang Lin.
\newblock Knowledge-routed visual question reasoning: Challenges for deep representation embedding.
\newblock \emph{IEEE Transactions on Neural Networks and Learning Systems}, 2021.

\bibitem[Cathey and Dowski(2002)]{cathey2002new}
W~Thomas Cathey and Edward~R Dowski.
\newblock New paradigm for imaging systems.
\newblock \emph{Applied optics}, 41\penalty0 (29):\penalty0 6080--6092, 2002.

\bibitem[Celikyilmaz et~al.(2020)Celikyilmaz, Clark, and Gao]{celikyilmaz2020evaluation}
Asli Celikyilmaz, Elizabeth Clark, and Jianfeng Gao.
\newblock Evaluation of text generation: A survey.
\newblock \emph{arXiv preprint arXiv:2006.14799}, 2020.

\bibitem[Chao et~al.(2019)Chao, Rastogi, Yavuz, Hakkani-T{\"u}r, Chen, and Lane]{chao2019learning}
Guan-Lin Chao, Abhinav Rastogi, Semih Yavuz, Dilek Hakkani-T{\"u}r, Jindong Chen, and Ian Lane.
\newblock Learning question-guided video representation for multi-turn video question answering.
\newblock \emph{arXiv preprint arXiv:1907.13280}, 2019.

\bibitem[Chen et~al.(2023{\natexlab{a}})Chen, Han, Zhao, Zhang, Shi, Xu, and Xu]{chen2023x}
Feilong Chen, Minglun Han, Haozhi Zhao, Qingyang Zhang, Jing Shi, Shuang Xu, and Bo~Xu.
\newblock X-llm: Bootstrapping advanced large language models by treating multi-modalities as foreign languages.
\newblock \emph{arXiv preprint arXiv:2305.04160}, 2023{\natexlab{a}}.

\bibitem[Chen et~al.(2015)Chen, Wang, Chen, Gao, Xu, and Nevatia]{chen2015abc}
Kan Chen, Jiang Wang, Liang-Chieh Chen, Haoyuan Gao, Wei Xu, and Ram Nevatia.
\newblock Abc-cnn: An attention based convolutional neural network for visual question answering.
\newblock \emph{arXiv preprint arXiv:1511.05960}, 2015.

\bibitem[Chen et~al.(2023{\natexlab{b}})Chen, Zhang, Zeng, Zhang, Zhu, and Zhao]{chen2023shikra}
Keqin Chen, Zhao Zhang, Weili Zeng, Richong Zhang, Feng Zhu, and Rui Zhao.
\newblock Shikra: Unleashing multimodal llm's referential dialogue magic.
\newblock \emph{arXiv preprint arXiv:2306.15195}, 2023{\natexlab{b}}.

\bibitem[Chen et~al.(2022)Chen, Wang, Changpinyo, Piergiovanni, Padlewski, Salz, Goodman, Grycner, Mustafa, Beyer, et~al.]{chen2022pali}
Xi~Chen, Xiao Wang, Soravit Changpinyo, AJ~Piergiovanni, Piotr Padlewski, Daniel Salz, Sebastian Goodman, Adam Grycner, Basil Mustafa, Lucas Beyer, et~al.
\newblock Pali: A jointly-scaled multilingual language-image model.
\newblock \emph{arXiv preprint arXiv:2209.06794}, 2022.

\bibitem[Chen et~al.(2023{\natexlab{c}})Chen, Wang, Beyer, Kolesnikov, Wu, Voigtlaender, Mustafa, Goodman, Alabdulmohsin, Padlewski, Salz, Xiong, Vlasic, Pavetic, Rong, Yu, Keysers, Zhai, and Soricut]{PaLI-3}
Xi~Chen, Xiao Wang, Lucas Beyer, Alexander Kolesnikov, Jialin Wu, Paul Voigtlaender, Basil Mustafa, Sebastian Goodman, Ibrahim Alabdulmohsin, Piotr Padlewski, Daniel Salz, Xi~Xiong, Daniel Vlasic, Filip Pavetic, Keran Rong, Tianli Yu, Daniel Keysers, Xiaohua Zhai, and Radu Soricut.
\newblock Pali-3 vision language models: Smaller, faster, stronger, 2023{\natexlab{c}}.

\bibitem[Chen et~al.(2020{\natexlab{a}})Chen, Jia, and Xiang]{chen2020review}
Xiaojun Chen, Shengbin Jia, and Yang Xiang.
\newblock A review: Knowledge reasoning over knowledge graph.
\newblock \emph{Expert Systems with Applications}, 141:\penalty0 112948, 2020{\natexlab{a}}.

\bibitem[Chen et~al.(2020{\natexlab{b}})Chen, Li, Yu, El~Kholy, Ahmed, Gan, Cheng, and Liu]{chen2020uniter}
Yen-Chun Chen, Linjie Li, Licheng Yu, Ahmed El~Kholy, Faisal Ahmed, Zhe Gan, Yu~Cheng, and Jingjing Liu.
\newblock Uniter: Universal image-text representation learning.
\newblock In \emph{European conference on computer vision}, pages 104--120. Springer, 2020{\natexlab{b}}.

\bibitem[Chen et~al.(2024{\natexlab{a}})Chen, Wang, Tian, Ye, Gao, Cui, Tong, Hu, Luo, Ma, et~al.]{chen2024far}
Zhe Chen, Weiyun Wang, Hao Tian, Shenglong Ye, Zhangwei Gao, Erfei Cui, Wenwen Tong, Kongzhi Hu, Jiapeng Luo, Zheng Ma, et~al.
\newblock How far are we to gpt-4v? closing the gap to commercial multimodal models with open-source suites.
\newblock \emph{arXiv preprint arXiv:2404.16821}, 2024{\natexlab{a}}.

\bibitem[Chen et~al.(2024{\natexlab{b}})Chen, Wu, Wang, Su, Chen, Xing, Zhong, Zhang, Zhu, Lu, et~al.]{chen2024internvl}
Zhe Chen, Jiannan Wu, Wenhai Wang, Weijie Su, Guo Chen, Sen Xing, Muyan Zhong, Qinglong Zhang, Xizhou Zhu, Lewei Lu, et~al.
\newblock Internvl: Scaling up vision foundation models and aligning for generic visual-linguistic tasks.
\newblock In \emph{Proceedings of the IEEE/CVF Conference on Computer Vision and Pattern Recognition}, pages 24185--24198, 2024{\natexlab{b}}.

\bibitem[Chen et~al.(2021)Chen, Chen, Geng, Pan, Yuan, and Chen]{chen2021zero}
Zhuo Chen, Jiaoyan Chen, Yuxia Geng, Jeff~Z Pan, Zonggang Yuan, and Huajun Chen.
\newblock Zero-shot visual question answering using knowledge graph.
\newblock In \emph{International Semantic Web Conference}, pages 146--162. Springer, 2021.

\bibitem[Cheng et~al.(2001)Cheng, Jiang, Sun, and Wang]{cheng2001color}
Heng-Da Cheng, X\_~H\_ Jiang, Ying Sun, and Jingli Wang.
\newblock Color image segmentation: advances and prospects.
\newblock \emph{Pattern recognition}, 34\penalty0 (12):\penalty0 2259--2281, 2001.

\bibitem[Cho et~al.(2014)Cho, Van~Merri{\"e}nboer, Gulcehre, Bahdanau, Bougares, Schwenk, and Bengio]{cho2014learning}
Kyunghyun Cho, Bart Van~Merri{\"e}nboer, Caglar Gulcehre, Dzmitry Bahdanau, Fethi Bougares, Holger Schwenk, and Yoshua Bengio.
\newblock Learning phrase representations using rnn encoder-decoder for statistical machine translation.
\newblock \emph{arXiv preprint arXiv:1406.1078}, 2014.

\bibitem[Chorowski et~al.(2015)Chorowski, Bahdanau, Serdyuk, Cho, and Bengio]{chorowski2015attention}
Jan~K Chorowski, Dzmitry Bahdanau, Dmitriy Serdyuk, Kyunghyun Cho, and Yoshua Bengio.
\newblock Attention-based models for speech recognition.
\newblock \emph{Advances in neural information processing systems}, 28, 2015.

\bibitem[Clark(1991)]{clark:pct}
Malcolm Clark.
\newblock Post congress tristesse.
\newblock In \emph{TeX90 Conference Proceedings}, pages 84--89. TeX Users Group, March 1991.

\bibitem[Clarkson(1985{\natexlab{a}})]{Clarkson85}
Kenneth~L. Clarkson.
\newblock \emph{Algorithms for Closest-Point Problems (Computational Geometry)}.
\newblock PhD thesis, Stanford University, Palo Alto, CA, 1985{\natexlab{a}}.
\newblock UMI Order Number: AAT 8506171.

\bibitem[Clarkson(1985{\natexlab{b}})]{Clarkson:1985:ACP:911891}
Kenneth~Lee Clarkson.
\newblock \emph{Algorithms for Closest-Point Problems (Computational Geometry)}.
\newblock PhD thesis, Stanford University, Stanford, CA, USA, 1985{\natexlab{b}}.
\newblock AAT 8506171.

\bibitem[Cohen()]{JCohen96}
Cohen.
\newblock Special issue: Digital libraries, November 1996.

\bibitem[Cohen et~al.(2007)Cohen, Nutt, and Sagic]{Cohen07}
Sarah Cohen, Werner Nutt, and Yehoshua Sagic.
\newblock Deciding equivalances among conjunctive aggregate queries.
\newblock \emph{J. ACM}, 54\penalty0 (2), April 2007.
\newblock \doi{10.1145/1219092.1219093}.
\newblock URL \url{http://doi.acm.org/10.1145/1219092.1219093}.

\bibitem[Conti et~al.(2009{\natexlab{a}})Conti, Di~Pietro, Mancini, and Mei]{1555162}
Mauro Conti, Roberto Di~Pietro, Luigi~V. Mancini, and Alessandro Mei.
\newblock (old) distributed data source verification in wireless sensor networks.
\newblock \emph{Inf. Fusion}, 10\penalty0 (4):\penalty0 342--353, 2009{\natexlab{a}}.
\newblock ISSN 1566-2535.
\newblock \doi{http://dx.doi.org/10.1016/j.inffus.2009.01.002}.

\bibitem[Conti et~al.(2009{\natexlab{b}})Conti, Di~Pietro, Mancini, and Mei]{Conti:2009:DDS:1555009.1555162}
Mauro Conti, Roberto Di~Pietro, Luigi~V. Mancini, and Alessandro Mei.
\newblock (new) distributed data source verification in wireless sensor networks.
\newblock \emph{Inf. Fusion}, 10\penalty0 (4):\penalty0 342--353, October 2009{\natexlab{b}}.
\newblock ISSN 1566-2535.
\newblock \doi{10.1016/j.inffus.2009.01.002}.
\newblock URL \url{http://portal.acm.org/citation.cfm?id=1555009.1555162}.

\bibitem[CROSSBOW()]{CROSSBOW}
CROSSBOW.
\newblock {XBOW} sensor motes specifications, 2008.
\newblock http://www.xbow.com.

\bibitem[Cui et~al.(2023)Cui, Zhou, Yang, Wu, Zhang, Zou, and Yao]{cui2023holistic}
Chenhang Cui, Yiyang Zhou, Xinyu Yang, Shirley Wu, Linjun Zhang, James Zou, and Huaxiu Yao.
\newblock Holistic analysis of hallucination in gpt-4v (ision): Bias and interference challenges.
\newblock \emph{arXiv preprint arXiv:2311.03287}, 2023.

\bibitem[Culler et~al.(2004)Culler, Estrin, and Srivastava]{Culler-01}
D.~Culler, D.~Estrin, and M.~Srivastava.
\newblock Overview of sensor networks.
\newblock \emph{IEEE Comput.}, 37\penalty0 (8 (Special Issue on Sensor Networks)):\penalty0 41--49, 2004.

\bibitem[Dai et~al.(2023)Dai, Li, Li, Tiong, Zhao, Wang, Li, Fung, and Hoi]{instructblip}
Wenliang Dai, Junnan Li, Dongxu Li, Anthony Meng~Huat Tiong, Junqi Zhao, Weisheng Wang, Boyang Li, Pascale Fung, and Steven C.~H. Hoi.
\newblock Instructblip: Towards general-purpose vision-language models with instruction tuning.
\newblock In Alice Oh, Tristan Naumann, Amir Globerson, Kate Saenko, Moritz Hardt, and Sergey Levine, editors, \emph{Advances in Neural Information Processing Systems 36: Annual Conference on Neural Information Processing Systems 2023, NeurIPS 2023, New Orleans, LA, USA, December 10 - 16, 2023}, 2023.

\bibitem[Davis et~al.(1993)Davis, Shrobe, and Szolovits]{davis1993knowledge}
Randall Davis, Howard Shrobe, and Peter Szolovits.
\newblock What is a knowledge representation?
\newblock \emph{AI magazine}, 14\penalty0 (1):\penalty0 17--17, 1993.

\bibitem[de~Faria et~al.(2023)de~Faria, de~Castro~Bastos, da~Silva, Fabris, de~Sousa~Uchoa, de~Aguiar~Neto, and dos Santos]{Faria2023VQA}
Ana Claudia Akemi~Matsuki de~Faria, Felype de~Castro~Bastos, Jose Victor Nogueira~Alves da~Silva, Vitor~Lopes Fabris, Valeska de~Sousa~Uchoa, D{\'{e}}cio~Gon{\c{c}}alves de~Aguiar~Neto, and Claudio Filipi~Goncalves dos Santos.
\newblock Visual question answering: {A} survey on techniques and common trends in recent literature.
\newblock \emph{CoRR}, abs/2305.11033, 2023.

\bibitem[Della~Pietra(1994)]{della1994mathematics}
Vincent~J Della~Pietra.
\newblock The mathematics of statistical machine translation: Parameter estimation.
\newblock \emph{Using Large Corpora}, page 223, 1994.

\bibitem[Deng et~al.(2009)Deng, Dong, Socher, Li, Li, and Fei-Fei]{deng2009imagenet}
Jia Deng, Wei Dong, Richard Socher, Li-Jia Li, Kai Li, and Li~Fei-Fei.
\newblock Imagenet: A large-scale hierarchical image database.
\newblock In \emph{2009 IEEE conference on computer vision and pattern recognition}, pages 248--255. Ieee, 2009.

\bibitem[Deng et~al.(2020{\natexlab{a}})Deng, Lam, Xie, Chen, Li, Yang, and Shen]{deng2020joint}
Yang Deng, Wai Lam, Yuexiang Xie, Daoyuan Chen, Yaliang Li, Min Yang, and Ying Shen.
\newblock Joint learning of answer selection and answer summary generation in community question answering.
\newblock In \emph{Proceedings of the AAAI Conference on Artificial Intelligence}, volume~34, pages 7651--7658, 2020{\natexlab{a}}.

\bibitem[Deng et~al.(2020{\natexlab{b}})Deng, Zhang, Li, Yang, Lam, and Shen]{deng2020bridging}
Yang Deng, Wenxuan Zhang, Yaliang Li, Min Yang, Wai Lam, and Ying Shen.
\newblock Bridging hierarchical and sequential context modeling for question-driven extractive answer summarization.
\newblock In \emph{Proceedings of the 43rd International ACM SIGIR Conference on Research and Development in Information Retrieval}, pages 1693--1696, 2020{\natexlab{b}}.

\bibitem[Dettmers et~al.(2023)Dettmers, Pagnoni, Holtzman, and Zettlemoyer]{qlora}
Tim Dettmers, Artidoro Pagnoni, Ari Holtzman, and Luke Zettlemoyer.
\newblock Qlora: Efficient finetuning of quantized llms.
\newblock In A.~Oh, T.~Neumann, A.~Globerson, K.~Saenko, M.~Hardt, and S.~Levine, editors, \emph{Advances in Neural Information Processing Systems}, volume~36, pages 10088--10115. Curran Associates, Inc., 2023.

\bibitem[Devlin et~al.(2018)Devlin, Chang, Lee, and Toutanova]{devlin2018bert}
Jacob Devlin, Ming-Wei Chang, Kenton Lee, and Kristina Toutanova.
\newblock Bert: Pre-training of deep bidirectional transformers for language understanding.
\newblock \emph{arXiv preprint arXiv:1810.04805}, 2018.

\bibitem[Dijkstra(1979)]{Dijkstra:1979:GSC:1241515.1241518}
E.~Dijkstra.
\newblock Go to statement considered harmful.
\newblock In \emph{Classics in software engineering (incoll)}, pages 27--33. Yourdon Press, Upper Saddle River, NJ, USA, 1979.
\newblock ISBN 0-917072-14-6.
\newblock URL \url{http://portal.acm.org/citation.cfm?id=1241515.1241518}.

\bibitem[Ding et~al.(2022)Ding, Yu, Liu, Hu, Cui, and Wu]{ding2022mukea}
Yang Ding, Jing Yu, Bang Liu, Yue Hu, Mingxin Cui, and Qi~Wu.
\newblock Mukea: Multimodal knowledge extraction and accumulation for knowledge-based visual question answering.
\newblock In \emph{Proceedings of the IEEE/CVF Conference on Computer Vision and Pattern Recognition}, pages 5089--5098, 2022.

\bibitem[Ding et~al.(2023)Ding, Luo, Chung, and Han]{ding2023vqa}
Yihao Ding, Siwen Luo, Hyunsuk Chung, and Soyeon~Caren Han.
\newblock Vqa: A new dataset for real-world vqa on pdf documents.
\newblock In \emph{Joint European Conference on Machine Learning and Knowledge Discovery in Databases}, pages 585--601. Springer, 2023.

\bibitem[Dong et~al.(2022)Dong, Li, Gong, Chen, Li, Shen, and Yang]{dong2021survey}
Chenhe Dong, Yinghui Li, Haifan Gong, Miaoxin Chen, Junxin Li, Ying Shen, and Min Yang.
\newblock A survey of natural language generation.
\newblock \emph{ACM Comput. Surv.}, 55\penalty0 (8), dec 2022.
\newblock ISSN 0360-0300.
\newblock \doi{10.1145/3554727}.
\newblock URL \url{https://doi.org/10.1145/3554727}.

\bibitem[Dong et~al.()Dong, Li, Dai, Zheng, Wu, Chang, Sun, Xu, Li, and Sui]{surveyicl}
Qingxiu Dong, Lei Li, Damai Dai, Ce~Zheng, Zhiyong Wu, Baobao Chang, Xu~Sun, Jingjing Xu, Lei Li, and Zhifang Sui.
\newblock A survey for in-context learning.
\newblock \emph{CoRR}.

\bibitem[Dong et~al.(2023)Dong, Li, Dai, Zheng, Wu, Chang, Sun, Xu, Li, and Sui]{ICL74}
Qingxiu Dong, Lei Li, Damai Dai, Ce~Zheng, Zhiyong Wu, Baobao Chang, Xu~Sun, Jingjing Xu, Lei Li, and Zhifang Sui.
\newblock A survey for in-context learning.
\newblock \emph{CoRR}, abs/2301.00234, 2023.

\bibitem[Dong et~al.(2014)Dong, Gabrilovich, Heitz, Horn, Lao, Murphy, Strohmann, Sun, and Zhang]{dong2014knowledge}
Xin Dong, Evgeniy Gabrilovich, Geremy Heitz, Wilko Horn, Ni~Lao, Kevin Murphy, Thomas Strohmann, Shaohua Sun, and Wei Zhang.
\newblock Knowledge vault: A web-scale approach to probabilistic knowledge fusion.
\newblock In \emph{Proceedings of the 20th ACM SIGKDD international conference on Knowledge discovery and data mining}, pages 601--610, 2014.

\bibitem[Dosovitskiy et~al.(2020)Dosovitskiy, Beyer, Kolesnikov, Weissenborn, Zhai, Unterthiner, Dehghani, Minderer, Heigold, Gelly, et~al.]{dosovitskiy2020image}
Alexey Dosovitskiy, Lucas Beyer, Alexander Kolesnikov, Dirk Weissenborn, Xiaohua Zhai, Thomas Unterthiner, Mostafa Dehghani, Matthias Minderer, Georg Heigold, Sylvain Gelly, et~al.
\newblock An image is worth 16x16 words: Transformers for image recognition at scale.
\newblock \emph{arXiv preprint arXiv:2010.11929}, 2020.

\bibitem[Dou et~al.(2022)Dou, Xu, Gan, Wang, Wang, Wang, Zhu, Zhang, Yuan, Peng, et~al.]{dou2022empirical}
Zi-Yi Dou, Yichong Xu, Zhe Gan, Jianfeng Wang, Shuohang Wang, Lijuan Wang, Chenguang Zhu, Pengchuan Zhang, Lu~Yuan, Nanyun Peng, et~al.
\newblock An empirical study of training end-to-end vision-and-language transformers.
\newblock In \emph{Proceedings of the IEEE/CVF Conference on Computer Vision and Pattern Recognition}, pages 18166--18176, 2022.

\bibitem[Douglass et~al.(1998)Douglass, Harel, and Trakhtenbrot]{Douglass98}
Bruce~P. Douglass, David Harel, and Mark~B. Trakhtenbrot.
\newblock Statecarts in use: structured analysis and object-orientation.
\newblock In Grzegorz Rozenberg and Frits~W. Vaandrager, editors, \emph{Lectures on Embedded Systems}, volume 1494 of \emph{Lecture Notes in Computer Science}, pages 368--394. Springer-Verlag, London, 1998.
\newblock \doi{10.1007/3-540-65193-4_29}.
\newblock URL \url{http://dx.doi.org/10.1007/3-540-65193-4_29}.

\bibitem[Driess et~al.(2023)Driess, Xia, Sajjadi, Lynch, Chowdhery, Ichter, Wahid, Tompson, Vuong, Yu, Huang, Chebotar, Sermanet, Duckworth, Levine, Vanhoucke, Hausman, Toussaint, Greff, Zeng, Mordatch, and Florence]{driess2023palme}
Danny Driess, Fei Xia, Mehdi S.~M. Sajjadi, Corey Lynch, Aakanksha Chowdhery, Brian Ichter, Ayzaan Wahid, Jonathan Tompson, Quan Vuong, Tianhe Yu, Wenlong Huang, Yevgen Chebotar, Pierre Sermanet, Daniel Duckworth, Sergey Levine, Vincent Vanhoucke, Karol Hausman, Marc Toussaint, Klaus Greff, Andy Zeng, Igor Mordatch, and Pete Florence.
\newblock Palm-e: An embodied multimodal language model.
\newblock In Andreas Krause, Emma Brunskill, Kyunghyun Cho, Barbara Engelhardt, Sivan Sabato, and Jonathan Scarlett, editors, \emph{International Conference on Machine Learning, {ICML} 2023, 23-29 July 2023, Honolulu, Hawaii, {USA}}, volume 202 of \emph{Proceedings of Machine Learning Research}, pages 8469--8488. {PMLR}, 2023.

\bibitem[Dunlop and Basili(1985)]{7:1:137}
D.~D. Dunlop and V.~R. Basili.
\newblock Generalizing specifications for uniformly implemented loops.
\newblock \emph{ACM Trans. Program. Lang. Syst.}, 7\penalty0 (1):\penalty0 137--158, January 1985.

\bibitem[Editor(2007)]{Editor00}
Ian Editor, editor.
\newblock \emph{The title of book one}, volume~9 of \emph{The name of the series one}.
\newblock University of Chicago Press, Chicago, 1st. edition, 2007.
\newblock \doi{10.1007/3-540-09237-4}.
\newblock URL \url{http://dx.doi.org/10.1007/3-540-09456-9}.

\bibitem[Editor(2008)]{Editor00a}
Ian Editor, editor.
\newblock \emph{The title of book two}, chapter 100.
\newblock The name of the series two. University of Chicago Press, Chicago, 2nd. edition, 2008.
\newblock \doi{10.1007/3-540-09237-4}.
\newblock URL \url{http://dx.doi.org/10.1007/3-540-09456-9}.

\bibitem[El~Ayadi et~al.(2011)El~Ayadi, Kamel, and Karray]{el2011survey}
Moataz El~Ayadi, Mohamed~S Kamel, and Fakhri Karray.
\newblock Survey on speech emotion recognition: Features, classification schemes, and databases.
\newblock \emph{Pattern recognition}, 44\penalty0 (3):\penalty0 572--587, 2011.

\bibitem[Farhadi et~al.(2010)Farhadi, Hejrati, Sadeghi, Young, Rashtchian, Hockenmaier, and Forsyth]{farhadi2010every}
Ali Farhadi, Mohsen Hejrati, Mohammad~Amin Sadeghi, Peter Young, Cyrus Rashtchian, Julia Hockenmaier, and David Forsyth.
\newblock Every picture tells a story: Generating sentences from images.
\newblock In \emph{European conference on computer vision}, pages 15--29. Springer, 2010.

\bibitem[Farneb{\"a}ck(2003)]{farneback2003two}
Gunnar Farneb{\"a}ck.
\newblock Two-frame motion estimation based on polynomial expansion.
\newblock In \emph{Scandinavian conference on Image analysis}, pages 363--370. Springer, 2003.

\bibitem[Fear(2005)]{Fear05}
Simon Fear.
\newblock \emph{Publication quality tables in {\LaTeX}}, April 2005.
\newblock \url{http://www.ctan.org/pkg/booktabs}.

\bibitem[Fu et~al.(2023{\natexlab{a}})Fu, Chen, Shen, Qin, Zhang, Lin, Yang, Zheng, Li, Sun, Wu, and Ji]{fu2023mme}
Chaoyou Fu, Peixian Chen, Yunhang Shen, Yulei Qin, Mengdan Zhang, Xu~Lin, Jinrui Yang, Xiawu Zheng, Ke~Li, Xing Sun, Yunsheng Wu, and Rongrong Ji.
\newblock Mme: A comprehensive evaluation benchmark for multimodal large language models.
\newblock \emph{arXiv preprint arXiv:2306.13394}, 2023{\natexlab{a}}.

\bibitem[Fu et~al.(2023{\natexlab{b}})Fu, Zhang, Lin, Wang, Gao, Luo, Huang, Zhang, Qiu, Ye, et~al.]{fu2023challenger}
Chaoyou Fu, Renrui Zhang, Haojia Lin, Zihan Wang, Timin Gao, Yongdong Luo, Yubo Huang, Zhengye Zhang, Longtian Qiu, Gaoxiang Ye, et~al.
\newblock A challenger to gpt-4v? early explorations of gemini in visual expertise.
\newblock \emph{arXiv preprint arXiv:2312.12436}, 2023{\natexlab{b}}.

\bibitem[Fukui et~al.(2016{\natexlab{a}})Fukui, Park, Yang, Rohrbach, Darrell, and Rohrbach]{fukui-etal-2016-multimodal}
Akira Fukui, Dong~Huk Park, Daylen Yang, Anna Rohrbach, Trevor Darrell, and Marcus Rohrbach.
\newblock Multimodal compact bilinear pooling for visual question answering and visual grounding.
\newblock In \emph{Proceedings of the 2016 Conference on Empirical Methods in Natural Language Processing}, pages 457--468, Austin, Texas, November 2016{\natexlab{a}}. Association for Computational Linguistics.
\newblock \doi{10.18653/v1/D16-1044}.
\newblock URL \url{https://aclanthology.org/D16-1044}.

\bibitem[Fukui et~al.(2016{\natexlab{b}})Fukui, Park, Yang, Rohrbach, Darrell, and Rohrbach]{fukui2016multimodal}
Akira Fukui, Dong~Huk Park, Daylen Yang, Anna Rohrbach, Trevor Darrell, and Marcus Rohrbach.
\newblock Multimodal compact bilinear pooling for visual question answering and visual grounding.
\newblock In \emph{Proceedings of the 2016 Conference on Empirical Methods in Natural Language Processing}, pages 457--468, Austin, Texas, November 2016{\natexlab{b}}. Association for Computational Linguistics.
\newblock \doi{10.18653/v1/D16-1044}.
\newblock URL \url{https://aclanthology.org/D16-1044}.

\bibitem[Gan et~al.(2020)Gan, Chen, Li, Zhu, Cheng, and Liu]{gan2020large}
Zhe Gan, Yen-Chun Chen, Linjie Li, Chen Zhu, Yu~Cheng, and Jingjing Liu.
\newblock Large-scale adversarial training for vision-and-language representation learning.
\newblock \emph{Advances in Neural Information Processing Systems}, 33:\penalty0 6616--6628, 2020.

\bibitem[Gao et~al.(2022)Gao, Ping, Thattai, Reganti, Wu, and Natarajan]{gao2022transform}
Feng Gao, Qing Ping, Govind Thattai, Aishwarya Reganti, Ying~Nian Wu, and Prem Natarajan.
\newblock Transform-retrieve-generate: Natural language-centric outside-knowledge visual question answering.
\newblock In \emph{Proceedings of the IEEE/CVF Conference on Computer Vision and Pattern Recognition}, pages 5067--5077, 2022.

\bibitem[Gao et~al.(2015)Gao, Mao, Zhou, Huang, Wang, and Xu]{gao2015you}
Haoyuan Gao, Junhua Mao, Jie Zhou, Zhiheng Huang, Lei Wang, and Wei Xu.
\newblock Are you talking to a machine? dataset and methods for multilingual image question.
\newblock \emph{Advances in neural information processing systems}, 28, 2015.

\bibitem[Gao et~al.(2024)Gao, Wu, Blair, and Pagnucco]{gao2024lora}
Jingying Gao, Qi~Wu, Alan Blair, and Maurice Pagnucco.
\newblock Lora: A logical reasoning augmented dataset for visual question answering.
\newblock \emph{Advances in Neural Information Processing Systems}, 36, 2024.

\bibitem[Gao et~al.(2018{\natexlab{a}})Gao, Ge, Chen, and Nevatia]{gao2018motion}
Jiyang Gao, Runzhou Ge, Kan Chen, and Ram Nevatia.
\newblock Motion-appearance co-memory networks for video question answering.
\newblock In \emph{Proceedings of the IEEE Conference on Computer Vision and Pattern Recognition}, pages 6576--6585, 2018{\natexlab{a}}.

\bibitem[Gao et~al.(2018{\natexlab{b}})Gao, Zeng, Song, Liu, and Shen]{gao2018examine}
Lianli Gao, Pengpeng Zeng, Jingkuan Song, Xianglong Liu, and Heng~Tao Shen.
\newblock Examine before you answer: Multi-task learning with adaptive-attentions for multiple-choice vqa.
\newblock In \emph{Proceedings of the 26th ACM international conference on Multimedia}, pages 1742--1750, 2018{\natexlab{b}}.

\bibitem[Gao et~al.(2019{\natexlab{a}})Gao, Zeng, Song, Li, Liu, Mei, and Shen]{gao2019structured}
Lianli Gao, Pengpeng Zeng, Jingkuan Song, Yuan-Fang Li, Wu~Liu, Tao Mei, and Heng~Tao Shen.
\newblock Structured two-stream attention network for video question answering.
\newblock In \emph{Proceedings of the AAAI Conference on Artificial Intelligence}, volume~33, pages 6391--6398, 2019{\natexlab{a}}.

\bibitem[Gao et~al.(2018{\natexlab{c}})Gao, Li, Li, Lu, Li, Hoi, and Wang]{gao2018question}
Peng Gao, Hongsheng Li, Shuang Li, Pan Lu, Yikang Li, Steven~CH Hoi, and Xiaogang Wang.
\newblock Question-guided hybrid convolution for visual question answering.
\newblock In \emph{Proceedings of the European Conference on Computer Vision (ECCV)}, pages 469--485, 2018{\natexlab{c}}.

\bibitem[Gao et~al.(2019{\natexlab{b}})Gao, Jiang, You, Lu, Hoi, Wang, and Li]{gao2019dynamic}
Peng Gao, Zhengkai Jiang, Haoxuan You, Pan Lu, Steven~CH Hoi, Xiaogang Wang, and Hongsheng Li.
\newblock Dynamic fusion with intra-and inter-modality attention flow for visual question answering.
\newblock In \emph{Proceedings of the IEEE/CVF conference on computer vision and pattern recognition}, pages 6639--6648, 2019{\natexlab{b}}.

\bibitem[Gard{\`e}res et~al.(2020)Gard{\`e}res, Ziaeefard, Abeloos, and Lecue]{garderes2020conceptbert}
Fran{\c{c}}ois Gard{\`e}res, Maryam Ziaeefard, Baptiste Abeloos, and Freddy Lecue.
\newblock Conceptbert: Concept-aware representation for visual question answering.
\newblock In \emph{Findings of the Association for Computational Linguistics: EMNLP 2020}, pages 489--498, 2020.

\bibitem[Geiger and Meek(2005)]{GM05}
Dan Geiger and Christopher Meek.
\newblock Structured variational inference procedures and their realizations (as incol).
\newblock In \emph{Proceedings of Tenth International Workshop on Artificial Intelligence and Statistics, {\rm The Barbados}}. The Society for Artificial Intelligence and Statistics, January 2005.

\bibitem[Gerndt(1989)]{gerndt:89}
Michael Gerndt.
\newblock \emph{Automatic Parallelization for Distributed-Memory Multiprocessing Systems}.
\newblock PhD thesis, University of Bonn, Bonn, Germany, December 1989.

\bibitem[Goldwater et~al.(2009)Goldwater, Griffiths, and Johnson]{goldwater2009bayesian}
Sharon Goldwater, Thomas~L Griffiths, and Mark Johnson.
\newblock A bayesian framework for word segmentation: Exploring the effects of context.
\newblock \emph{Cognition}, 112\penalty0 (1):\penalty0 21--54, 2009.

\bibitem[Gong et~al.(2023)Gong, Lyu, Zhang, Wang, Zheng, Zhao, Liu, Zhang, Luo, and Chen]{multimodalgpt}
Tao Gong, Chengqi Lyu, Shilong Zhang, Yudong Wang, Miao Zheng, Qian Zhao, Kuikun Liu, Wenwei Zhang, Ping Luo, and Kai Chen.
\newblock Multimodal-gpt: {A} vision and language model for dialogue with humans.
\newblock \emph{CoRR}, abs/2305.04790, 2023.

\bibitem[Goossens et~al.(1999)Goossens, Rahtz, Moore, and Sutor]{Goossens:1999:LWC:553897}
Michel Goossens, S.~P. Rahtz, Ross Moore, and Robert~S. Sutor.
\newblock \emph{The Latex Web Companion: Integrating TEX, HTML, and XML}.
\newblock Addison-Wesley Longman Publishing Co., Inc., Boston, MA, USA, 1st edition, 1999.
\newblock ISBN 0201433117.

\bibitem[Goyal et~al.(2017)Goyal, Khot, Summers-Stay, Batra, and Parikh]{goyal2017making}
Yash Goyal, Tejas Khot, Douglas Summers-Stay, Dhruv Batra, and Devi Parikh.
\newblock Making the v in vqa matter: Elevating the role of image understanding in visual question answering.
\newblock In \emph{Proceedings of the IEEE conference on computer vision and pattern recognition}, pages 6904--6913, 2017.

\bibitem[Graves(2012)]{graves2012long}
Alex Graves.
\newblock Long short-term memory.
\newblock \emph{Supervised sequence labelling with recurrent neural networks}, pages 37--45, 2012.

\bibitem[Grosz(1977)]{grosz1977representation}
Barbara~Jean Grosz.
\newblock \emph{The representation and use of focus in dialogue understanding.}
\newblock University of California, Berkeley, 1977.

\bibitem[Gundy et~al.(2007)Gundy, Balzarotti, and Vigna]{VanGundy07}
Matthew~Van Gundy, Davide Balzarotti, and Giovanni Vigna.
\newblock Catch me, if you can: Evading network signatures with web-based polymorphic worms.
\newblock In \emph{Proceedings of the first USENIX workshop on Offensive Technologies}, WOOT '07, Berkley, CA, 2007. USENIX Association.

\bibitem[Gundy et~al.(2008)Gundy, Balzarotti, and Vigna]{VanGundy08}
Matthew~Van Gundy, Davide Balzarotti, and Giovanni Vigna.
\newblock Catch me, if you can: Evading network signatures with web-based polymorphic worms.
\newblock In \emph{Proceedings of the first USENIX workshop on Offensive Technologies}, WOOT '08, pages 99--100, Berkley, CA, 2008. USENIX Association.

\bibitem[Gundy et~al.(2009)Gundy, Balzarotti, and Vigna]{VanGundy09}
Matthew~Van Gundy, Davide Balzarotti, and Giovanni Vigna.
\newblock Catch me, if you can: Evading network signatures with web-based polymorphic worms.
\newblock In \emph{Proceedings of the first USENIX workshop on Offensive Technologies}, WOOT '09, pages 90--100, Berkley, CA, 2009. USENIX Association.

\bibitem[Guo et~al.(2023)Guo, Li, Li, Tiong, Li, Tao, and Hoi]{guo2023images}
Jiaxian Guo, Junnan Li, Dongxu Li, Anthony Meng~Huat Tiong, Boyang Li, Dacheng Tao, and Steven Hoi.
\newblock From images to textual prompts: Zero-shot visual question answering with frozen large language models.
\newblock In \emph{Proceedings of the IEEE/CVF Conference on Computer Vision and Pattern Recognition}, pages 10867--10877, 2023.

\bibitem[Gupta and Kembhavi(2023{\natexlab{a}})]{Gupta_2023_CVPR}
Tanmay Gupta and Aniruddha Kembhavi.
\newblock Visual programming: Compositional visual reasoning without training.
\newblock In \emph{Proceedings of the IEEE/CVF Conference on Computer Vision and Pattern Recognition (CVPR)}, pages 14953--14962, June 2023{\natexlab{a}}.

\bibitem[Gupta and Kembhavi(2023{\natexlab{b}})]{gupta2023visual}
Tanmay Gupta and Aniruddha Kembhavi.
\newblock Visual programming: Compositional visual reasoning without training.
\newblock In \emph{Proceedings of the IEEE/CVF Conference on Computer Vision and Pattern Recognition}, pages 14953--14962, 2023{\natexlab{b}}.

\bibitem[Gurari et~al.(2018)Gurari, Li, Stangl, Guo, Lin, Grauman, Luo, and Bigham]{gurari2018vizwiz}
Danna Gurari, Qing Li, Abigale~J Stangl, Anhong Guo, Chi Lin, Kristen Grauman, Jiebo Luo, and Jeffrey~P Bigham.
\newblock Vizwiz grand challenge: Answering visual questions from blind people.
\newblock In \emph{Proceedings of the IEEE conference on computer vision and pattern recognition}, pages 3608--3617, 2018.

\bibitem[Hagerup et~al.(1993)Hagerup, Mehlhorn, and Munro]{Hagerup1993}
Torben Hagerup, Kurt Mehlhorn, and J.~Ian Munro.
\newblock Maintaining discrete probability distributions optimally.
\newblock In \emph{Proceedings of the 20th International Colloquium on Automata, Languages and Programming}, volume 700 of \emph{Lecture Notes in Computer Science}, pages 253--264, Berlin, 1993. Springer-Verlag.

\bibitem[Haralick et~al.(1973)Haralick, Shanmugam, and Dinstein]{haralick1973textural}
Robert~M Haralick, Karthikeyan Shanmugam, and Its'~Hak Dinstein.
\newblock Textural features for image classification.
\newblock \emph{IEEE Transactions on systems, man, and cybernetics}, \penalty0 (6):\penalty0 610--621, 1973.

\bibitem[Harel(1978)]{Harel78}
David Harel.
\newblock Logics of programs: Axiomatics and descriptive power.
\newblock MIT Research Lab Technical Report TR-200, Massachusetts Institute of Technology, Cambridge, MA, 1978.

\bibitem[Harel(1979)]{Harel79}
David Harel.
\newblock \emph{First-Order Dynamic Logic}, volume~68 of \emph{Lecture Notes in Computer Science}.
\newblock Springer-Verlag, New York, NY, 1979.
\newblock \doi{10.1007/3-540-09237-4}.
\newblock URL \url{http://dx.doi.org/10.1007/3-540-09237-4}.

\bibitem[Harvard CodeBlue()]{Harvard-01}
Harvard CodeBlue.
\newblock {CodeBlue}: Sensor networks for medical care, 2008.
\newblock http://www.eecs.harvard.edu/mdw/ proj/codeblue/.

\bibitem[Haurilet et~al.(2019)Haurilet, Roitberg, and Stiefelhagen]{haurilet2019s}
Monica Haurilet, Alina Roitberg, and Rainer Stiefelhagen.
\newblock It's not about the journey; it's about the destination: Following soft paths under question-guidance for visual reasoning.
\newblock In \emph{Proceedings of the IEEE/CVF Conference on Computer Vision and Pattern Recognition}, pages 1930--1939, 2019.

\bibitem[He et~al.(2016)He, Zhang, Ren, and Sun]{he2016deep}
Kaiming He, Xiangyu Zhang, Shaoqing Ren, and Jian Sun.
\newblock Deep residual learning for image recognition.
\newblock In \emph{Proceedings of the IEEE conference on computer vision and pattern recognition}, pages 770--778, 2016.

\bibitem[He et~al.(2020{\natexlab{a}})He, Deng, Wang, Li, Zhang, and Wang]{he2020lightgcn}
Xiangnan He, Kuan Deng, Xiang Wang, Yan Li, Yongdong Zhang, and Meng Wang.
\newblock Lightgcn: Simplifying and powering graph convolution network for recommendation.
\newblock In \emph{Proceedings of the 43rd International ACM SIGIR conference on research and development in Information Retrieval}, pages 639--648, 2020{\natexlab{a}}.

\bibitem[He et~al.(2020{\natexlab{b}})He, Zhang, Mou, Xing, and Xie]{he2020pathvqa}
Xuehai He, Yichen Zhang, Luntian Mou, Eric Xing, and Pengtao Xie.
\newblock Pathvqa: 30000+ questions for medical visual question answering.
\newblock \emph{arXiv preprint arXiv:2003.10286}, 2020{\natexlab{b}}.

\bibitem[Heering and Klint(1985)]{7:2:183}
J.~Heering and P.~Klint.
\newblock Towards monolingual programming environments.
\newblock \emph{ACM Trans. Program. Lang. Syst.}, 7\penalty0 (2):\penalty0 183--213, April 1985.

\bibitem[Heo et~al.(2022)Heo, Kim, Choi, and Zhang]{heo2022hypergraph}
Yu-Jung Heo, Eun-Sol Kim, Woo~Suk Choi, and Byoung-Tak Zhang.
\newblock Hypergraph transformer: Weakly-supervised multi-hop reasoning for knowledge-based visual question answering.
\newblock \emph{arXiv preprint arXiv:2204.10448}, 2022.

\bibitem[Herdade et~al.(2019)Herdade, Kappeler, Boakye, and Soares]{herdade2019image}
Simao Herdade, Armin Kappeler, Kofi Boakye, and Joao Soares.
\newblock Image captioning: Transforming objects into words.
\newblock \emph{Advances in Neural Information Processing Systems}, 32, 2019.

\bibitem[Herlihy(1993)]{herlihy:methodology}
Maurice Herlihy.
\newblock A methodology for implementing highly concurrent data objects.
\newblock \emph{ACM Trans. Program. Lang. Syst.}, 15\penalty0 (5):\penalty0 745--770, November 1993.
\newblock \doi{10.1145/161468.161469}.

\bibitem[Himakunthala et~al.(2023)Himakunthala, Ouyang, Rose, He, Mei, Lu, Sonar, Saxon, and Wang]{DBLP:conf/emnlp/HimakunthalaORH23}
Vaishnavi Himakunthala, Andy Ouyang, Daniel Rose, Ryan He, Alex Mei, Yujie Lu, Chinmay Sonar, Michael Saxon, and William~Yang Wang.
\newblock Let's think frame by frame with {VIP:} {A} video infilling and prediction dataset for evaluating video chain-of-thought.
\newblock In Houda Bouamor, Juan Pino, and Kalika Bali, editors, \emph{Proceedings of the 2023 Conference on Empirical Methods in Natural Language Processing, {EMNLP} 2023, Singapore, December 6-10, 2023}, pages 204--219. Association for Computational Linguistics, 2023.

\bibitem[Hirschman and Gaizauskas(2001)]{hirschman2001natural}
Lynette Hirschman and Robert Gaizauskas.
\newblock Natural language question answering: the view from here.
\newblock \emph{natural language engineering}, 7\penalty0 (4):\penalty0 275--300, 2001.

\bibitem[Hoare(1972)]{Hoare:1972:CIN:1243380.1243382}
C.~A.~R. Hoare.
\newblock Chapter ii: Notes on data structuring.
\newblock In O.~J. Dahl, E.~W. Dijkstra, and C.~A.~R. Hoare, editors, \emph{Structured programming (incoll)}, pages 83--174. Academic Press Ltd., London, UK, UK, 1972.
\newblock ISBN 0-12-200550-3.
\newblock URL \url{http://portal.acm.org/citation.cfm?id=1243380.1243382}.

\bibitem[Hogan et~al.(2021)Hogan, Blomqvist, Cochez, d’Amato, Melo, Gutierrez, Kirrane, Gayo, Navigli, Neumaier, et~al.]{hogan2021knowledge}
Aidan Hogan, Eva Blomqvist, Michael Cochez, Claudia d’Amato, Gerard~De Melo, Claudio Gutierrez, Sabrina Kirrane, Jos{\'e} Emilio~Labra Gayo, Roberto Navigli, Sebastian Neumaier, et~al.
\newblock Knowledge graphs.
\newblock \emph{ACM Computing Surveys (CSUR)}, 54\penalty0 (4):\penalty0 1--37, 2021.

\bibitem[Hollis(1999)]{Hollis:1999:VBD:519964}
Billy~S. Hollis.
\newblock \emph{Visual Basic 6: Design, Specification, and Objects with Other}.
\newblock Prentice Hall PTR, Upper Saddle River, NJ, USA, 1st edition, 1999.
\newblock ISBN 0130850845.

\bibitem[H{\"o}rmander(1985{\natexlab{a}})]{MR781536}
Lars H{\"o}rmander.
\newblock \emph{The analysis of linear partial differential operators. {IV}}, volume 275 of \emph{Grundlehren der Mathematischen Wissenschaften [Fundamental Principles of Mathematical Sciences]}.
\newblock Springer-Verlag, Berlin, Germany, 1985{\natexlab{a}}.
\newblock ISBN 3-540-13829-3.
\newblock Fourier integral operators.

\bibitem[H{\"o}rmander(1985{\natexlab{b}})]{MR781537}
Lars H{\"o}rmander.
\newblock \emph{The analysis of linear partial differential operators. {III}}, volume 275 of \emph{Grundlehren der Mathematischen Wissenschaften [Fundamental Principles of Mathematical Sciences]}.
\newblock Springer-Verlag, Berlin, Germany, 1985{\natexlab{b}}.
\newblock ISBN 3-540-13828-5.
\newblock Pseudodifferential operators.

\bibitem[Hossain et~al.(2019)Hossain, Sohel, Shiratuddin, and Laga]{hossain2019comprehensive}
MD~Zakir Hossain, Ferdous Sohel, Mohd~Fairuz Shiratuddin, and Hamid Laga.
\newblock A comprehensive survey of deep learning for image captioning.
\newblock \emph{ACM Computing Surveys (CsUR)}, 51\penalty0 (6):\penalty0 1--36, 2019.

\bibitem[Hu et~al.(2016)Hu, Wang, Wang, Liu, and He]{hu2016human}
Fengye Hu, Lu~Wang, Shanshan Wang, Xiaolan Liu, and Gengxin He.
\newblock A human body posture recognition algorithm based on bp neural network for wireless body area networks.
\newblock \emph{China Communications}, 13\penalty0 (8):\penalty0 198--208, 2016.

\bibitem[Hu et~al.(2021)Hu, Liu, Zhao, and Jin]{hu2021mmgcn}
Jingwen Hu, Yuchen Liu, Jinming Zhao, and Qin Jin.
\newblock {MMGCN}: Multimodal fusion via deep graph convolution network for emotion recognition in conversation.
\newblock In \emph{Proceedings of the 59th Annual Meeting of the Association for Computational Linguistics and the 11th International Joint Conference on Natural Language Processing (Volume 1: Long Papers)}, pages 5666--5675, Online, August 2021. Association for Computational Linguistics.
\newblock \doi{10.18653/v1/2021.acl-long.440}.
\newblock URL \url{https://aclanthology.org/2021.acl-long.440}.

\bibitem[Hu et~al.(2019)Hu, Rohrbach, Darrell, and Saenko]{hu2019language}
Ronghang Hu, Anna Rohrbach, Trevor Darrell, and Kate Saenko.
\newblock Language-conditioned graph networks for relational reasoning.
\newblock In \emph{Proceedings of the IEEE/CVF international conference on computer vision}, pages 10294--10303, 2019.

\bibitem[Hu et~al.(2022)Hu, Gan, Wang, Yang, Liu, Lu, and Wang]{hu2022scaling}
Xiaowei Hu, Zhe Gan, Jianfeng Wang, Zhengyuan Yang, Zicheng Liu, Yumao Lu, and Lijuan Wang.
\newblock Scaling up vision-language pre-training for image captioning.
\newblock In \emph{Proceedings of the IEEE/CVF Conference on Computer Vision and Pattern Recognition}, pages 17980--17989, 2022.

\bibitem[Huang et~al.(2020{\natexlab{a}})Huang, Chen, Zeng, Du, Tan, and Gan]{huang2020location}
Deng Huang, Peihao Chen, Runhao Zeng, Qing Du, Mingkui Tan, and Chuang Gan.
\newblock Location-aware graph convolutional networks for video question answering.
\newblock In \emph{Proceedings of the AAAI Conference on Artificial Intelligence}, volume~34, pages 11021--11028, 2020{\natexlab{a}}.

\bibitem[Huang et~al.(2017)Huang, Liu, Van Der~Maaten, and Weinberger]{huang2017densely}
Gao Huang, Zhuang Liu, Laurens Van Der~Maaten, and Kilian~Q Weinberger.
\newblock Densely connected convolutional networks.
\newblock In \emph{Proceedings of the IEEE conference on computer vision and pattern recognition}, pages 4700--4708, 2017.

\bibitem[Huang et~al.(2020{\natexlab{b}})Huang, Wei, Cai, Zheng, Chen, Leung, and Li]{huang2020aligned}
Qingbao Huang, Jielong Wei, Yi~Cai, Changmeng Zheng, Junying Chen, Ho-fung Leung, and Qing Li.
\newblock Aligned dual channel graph convolutional network for visual question answering.
\newblock In \emph{Proceedings of the 58th annual meeting of the association for computational linguistics}, pages 7166--7176, 2020{\natexlab{b}}.

\bibitem[Huang et~al.(2020{\natexlab{c}})Huang, Zeng, Liu, Fu, and Fu]{huang2020pixel}
Zhicheng Huang, Zhaoyang Zeng, Bei Liu, Dongmei Fu, and Jianlong Fu.
\newblock Pixel-bert: Aligning image pixels with text by deep multi-modal transformers.
\newblock \emph{arXiv preprint arXiv:2004.00849}, 2020{\natexlab{c}}.

\bibitem[Huang et~al.(2021)Huang, Zeng, Huang, Liu, Fu, and Fu]{huang2021seeing}
Zhicheng Huang, Zhaoyang Zeng, Yupan Huang, Bei Liu, Dongmei Fu, and Jianlong Fu.
\newblock Seeing out of the box: End-to-end pre-training for vision-language representation learning.
\newblock In \emph{Proceedings of the IEEE/CVF Conference on Computer Vision and Pattern Recognition}, pages 12976--12985, 2021.

\bibitem[Hudson and Manning(2019)]{hudson2019gqa}
Drew~A Hudson and Christopher~D Manning.
\newblock Gqa: A new dataset for real-world visual reasoning and compositional question answering.
\newblock In \emph{Proceedings of the IEEE/CVF conference on computer vision and pattern recognition}, pages 6700--6709, 2019.

\bibitem[IEEE()]{2004:ITE:1009386.1010128}
IEEE.
\newblock Ieee tcsc executive committee.
\newblock In \emph{Proceedings of the IEEE International Conference on Web Services}, ICWS '04, pages 21--22, Washington, DC, USA, 2004. IEEE Computer Society.
\newblock ISBN 0-7695-2167-3.
\newblock \doi{http://dx.doi.org/10.1109/ICWS.2004.64}.
\newblock URL \url{http://dx.doi.org/10.1109/ICWS.2004.64}.

\bibitem[Ilievski and Feng(2017)]{ilievski2017generative}
Ilija Ilievski and Jiashi Feng.
\newblock Generative attention model with adversarial self-learning for visual question answering.
\newblock In \emph{Proceedings of the on Thematic Workshops of ACM Multimedia 2017}, pages 415--423, 2017.

\bibitem[Ilievski et~al.(2016)Ilievski, Yan, and Feng]{ilievski2016focused}
Ilija Ilievski, Shuicheng Yan, and Jiashi Feng.
\newblock A focused dynamic attention model for visual question answering.
\newblock \emph{arXiv preprint arXiv:1604.01485}, 2016.

\bibitem[Ishmam et~al.(2024)Ishmam, Shovon, Mridha, and Dey]{ISHMAM2024102270}
Md.~Farhan Ishmam, Md. Sakib~Hossain Shovon, M.F. Mridha, and Nilanjan Dey.
\newblock From image to language: A critical analysis of visual question answering (vqa) approaches, challenges, and opportunities.
\newblock \emph{Information Fusion}, 106:\penalty0 102270, 2024.
\newblock ISSN 1566-2535.

\bibitem[Islam and Moushi(2024)]{islam2024gpt}
Raisa Islam and Owana~Marzia Moushi.
\newblock Gpt-4o: The cutting-edge advancement in multimodal llm.
\newblock \emph{Authorea Preprints}, 2024.

\bibitem[Iyer et~al.(2022)Iyer, Lin, Pasunuru, Mihaylov, Simig, Yu, Shuster, Wang, Liu, Koura, et~al.]{iyer2022opt}
Srinivasan Iyer, Xi~Victoria Lin, Ramakanth Pasunuru, Todor Mihaylov, Daniel Simig, Ping Yu, Kurt Shuster, Tianlu Wang, Qing Liu, Punit~Singh Koura, et~al.
\newblock Opt-iml: Scaling language model instruction meta learning through the lens of generalization.
\newblock \emph{arXiv preprint arXiv:2212.12017}, 2022.

\bibitem[Jabri et~al.(2016)Jabri, Joulin, and Maaten]{jabri2016revisiting}
Allan Jabri, Armand Joulin, and Laurens van~der Maaten.
\newblock Revisiting visual question answering baselines.
\newblock In \emph{European conference on computer vision}, pages 727--739. Springer, 2016.

\bibitem[Jang et~al.(2017)Jang, Song, Yu, Kim, and Kim]{jang2017tgif}
Yunseok Jang, Yale Song, Youngjae Yu, Youngjin Kim, and Gunhee Kim.
\newblock Tgif-qa: Toward spatio-temporal reasoning in visual question answering.
\newblock In \emph{Proceedings of the IEEE conference on computer vision and pattern recognition}, pages 2758--2766, 2017.

\bibitem[Jia et~al.(2021)Jia, Yang, Xia, Chen, Parekh, Pham, Le, Sung, Li, and Duerig]{jia2021scaling}
Chao Jia, Yinfei Yang, Ye~Xia, Yi-Ting Chen, Zarana Parekh, Hieu Pham, Quoc Le, Yun-Hsuan Sung, Zhen Li, and Tom Duerig.
\newblock Scaling up visual and vision-language representation learning with noisy text supervision.
\newblock In \emph{International Conference on Machine Learning}, pages 4904--4916. PMLR, 2021.

\bibitem[Jiang et~al.(2015)Jiang, Wang, Porikli, and Li]{jiang2015compositional}
Aiwen Jiang, Fang Wang, Fatih Porikli, and Yi~Li.
\newblock Compositional memory for visual question answering.
\newblock \emph{arXiv preprint arXiv:1511.05676}, 2015.

\bibitem[Jiang et~al.(2020{\natexlab{a}})Jiang, Misra, Rohrbach, Learned-Miller, and Chen]{jiang2020defense}
Huaizu Jiang, Ishan Misra, Marcus Rohrbach, Erik Learned-Miller, and Xinlei Chen.
\newblock In defense of grid features for visual question answering.
\newblock In \emph{Proceedings of the IEEE/CVF Conference on Computer Vision and Pattern Recognition}, pages 10267--10276, 2020{\natexlab{a}}.

\bibitem[Jiang et~al.(2020{\natexlab{b}})Jiang, Chen, Lin, Zhao, and Gao]{jiang2020divide}
Jianwen Jiang, Ziqiang Chen, Haojie Lin, Xibin Zhao, and Yue Gao.
\newblock Divide and conquer: Question-guided spatio-temporal contextual attention for video question answering.
\newblock In \emph{Proceedings of the AAAI Conference on Artificial Intelligence}, volume~34, pages 11101--11108, 2020{\natexlab{b}}.

\bibitem[Jiang et~al.(2018)Jiang, Natarajan, Chen, Rohrbach, Batra, and Parikh]{jiang2018pythia}
Yu~Jiang, Vivek Natarajan, Xinlei Chen, Marcus Rohrbach, Dhruv Batra, and Devi Parikh.
\newblock Pythia v0. 1: the winning entry to the vqa challenge 2018.
\newblock \emph{arXiv preprint arXiv:1807.09956}, 2018.

\bibitem[Jiang et~al.(2024)Jiang, Peng, Feng, Li, and Li]{DBLP:journals/corr/abs-2405-06705}
Zhuoxuan Jiang, Haoyuan Peng, Shanshan Feng, Fan Li, and Dongsheng Li.
\newblock Llms can find mathematical reasoning mistakes by pedagogical chain-of-thought.
\newblock \emph{CoRR}, abs/2405.06705, 2024.

\bibitem[Jin et~al.(2019)Jin, Zhao, Gu, Yu, Xiao, and Zhuang]{jin2019multi}
Weike Jin, Zhou Zhao, Mao Gu, Jun Yu, Jun Xiao, and Yueting Zhuang.
\newblock Multi-interaction network with object relation for video question answering.
\newblock In \emph{Proceedings of the 27th ACM international conference on multimedia}, pages 1193--1201, 2019.

\bibitem[Johnson et~al.(2016)Johnson, Karpathy, and Fei-Fei]{johnson2016densecap}
Justin Johnson, Andrej Karpathy, and Li~Fei-Fei.
\newblock Densecap: Fully convolutional localization networks for dense captioning.
\newblock In \emph{Proceedings of the IEEE conference on computer vision and pattern recognition}, pages 4565--4574, 2016.

\bibitem[Johnson et~al.(2017)Johnson, Hariharan, Van Der~Maaten, Fei-Fei, Lawrence~Zitnick, and Girshick]{johnson2017clevr}
Justin Johnson, Bharath Hariharan, Laurens Van Der~Maaten, Li~Fei-Fei, C~Lawrence~Zitnick, and Ross Girshick.
\newblock Clevr: A diagnostic dataset for compositional language and elementary visual reasoning.
\newblock In \emph{Proceedings of the IEEE conference on computer vision and pattern recognition}, pages 2901--2910, 2017.

\bibitem[Kafle and Kanan(2016)]{kafle2016answer}
Kushal Kafle and Christopher Kanan.
\newblock Answer-type prediction for visual question answering.
\newblock In \emph{Proceedings of the IEEE conference on computer vision and pattern recognition}, pages 4976--4984, 2016.

\bibitem[Kafle and Kanan(2017)]{kafle2017visual}
Kushal Kafle and Christopher Kanan.
\newblock Visual question answering: Datasets, algorithms, and future challenges.
\newblock \emph{Computer Vision and Image Understanding}, 163:\penalty0 3--20, 2017.

\bibitem[Kamath et~al.(2021)Kamath, Singh, LeCun, Synnaeve, Misra, and Carion]{kamath2021mdetr}
Aishwarya Kamath, Mannat Singh, Yann LeCun, Gabriel Synnaeve, Ishan Misra, and Nicolas Carion.
\newblock Mdetr-modulated detection for end-to-end multi-modal understanding.
\newblock In \emph{Proceedings of the IEEE/CVF International Conference on Computer Vision}, pages 1780--1790, 2021.

\bibitem[Karpukhin et~al.(2020)Karpukhin, Oguz, Min, Lewis, Wu, Edunov, Chen, and Yih]{karpukhin2020dense}
Vladimir Karpukhin, Barlas Oguz, Sewon Min, Patrick Lewis, Ledell Wu, Sergey Edunov, Danqi Chen, and Wen-tau Yih.
\newblock Dense passage retrieval for open-domain question answering.
\newblock In \emph{Proceedings of the 2020 Conference on Empirical Methods in Natural Language Processing (EMNLP)}, pages 6769--6781, 2020.

\bibitem[Kazemi and Elqursh(2017)]{kazemi2017show}
Vahid Kazemi and Ali Elqursh.
\newblock Show, ask, attend, and answer: A strong baseline for visual question answering.
\newblock \emph{arXiv preprint arXiv:1704.03162}, 2017.

\bibitem[Khademi(2020)]{khademi2020multimodal}
Mahmoud Khademi.
\newblock Multimodal neural graph memory networks for visual question answering.
\newblock In \emph{Proceedings of the 58th Annual Meeting of the Association for Computational Linguistics}, pages 7177--7188, 2020.

\bibitem[Kim et~al.(2019)Kim, Kim, and Kwak]{kim2019textbook}
Daesik Kim, Seonhoon Kim, and Nojun Kwak.
\newblock Textbook question answering with multi-modal context graph understanding and self-supervised open-set comprehension.
\newblock In \emph{Proceedings of the 57th Annual Meeting of the Association for Computational Linguistics}, pages 3568--3584, 2019.

\bibitem[Kim et~al.(2016)Kim, Lee, Kwak, Heo, Kim, Ha, and Zhang]{kim2016multimodal}
Jin-Hwa Kim, Sang-Woo Lee, Donghyun Kwak, Min-Oh Heo, Jeonghee Kim, Jung-Woo Ha, and Byoung-Tak Zhang.
\newblock Multimodal residual learning for visual qa.
\newblock \emph{Advances in neural information processing systems}, 29, 2016.

\bibitem[Kim et~al.(2017{\natexlab{a}})Kim, On, Lim, Kim, Ha, and Zhang]{kim2016hadamard}
Jin-Hwa Kim, Kyoung-Woon On, Woosang Lim, Jeonghee Kim, Jung-Woo Ha, and Byoung-Tak Zhang.
\newblock Hadamard product for low-rank bilinear pooling.
\newblock In \emph{International Conference on Learning Representations}, 2017{\natexlab{a}}.
\newblock URL \url{https://openreview.net/forum?id=r1rhWnZkg}.

\bibitem[Kim et~al.(2018{\natexlab{a}})Kim, Jun, and Zhang]{kim2018bilinear}
Jin-Hwa Kim, Jaehyun Jun, and Byoung-Tak Zhang.
\newblock Bilinear attention networks.
\newblock \emph{Advances in neural information processing systems}, 31, 2018{\natexlab{a}}.

\bibitem[Kim et~al.(2017{\natexlab{b}})Kim, Heo, Choi, and Zhang]{kim2017deepstory}
Kyung-Min Kim, Min-Oh Heo, Seong-Ho Choi, and Byoung-Tak Zhang.
\newblock Deepstory: Video story qa by deep embedded memory networks.
\newblock \emph{arXiv preprint arXiv:1707.00836}, 2017{\natexlab{b}}.

\bibitem[Kim et~al.(2018{\natexlab{b}})Kim, Choi, Kim, and Zhang]{kim2018multimodal}
Kyung-Min Kim, Seong-Ho Choi, Jin-Hwa Kim, and Byoung-Tak Zhang.
\newblock Multimodal dual attention memory for video story question answering.
\newblock In \emph{Proceedings of the European Conference on Computer Vision (ECCV)}, pages 673--688, 2018{\natexlab{b}}.

\bibitem[Kim et~al.(2021)Kim, Son, and Kim]{kim2021vilt}
Wonjae Kim, Bokyung Son, and Ildoo Kim.
\newblock Vilt: Vision-and-language transformer without convolution or region supervision.
\newblock In \emph{International Conference on Machine Learning}, pages 5583--5594. PMLR, 2021.

\bibitem[Kipf and Welling(2017)]{kipf2016semi}
Thomas~N. Kipf and Max Welling.
\newblock Semi-supervised classification with graph convolutional networks.
\newblock In \emph{International Conference on Learning Representations}, 2017.
\newblock URL \url{https://openreview.net/forum?id=SJU4ayYgl}.

\bibitem[Kiros et~al.(2015)Kiros, Zhu, Salakhutdinov, Zemel, Urtasun, Torralba, and Fidler]{kiros2015skip}
Ryan Kiros, Yukun Zhu, Russ~R Salakhutdinov, Richard Zemel, Raquel Urtasun, Antonio Torralba, and Sanja Fidler.
\newblock Skip-thought vectors.
\newblock \emph{Advances in neural information processing systems}, 28, 2015.

\bibitem[Kirschmer and Voight(2010)]{Kirschmer:2010:AEI:1958016.1958018}
Markus Kirschmer and John Voight.
\newblock Algorithmic enumeration of ideal classes for quaternion orders.
\newblock \emph{SIAM J. Comput.}, 39\penalty0 (5):\penalty0 1714--1747, January 2010.
\newblock ISSN 0097-5397.
\newblock \doi{https://doi.org/10.1137/080734467}.
\newblock URL \url{http://dx.doi.org/10.1137/080734467}.

\bibitem[Knuth(1981{\natexlab{a}})]{book-minimal}
Donald~E. Knuth.
\newblock \emph{Seminumerical Algorithms}.
\newblock Addison-Wesley, 1981{\natexlab{a}}.

\bibitem[Knuth(1981{\natexlab{b}})]{test}
Donald~E. Knuth.
\newblock \emph{Seminumerical Algorithms}, volume~2 of \emph{The Art of Computer Programming}.
\newblock Addison-Wesley, Reading, MA, 2nd edition, 10~January 1981{\natexlab{b}}.

\bibitem[Knuth(1984)]{knuth:texbook}
Donald~E. Knuth.
\newblock \emph{The {\TeX{}book}}.
\newblock Addison-Wesley, Reading, MA., 1984.

\bibitem[Knuth(1997)]{Knuth97}
Donald~E. Knuth.
\newblock \emph{The Art of Computer Programming, Vol. 1: Fundamental Algorithms (3rd. ed.)}.
\newblock Addison Wesley Longman Publishing Co., Inc., 1997.

\bibitem[Knuth(1998)]{Knuth98}
Donald~E. Knuth.
\newblock \emph{The Art of Computer Programming}, volume~1 of \emph{Fundamental Algorithms}.
\newblock Addison Wesley Longman Publishing Co., Inc., 3rd edition, 1998.
\newblock (book).

\bibitem[Kodali and Berleant(2022)]{kodali2022recent}
Venkat Kodali and Daniel Berleant.
\newblock Recent, rapid advancement in visual question answering: a review.
\newblock In \emph{2022 IEEE International Conference on Electro Information Technology (eIT)}, pages 139--146. IEEE, 2022.

\bibitem[Koehn(2009)]{koehn2009statistical}
Philipp Koehn.
\newblock \emph{Statistical machine translation}.
\newblock Cambridge University Press, 2009.

\bibitem[Koh et~al.(2024)Koh, Fried, and Salakhutdinov]{koh2024generating}
Jing~Yu Koh, Daniel Fried, and Russ~R Salakhutdinov.
\newblock Generating images with multimodal language models.
\newblock \emph{Advances in Neural Information Processing Systems}, 36, 2024.

\bibitem[Kong(2001{\natexlab{a}})]{KA:2001}
Wei-Chang Kong.
\newblock The implementation of electronic commerce in smes in singapore (as incoll).
\newblock In \emph{E-commerce and cultural values}, pages 51--74. IGI Publishing, Hershey, PA, USA, 2001{\natexlab{a}}.
\newblock ISBN 1-59140-056-2.
\newblock URL \url{http://portal.acm.org/citation.cfm?id=887006.887010}.

\bibitem[Kong(2001{\natexlab{b}})]{KAGM:2001}
Wei-Chang Kong.
\newblock \emph{E-commerce and cultural values}, name of chapter: The implementation of electronic commerce in SMEs in Singapore (Inbook-w-chap-w-type), pages 51--74.
\newblock IGI Publishing, Hershey, PA, USA, 2001{\natexlab{b}}.
\newblock ISBN 1-59140-056-2.
\newblock URL \url{http://portal.acm.org/citation.cfm?id=887006.887010}.

\bibitem[Kong(2002)]{Kong:2002:IEC:887006.887010}
Wei-Chang Kong.
\newblock Chapter 9.
\newblock In Theerasak Thanasankit, editor, \emph{E-commerce and cultural values (Incoll-w-text (chap 9) 'title')}, pages 51--74. IGI Publishing, Hershey, PA, USA, 2002.
\newblock ISBN 1-59140-056-2.
\newblock URL \url{http://portal.acm.org/citation.cfm?id=887006.887010}.

\bibitem[Kong(2003)]{Kong:2003:IEC:887006.887011}
Wei-Chang Kong.
\newblock The implementation of electronic commerce in smes in singapore (incoll).
\newblock In Theerasak Thanasankit, editor, \emph{E-commerce and cultural values}, pages 51--74. IGI Publishing, Hershey, PA, USA, 2003.
\newblock ISBN 1-59140-056-2.
\newblock URL \url{http://portal.acm.org/citation.cfm?id=887006.887010}.

\bibitem[Kong(2004)]{Kong:2004:IEC:123456.887010}
Wei-Chang Kong.
\newblock \emph{E-commerce and cultural values - (InBook-num-in-chap)}, chapter~9, pages 51--74.
\newblock IGI Publishing, Hershey, PA, USA, 2004.
\newblock ISBN 1-59140-056-2.
\newblock URL \url{http://portal.acm.org/citation.cfm?id=887006.887010}.

\bibitem[Kong(2005)]{Kong:2005:IEC:887006.887010}
Wei-Chang Kong.
\newblock \emph{E-commerce and cultural values (Inbook-text-in-chap)}, chapter: The implementation of electronic commerce in SMEs in Singapore, pages 51--74.
\newblock IGI Publishing, Hershey, PA, USA, 2005.
\newblock ISBN 1-59140-056-2.
\newblock URL \url{http://portal.acm.org/citation.cfm?id=887006.887010}.

\bibitem[Kong(2006)]{Kong:2006:IEC:887006.887010}
Wei-Chang Kong.
\newblock \emph{E-commerce and cultural values (Inbook-num chap)}, chapter (in type field)~22, pages 51--74.
\newblock IGI Publishing, Hershey, PA, USA, 2006.
\newblock ISBN 1-59140-056-2.
\newblock URL \url{http://portal.acm.org/citation.cfm?id=887006.887010}.

\bibitem[Korach et~al.(1984)Korach, Rotem, and Santoro]{6:3:380}
E.~Korach, D.~Rotem, and N.~Santoro.
\newblock Distributed algorithms for finding centers and medians in networks.
\newblock \emph{ACM Trans. Program. Lang. Syst.}, 6\penalty0 (3):\penalty0 380--401, July 1984.

\bibitem[Kornerup(1994)]{ko94}
Jacob Kornerup.
\newblock Mapping powerlists onto hypercubes.
\newblock Master's thesis, The University of Texas at Austin, 1994.
\newblock (In preparation).

\bibitem[Kosiur(2001)]{Kosiur01}
David Kosiur.
\newblock \emph{Understanding Policy-Based Networking}.
\newblock Wiley, New York, NY, 2nd. edition, 2001.

\bibitem[Krishna et~al.(2017)Krishna, Zhu, Groth, Johnson, Hata, Kravitz, Chen, Kalantidis, Li, Shamma, et~al.]{krishna2017visual}
Ranjay Krishna, Yuke Zhu, Oliver Groth, Justin Johnson, Kenji Hata, Joshua Kravitz, Stephanie Chen, Yannis Kalantidis, Li-Jia Li, David~A Shamma, et~al.
\newblock Visual genome: Connecting language and vision using crowdsourced dense image annotations.
\newblock \emph{International journal of computer vision}, 123\penalty0 (1):\penalty0 32--73, 2017.

\bibitem[Krizhevsky et~al.(2017)Krizhevsky, Sutskever, and Hinton]{krizhevsky2017imagenet}
Alex Krizhevsky, Ilya Sutskever, and Geoffrey~E Hinton.
\newblock Imagenet classification with deep convolutional neural networks.
\newblock \emph{Communications of the ACM}, 60\penalty0 (6):\penalty0 84--90, 2017.

\bibitem[Kwiatkowski et~al.(2019)Kwiatkowski, Palomaki, Redfield, Collins, Parikh, Alberti, Epstein, Polosukhin, Devlin, Lee, et~al.]{kwiatkowski2019natural}
Tom Kwiatkowski, Jennimaria Palomaki, Olivia Redfield, Michael Collins, Ankur Parikh, Chris Alberti, Danielle Epstein, Illia Polosukhin, Jacob Devlin, Kenton Lee, et~al.
\newblock Natural questions: a benchmark for question answering research.
\newblock \emph{Transactions of the Association for Computational Linguistics}, 7:\penalty0 453--466, 2019.

\bibitem[Lafferty et~al.(2001)Lafferty, McCallum, and Pereira]{lafferty2001conditional}
John Lafferty, Andrew McCallum, and Fernando~CN Pereira.
\newblock Conditional random fields: Probabilistic models for segmenting and labeling sequence data.
\newblock 2001.

\bibitem[Lamport(1986)]{Lamport:LaTeX}
Leslie Lamport.
\newblock \emph{\it {\LaTeX}: A Document Preparation System}.
\newblock Addison-Wesley, Reading, MA., 1986.

\bibitem[Le et~al.(2024)Le, Nguyen, Do, Nguyen, and Nguyen]{DBLP:conf/jsai/LeNDNN24}
Nguyen{-}Khang Le, Dieu{-}Hien Nguyen, Dinh{-}Truong Do, Chau Nguyen, and Le~Minh Nguyen.
\newblock Vietnamese elementary math reasoning using large language model with refined translation and dense-retrieved chain-of-thought.
\newblock In Toyotaro Suzumura and Mayumi Bono, editors, \emph{New Frontiers in Artificial Intelligence - {JSAI} International Symposium on Artificial Intelligence, JSAI-isAI 2024, Hamamatsu, Japan, May 28-29, 2024, Proceedings}, volume 14741 of \emph{Lecture Notes in Computer Science}, pages 260--268. Springer, 2024.

\bibitem[Le and Mikolov(2014)]{le2014distributed}
Quoc Le and Tomas Mikolov.
\newblock Distributed representations of sentences and documents.
\newblock In \emph{International conference on machine learning}, pages 1188--1196. PMLR, 2014.

\bibitem[Le et~al.(2020)Le, Le, Venkatesh, and Tran]{le2020hierarchical}
Thao~Minh Le, Vuong Le, Svetha Venkatesh, and Truyen Tran.
\newblock Hierarchical conditional relation networks for video question answering.
\newblock In \emph{Proceedings of the IEEE/CVF conference on computer vision and pattern recognition}, pages 9972--9981, 2020.

\bibitem[LeCun et~al.(1989)LeCun, Boser, Denker, Henderson, Howard, Hubbard, and Jackel]{lecun1989backpropagation}
Yann LeCun, Bernhard Boser, John~S Denker, Donnie Henderson, Richard~E Howard, Wayne Hubbard, and Lawrence~D Jackel.
\newblock Backpropagation applied to handwritten zip code recognition.
\newblock \emph{Neural computation}, 1\penalty0 (4):\penalty0 541--551, 1989.

\bibitem[Lee(1981)]{Lee:1978:TQA:800025.1198348}
Jan Lee.
\newblock Transcript of question and answer session.
\newblock In Richard~L. Wexelblat, editor, \emph{History of programming languages I (incoll)}, pages 68--71. ACM, New York, NY, USA, 1981.
\newblock ISBN 0-12-745040-8.
\newblock \doi{http://doi.acm.org/10.1145/800025.1198348}.
\newblock URL \url{http://doi.acm.org/10.1145/800025.1198348}.

\bibitem[Lee(2005)]{Lee05}
Newton Lee.
\newblock Interview with bill kinder: January 13, 2005.
\newblock \emph{Comput. Entertain.}, 3\penalty0 (1):\penalty0 4, Jan.-March 2005.
\newblock \doi{10.1145/1057270.1057278}.
\newblock URL \url{http://doi.acm.org/10.1145/1057270.1057278}.

\bibitem[Lei et~al.(2020)Lei, Wu, Liu, Li, Wang, Tang, and Li]{lei2020multi}
Chenyi Lei, Lei Wu, Dong Liu, Zhao Li, Guoxin Wang, Haihong Tang, and Houqiang Li.
\newblock Multi-question learning for visual question answering.
\newblock In \emph{Proceedings of the AAAI Conference on Artificial Intelligence}, volume~34, pages 11328--11335, 2020.

\bibitem[Lei et~al.(2018)Lei, Yu, Bansal, and Berg]{lei2018tvqa}
Jie Lei, Licheng Yu, Mohit Bansal, and Tamara~L Berg.
\newblock Tvqa: Localized, compositional video question answering.
\newblock \emph{arXiv preprint arXiv:1809.01696}, 2018.

\bibitem[Lerner et~al.(2022)Lerner, Ferret, Guinaudeau, Le~Borgne, Besan{\c{c}}on, Moreno, and Lov{\'o}n~Melgarejo]{lerner2022viquae}
Paul Lerner, Olivier Ferret, Camille Guinaudeau, Herv{\'e} Le~Borgne, Romaric Besan{\c{c}}on, Jos{\'e}~G Moreno, and Jes{\'u}s Lov{\'o}n~Melgarejo.
\newblock Viquae, a dataset for knowledge-based visual question answering about named entities.
\newblock In \emph{Proceedings of the 45th International ACM SIGIR Conference on Research and Development in Information Retrieval}, pages 3108--3120, 2022.

\bibitem[Li et~al.(2023{\natexlab{a}})Li, Ge, Ge, Wang, Wang, Zhang, and Shan]{li2023seed-2}
Bohao Li, Yuying Ge, Yixiao Ge, Guangzhi Wang, Rui Wang, Ruimao Zhang, and Ying Shan.
\newblock Seed-bench-2: Benchmarking multimodal large language models.
\newblock \emph{arXiv preprint arXiv:2311.17092}, 2023{\natexlab{a}}.

\bibitem[Li et~al.(2024{\natexlab{a}})Li, Ge, Chen, Ge, Zhang, and Shan]{li2024seed}
Bohao Li, Yuying Ge, Yi~Chen, Yixiao Ge, Ruimao Zhang, and Ying Shan.
\newblock Seed-bench-2-plus: Benchmarking multimodal large language models with text-rich visual comprehension.
\newblock \emph{arXiv preprint arXiv:2404.16790}, 2024{\natexlab{a}}.

\bibitem[Li et~al.(2024{\natexlab{b}})Li, Wang, Wang, Ge, Ge, and Shan]{li2023seed}
Bohao Li, Rui Wang, Guangzhi Wang, Yuying Ge, Yixiao Ge, and Ying Shan.
\newblock Seed-bench: Benchmarking multimodal llms with generative comprehension.
\newblock In \emph{CVPR}, 2024{\natexlab{b}}.

\bibitem[Li et~al.(2008)Li, Buyuktur, Hutchful, Sant, and Nainwal]{Li:2008:PUC:1358628.1358946}
Cheng-Lun Li, Ayse~G. Buyuktur, David~K. Hutchful, Natasha~B. Sant, and Satyendra~K. Nainwal.
\newblock Portalis: using competitive online interactions to support aid initiatives for the homeless.
\newblock In \emph{CHI '08 extended abstracts on Human factors in computing systems}, pages 3873--3878, New York, NY, USA, 2008. ACM.
\newblock ISBN 978-1-60558-012-X.
\newblock \doi{10.1145/1358628.1358946}.
\newblock URL \url{http://portal.acm.org/citation.cfm?id=1358628.1358946}.

\bibitem[Li et~al.(2024{\natexlab{c}})Li, Li, and Hoi]{li2024blip}
Dongxu Li, Junnan Li, and Steven Hoi.
\newblock Blip-diffusion: Pre-trained subject representation for controllable text-to-image generation and editing.
\newblock \emph{Advances in Neural Information Processing Systems}, 36, 2024{\natexlab{c}}.

\bibitem[Li et~al.(2022{\natexlab{a}})Li, Zhang, Zhang, Liu, Guo, Ni, Zhang, and Zhang]{li2022vision}
Feng Li, Hao Zhang, Yi-Fan Zhang, Shilong Liu, Jian Guo, Lionel~M Ni, PengChuan Zhang, and Lei Zhang.
\newblock Vision-language intelligence: Tasks, representation learning, and large models.
\newblock \emph{arXiv preprint arXiv:2203.01922}, 2022{\natexlab{a}}.

\bibitem[Li et~al.(2020{\natexlab{a}})Li, Duan, Fang, Gong, and Jiang]{li2020unicoder}
Gen Li, Nan Duan, Yuejian Fang, Ming Gong, and Daxin Jiang.
\newblock Unicoder-vl: A universal encoder for vision and language by cross-modal pre-training.
\newblock In \emph{Proceedings of the AAAI Conference on Artificial Intelligence}, volume~34, pages 11336--11344, 2020{\natexlab{a}}.

\bibitem[Li et~al.(2017)Li, Su, and Zhu]{li2017incorporating}
Guohao Li, Hang Su, and Wenwu Zhu.
\newblock Incorporating external knowledge to answer open-domain visual questions with dynamic memory networks.
\newblock \emph{arXiv preprint arXiv:1712.00733}, 2017.

\bibitem[Li et~al.(2020{\natexlab{b}})Li, Wang, and Zhu]{li2020boosting}
Guohao Li, Xin Wang, and Wenwu Zhu.
\newblock Boosting visual question answering with context-aware knowledge aggregation.
\newblock In \emph{Proceedings of the 28th ACM International Conference on Multimedia}, pages 1227--1235, 2020{\natexlab{b}}.

\bibitem[Li et~al.(2021{\natexlab{a}})Li, Selvaraju, Gotmare, Joty, Xiong, and Hoi]{2021Align}
J.~Li, R.~R. Selvaraju, A.~D. Gotmare, S.~Joty, C.~Xiong, and S.~Hoi.
\newblock Align before fuse: Vision and language representation learning with momentum distillation, 2021{\natexlab{a}}.

\bibitem[Li et~al.(2020{\natexlab{c}})Li, Sun, Han, and Li]{li2020survey}
Jing Li, Aixin Sun, Jianglei Han, and Chenliang Li.
\newblock A survey on deep learning for named entity recognition.
\newblock \emph{IEEE Transactions on Knowledge and Data Engineering}, 34\penalty0 (1):\penalty0 50--70, 2020{\natexlab{c}}.

\bibitem[Li et~al.(2021{\natexlab{b}})Li, Selvaraju, Gotmare, Joty, Xiong, and Hoi]{li2021align}
Junnan Li, Ramprasaath Selvaraju, Akhilesh Gotmare, Shafiq Joty, Caiming Xiong, and Steven Chu~Hong Hoi.
\newblock Align before fuse: Vision and language representation learning with momentum distillation.
\newblock \emph{Advances in neural information processing systems}, 34:\penalty0 9694--9705, 2021{\natexlab{b}}.

\bibitem[Li et~al.(2022{\natexlab{b}})Li, Li, Xiong, and Hoi]{li2022blip}
Junnan Li, Dongxu Li, Caiming Xiong, and Steven Hoi.
\newblock Blip: Bootstrapping language-image pre-training for unified vision-language understanding and generation.
\newblock \emph{arXiv preprint arXiv:2201.12086}, 2022{\natexlab{b}}.

\bibitem[Li et~al.(2023{\natexlab{b}})Li, Li, Savarese, and Hoi]{li2023blip}
Junnan Li, Dongxu Li, Silvio Savarese, and Steven Hoi.
\newblock Blip-2: Bootstrapping language-image pre-training with frozen image encoders and large language models.
\newblock In \emph{International conference on machine learning}, pages 19730--19742. PMLR, 2023{\natexlab{b}}.

\bibitem[Li et~al.(2023{\natexlab{c}})Li, He, Wang, Li, Wang, Luo, Wang, Wang, and Qiao]{li2023videochat}
KunChang Li, Yinan He, Yi~Wang, Yizhuo Li, Wenhai Wang, Ping Luo, Yali Wang, Limin Wang, and Yu~Qiao.
\newblock Videochat: Chat-centric video understanding.
\newblock \emph{arXiv preprint arXiv:2305.06355}, 2023{\natexlab{c}}.

\bibitem[Li et~al.(2019{\natexlab{a}})Li, Yatskar, Yin, Hsieh, and Chang]{li2019simple}
LH~Li, M~Yatskar, D~Yin, CJ~Hsieh, and KW~Chang.
\newblock A simple and performant baseline for vision and language.
\newblock \emph{arXiv preprint arXiv:1908.03557}, 2019{\natexlab{a}}.

\bibitem[Li et~al.(2019{\natexlab{b}})Li, Gan, Cheng, and Liu]{li2019relation}
Linjie Li, Zhe Gan, Yu~Cheng, and Jingjing Liu.
\newblock Relation-aware graph attention network for visual question answering.
\newblock In \emph{Proceedings of the IEEE/CVF international conference on computer vision}, pages 10313--10322, 2019{\natexlab{b}}.

\bibitem[Li et~al.(2019{\natexlab{c}})Li, Yatskar, Yin, Hsieh, and Chang]{li2019visualbert}
Liunian~Harold Li, Mark Yatskar, Da~Yin, Cho-Jui Hsieh, and Kai-Wei Chang.
\newblock Visualbert: A simple and performant baseline for vision and language.
\newblock \emph{arXiv preprint arXiv:1908.03557}, 2019{\natexlab{c}}.

\bibitem[Li and Moens(2022)]{li2022dynamic}
Mingxiao Li and Marie-Francine Moens.
\newblock Dynamic key-value memory enhanced multi-step graph reasoning for knowledge-based visual question answering.
\newblock \emph{arXiv preprint arXiv:2203.02985}, 2022.

\bibitem[Li et~al.(2022{\natexlab{c}})Li, Xiao, Bhanu, Sheng, and Hong]{li2022inner}
Qun Li, Fu~Xiao, Bir Bhanu, Biyun Sheng, and Richang Hong.
\newblock Inner knowledge-based img2doc scheme for visual question answering.
\newblock \emph{ACM Transactions on Multimedia Computing, Communications, and Applications (TOMM)}, 18\penalty0 (3):\penalty0 1--21, 2022{\natexlab{c}}.

\bibitem[Li et~al.(2022{\natexlab{d}})Li, Xiao, Bhanu, Sheng, and Hong]{li_inner_2022}
Qun Li, Fu~Xiao, Bir Bhanu, Biyun Sheng, and Richang Hong.
\newblock Inner {Knowledge}-based {Img2Doc} {Scheme} for {Visual} {Question} {Answering}.
\newblock \emph{ACM Transactions on Multimedia Computing, Communications, and Applications}, 18\penalty0 (3):\penalty0 1--21, 2022{\natexlab{d}}.
\newblock ISSN 1551-6857, 1551-6865.

\bibitem[Li and Jia(2016)]{li2016visual}
Ruiyu Li and Jiaya Jia.
\newblock Visual question answering with question representation update (qru).
\newblock \emph{Advances in Neural Information Processing Systems}, 29, 2016.

\bibitem[Li et~al.(2020{\natexlab{d}})Li, Gao, Niu, Xiao, Liu, Liu, Wu, and Wang]{li2020unimo}
Wei Li, Can Gao, Guocheng Niu, Xinyan Xiao, Hao Liu, Jiachen Liu, Hua Wu, and Haifeng Wang.
\newblock Unimo: Towards unified-modal understanding and generation via cross-modal contrastive learning.
\newblock \emph{arXiv preprint arXiv:2012.15409}, 2020{\natexlab{d}}.

\bibitem[Li et~al.(2019{\natexlab{d}})Li, Chen, Hu, and Yang]{li2019understanding}
Xiang Li, Shuo Chen, Xiaolin Hu, and Jian Yang.
\newblock Understanding the disharmony between dropout and batch normalization by variance shift.
\newblock In \emph{Proceedings of the IEEE/CVF conference on computer vision and pattern recognition}, pages 2682--2690, 2019{\natexlab{d}}.

\bibitem[Li et~al.(2020{\natexlab{e}})Li, Yin, Li, Zhang, Hu, Zhang, Wang, Hu, Dong, Wei, et~al.]{li2020oscar}
Xiujun Li, Xi~Yin, Chunyuan Li, Pengchuan Zhang, Xiaowei Hu, Lei Zhang, Lijuan Wang, Houdong Hu, Li~Dong, Furu Wei, et~al.
\newblock Oscar: Object-semantics aligned pre-training for vision-language tasks.
\newblock In \emph{European Conference on Computer Vision}, pages 121--137. Springer, 2020{\natexlab{e}}.

\bibitem[Li et~al.(2024{\natexlab{d}})Li, Li, Yu, Wu, Liu, Li, Wei, and Deng]{li2024mllm}
Yanjie Li, Weijun Li, Lina Yu, Min Wu, Jingyi Liu, Wenqiang Li, Shu Wei, and Yusong Deng.
\newblock Mllm-sr: Conversational symbolic regression base multi-modal large language models.
\newblock \emph{arXiv preprint arXiv:2406.05410}, 2024{\natexlab{d}}.

\bibitem[Li et~al.(2018)Li, Duan, Zhou, Chu, Ouyang, Wang, and Zhou]{li2018visual}
Yikang Li, Nan Duan, Bolei Zhou, Xiao Chu, Wanli Ouyang, Xiaogang Wang, and Ming Zhou.
\newblock Visual question generation as dual task of visual question answering.
\newblock In \emph{Proceedings of the IEEE conference on computer vision and pattern recognition}, pages 6116--6124, 2018.

\bibitem[Li et~al.(2023{\natexlab{d}})Li, Xu, Chen, Huang, Li, Jiang, Li, Zhou, Zheng, and Shen]{li2023towards}
Yinghui Li, Zishan Xu, Shaoshen Chen, Haojing Huang, Yangning Li, Yong Jiang, Zhongli Li, Qingyu Zhou, Hai-Tao Zheng, and Ying Shen.
\newblock Towards real-world writing assistance: A chinese character checking benchmark with faked and misspelled characters.
\newblock \emph{arXiv preprint arXiv:2311.11268}, 2023{\natexlab{d}}.

\bibitem[Li et~al.(2024{\natexlab{e}})Li, Du, Liu, Zhang, Liu, Zhang, and Cai]{li2024eagle}
Zhihao Li, Yao Du, Yang Liu, Yan Zhang, Yufang Liu, Mengdi Zhang, and Xunliang Cai.
\newblock Eagle: Elevating geometric reasoning through llm-empowered visual instruction tuning.
\newblock \emph{arXiv preprint arXiv:2408.11397}, 2024{\natexlab{e}}.

\bibitem[Liang et~al.(2018)Liang, Jiang, Cao, Li, and Hauptmann]{liang2018focal}
Junwei Liang, Lu~Jiang, Liangliang Cao, Li-Jia Li, and Alexander~G Hauptmann.
\newblock Focal visual-text attention for visual question answering.
\newblock In \emph{Proceedings of the IEEE Conference on Computer Vision and Pattern Recognition}, pages 6135--6143, 2018.

\bibitem[Liang et~al.(2020)Liang, Niu, Reganti, Thattai, and Tur]{liang2020lrta}
Weixin Liang, Feiyang Niu, Aishwarya Reganti, Govind Thattai, and Gokhan Tur.
\newblock Lrta: a transparent neural-symbolic reasoning framework with modular supervision for visual question answering.
\newblock \emph{arXiv preprint arXiv:2011.10731}, 2020.

\bibitem[Liang et~al.(2021{\natexlab{a}})Liang, Jiang, and Liu]{liang2021graphvqa}
Weixin Liang, Yanhao Jiang, and Zixuan Liu.
\newblock Graphvqa: Language-guided graph neural networks for scene graph question answering.
\newblock \emph{NAACL-HLT 2021}, page~79, 2021{\natexlab{a}}.

\bibitem[Liang et~al.(2021{\natexlab{b}})Liang, Wang, Duan, and Zhu]{liang2021multi}
Yaoyuan Liang, Xin Wang, Xuguang Duan, and Wenwu Zhu.
\newblock Multi-modal contextual graph neural network for text visual question answering.
\newblock In \emph{2020 25th International Conference on Pattern Recognition (ICPR)}, pages 3491--3498. IEEE, 2021{\natexlab{b}}.

\bibitem[Lin(2004)]{lin2004rouge}
Chin-Yew Lin.
\newblock Rouge: A package for automatic evaluation of summaries.
\newblock In \emph{Text summarization branches out}, pages 74--81, 2004.

\bibitem[Lin et~al.(2020)Lin, Yang, Zhang, Liu, Zhou, and Yang]{lin2020interbert}
Junyang Lin, An~Yang, Yichang Zhang, Jie Liu, Jingren Zhou, and Hongxia Yang.
\newblock Interbert: Vision-and-language interaction for multi-modal pretraining.
\newblock \emph{arXiv preprint arXiv:2003.13198}, 2020.

\bibitem[Lin et~al.(2014)Lin, Maire, Belongie, Hays, Perona, Ramanan, Doll{\'a}r, and Zitnick]{lin2014microsoft}
Tsung-Yi Lin, Michael Maire, Serge Belongie, James Hays, Pietro Perona, Deva Ramanan, Piotr Doll{\'a}r, and C~Lawrence Zitnick.
\newblock Microsoft coco: Common objects in context.
\newblock In \emph{European conference on computer vision}, pages 740--755. Springer, 2014.

\bibitem[Lin et~al.(2015)Lin, RoyChowdhury, and Maji]{lin2015bilinear}
Tsung-Yu Lin, Aruni RoyChowdhury, and Subhransu Maji.
\newblock Bilinear cnn models for fine-grained visual recognition.
\newblock In \emph{Proceedings of the IEEE international conference on computer vision}, pages 1449--1457, 2015.

\bibitem[Liu et~al.(2022{\natexlab{a}})Liu, Hsu, Auli, and Baevski]{liu2022towards}
Alexander~H Liu, Wei-Ning Hsu, Michael Auli, and Alexei Baevski.
\newblock Towards end-to-end unsupervised speech recognition.
\newblock \emph{arXiv preprint arXiv:2204.02492}, 2022{\natexlab{a}}.

\bibitem[Liu et~al.(2023{\natexlab{a}})Liu, Guan, Li, Chen, Yacoob, Manocha, and Zhou]{liu2023hallusionbench}
Fuxiao Liu, Tianrui Guan, Zongxia Li, Lichang Chen, Yaser Yacoob, Dinesh Manocha, and Tianyi Zhou.
\newblock Hallusionbench: You see what you think? or you think what you see? an image-context reasoning benchmark challenging for gpt-4v (ision), llava-1.5, and other multi-modality models.
\newblock \emph{arXiv preprint arXiv:2310.14566}, 2023{\natexlab{a}}.

\bibitem[Liu et~al.(2022{\natexlab{b}})Liu, Tam, Muqeeth, Mohta, Huang, Bansal, and Raffel]{liu2022few}
Haokun Liu, Derek Tam, Mohammed Muqeeth, Jay Mohta, Tenghao Huang, Mohit Bansal, and Colin~A Raffel.
\newblock Few-shot parameter-efficient fine-tuning is better and cheaper than in-context learning.
\newblock \emph{Advances in Neural Information Processing Systems}, 35:\penalty0 1950--1965, 2022{\natexlab{b}}.

\bibitem[Liu et~al.(2023{\natexlab{b}})Liu, Li, Wu, and Lee]{liu2023llava}
Haotian Liu, Chunyuan Li, Qingyang Wu, and Yong~Jae Lee.
\newblock Visual instruction tuning, 2023{\natexlab{b}}.

\bibitem[Liu et~al.(2024)Liu, Li, Wu, and Lee]{liu2024visual}
Haotian Liu, Chunyuan Li, Qingyang Wu, and Yong~Jae Lee.
\newblock Visual instruction tuning.
\newblock \emph{Advances in neural information processing systems}, 36, 2024.

\bibitem[Liu and Singh(2004)]{liu2004conceptnet}
Hugo Liu and Push Singh.
\newblock Conceptnet—a practical commonsense reasoning tool-kit.
\newblock \emph{BT technology journal}, 22\penalty0 (4):\penalty0 211--226, 2004.

\bibitem[Liu et~al.(2021)Liu, Lin, Cao, Hu, Wei, Zhang, Lin, and Guo]{liu2021swin}
Ze~Liu, Yutong Lin, Yue Cao, Han Hu, Yixuan Wei, Zheng Zhang, Stephen Lin, and Baining Guo.
\newblock Swin transformer: Hierarchical vision transformer using shifted windows.
\newblock In \emph{Proceedings of the IEEE/CVF International Conference on Computer Vision}, pages 10012--10022, 2021.

\bibitem[Liu et~al.(2018)Liu, Shen, Lakshminarasimhan, Liang, Zadeh, and Morency]{liu2018efficient}
Zhun Liu, Ying Shen, Varun~Bharadhwaj Lakshminarasimhan, Paul~Pu Liang, Amir Zadeh, and Louis-Philippe Morency.
\newblock Efficient low-rank multimodal fusion with modality-specific factors.
\newblock \emph{arXiv preprint arXiv:1806.00064}, 2018.

\bibitem[Lu and Weng(2007)]{lu2007survey}
Dengsheng Lu and Qihao Weng.
\newblock A survey of image classification methods and techniques for improving classification performance.
\newblock \emph{International journal of Remote sensing}, 28\penalty0 (5):\penalty0 823--870, 2007.

\bibitem[Lu et~al.(2016)Lu, Yang, Batra, and Parikh]{lu2016hierarchical}
Jiasen Lu, Jianwei Yang, Dhruv Batra, and Devi Parikh.
\newblock Hierarchical question-image co-attention for visual question answering.
\newblock \emph{Advances in neural information processing systems}, 29, 2016.

\bibitem[Lu et~al.(2019)Lu, Batra, Parikh, and Lee]{lu2019vilbert}
Jiasen Lu, Dhruv Batra, Devi Parikh, and Stefan Lee.
\newblock Vilbert: Pretraining task-agnostic visiolinguistic representations for vision-and-language tasks.
\newblock \emph{Advances in neural information processing systems}, 32, 2019.

\bibitem[Lu et~al.(2020)Lu, Goswami, Rohrbach, Parikh, and Lee]{lu202012}
Jiasen Lu, Vedanuj Goswami, Marcus Rohrbach, Devi Parikh, and Stefan Lee.
\newblock 12-in-1: Multi-task vision and language representation learning.
\newblock In \emph{Proceedings of the IEEE/CVF Conference on Computer Vision and Pattern Recognition}, pages 10437--10446, 2020.

\bibitem[Lu et~al.(2018{\natexlab{a}})Lu, Ji, Zhang, Duan, Zhou, and Wang]{lu2018r}
Pan Lu, Lei Ji, Wei Zhang, Nan Duan, Ming Zhou, and Jianyong Wang.
\newblock R-vqa: learning visual relation facts with semantic attention for visual question answering.
\newblock In \emph{Proceedings of the 24th ACM SIGKDD International Conference on Knowledge Discovery \& Data Mining}, pages 1880--1889, 2018{\natexlab{a}}.

\bibitem[Lu et~al.(2018{\natexlab{b}})Lu, Li, Zhang, Wang, and Wang]{lu2018co}
Pan Lu, Hongsheng Li, Wei Zhang, Jianyong Wang, and Xiaogang Wang.
\newblock Co-attending free-form regions and detections with multi-modal multiplicative feature embedding for visual question answering.
\newblock In \emph{Proceedings of the AAAI Conference on Artificial Intelligence}, volume~32, 2018{\natexlab{b}}.

\bibitem[Lu et~al.(2021)Lu, Qiu, Chen, Xia, Zhao, Zhang, Yu, Liang, and Zhu]{lu2021iconqa}
Pan Lu, Liang Qiu, Jiaqi Chen, Tanglin Xia, Yizhou Zhao, Wei Zhang, Zhou Yu, Xiaodan Liang, and Song-Chun Zhu.
\newblock Iconqa: A new benchmark for abstract diagram understanding and visual language reasoning.
\newblock In \emph{NeurIPS Datasets and Benchmarks}, 2021.

\bibitem[Lu et~al.(2022{\natexlab{a}})Lu, Mishra, Xia, Qiu, Chang, Zhu, Tafjord, Clark, and Kalyan]{scienceqa}
Pan Lu, Swaroop Mishra, Tanglin Xia, Liang Qiu, Kai{-}Wei Chang, Song{-}Chun Zhu, Oyvind Tafjord, Peter Clark, and Ashwin Kalyan.
\newblock Learn to explain: Multimodal reasoning via thought chains for science question answering.
\newblock In Sanmi Koyejo, S.~Mohamed, A.~Agarwal, Danielle Belgrave, K.~Cho, and A.~Oh, editors, \emph{Advances in Neural Information Processing Systems 35: Annual Conference on Neural Information Processing Systems 2022, NeurIPS 2022, New Orleans, LA, USA, November 28 - December 9, 2022}, 2022{\natexlab{a}}.

\bibitem[Lu et~al.(2024{\natexlab{a}})Lu, Bansal, Xia, Liu, Li, Hajishirzi, Cheng, Chang, Galley, and Gao]{lu2023mathvista}
Pan Lu, Hritik Bansal, Tony Xia, Jiacheng Liu, Chunyuan Li, Hannaneh Hajishirzi, Hao Cheng, Kai-Wei Chang, Michel Galley, and Jianfeng Gao.
\newblock Mathvista: Evaluating math reasoning in visual contexts with gpt-4v, bard, and other large multimodal models.
\newblock In \emph{ICLR}, 2024{\natexlab{a}}.

\bibitem[Lu et~al.(2024{\natexlab{b}})Lu, Peng, Cheng, Galley, Chang, Wu, Zhu, and Gao]{lu2024chameleon}
Pan Lu, Baolin Peng, Hao Cheng, Michel Galley, Kai-Wei Chang, Ying~Nian Wu, Song-Chun Zhu, and Jianfeng Gao.
\newblock Chameleon: Plug-and-play compositional reasoning with large language models.
\newblock \emph{Advances in Neural Information Processing Systems}, 36, 2024{\natexlab{b}}.

\bibitem[Lu et~al.(2022{\natexlab{b}})Lu, Bartolo, Moore, Riedel, and Stenetorp]{lu2022fantastically}
Yao Lu, Max Bartolo, Alastair Moore, Sebastian Riedel, and Pontus Stenetorp.
\newblock Fantastically ordered prompts and where to find them: Overcoming few-shot prompt order sensitivity.
\newblock In \emph{Proceedings of the 60th Annual Meeting of the Association for Computational Linguistics (Volume 1: Long Papers)}, pages 8086--8098, 2022{\natexlab{b}}.

\bibitem[Ma et~al.(2018)Ma, Shen, Dick, Wu, Wang, van~den Hengel, and Reid]{ma2018visual}
Chao Ma, Chunhua Shen, Anthony Dick, Qi~Wu, Peng Wang, Anton van~den Hengel, and Ian Reid.
\newblock Visual question answering with memory-augmented networks.
\newblock In \emph{Proceedings of the IEEE conference on computer vision and pattern recognition}, pages 6975--6984, 2018.

\bibitem[Ma et~al.(2024)Ma, Wang, Kong, Wang, Liu, Pei, and Zhao]{10438044}
Jie Ma, Pinghui Wang, Dechen Kong, Zewei Wang, Jun Liu, Hongbin Pei, and Junzhou Zhao.
\newblock Robust visual question answering: Datasets, methods, and future challenges.
\newblock \emph{IEEE Transactions on Pattern Analysis and Machine Intelligence}, pages 1--20, 2024.
\newblock \doi{10.1109/TPAMI.2024.3366154}.

\bibitem[Ma et~al.(2016)Ma, Lu, and Li]{ma2016learning}
Lin Ma, Zhengdong Lu, and Hang Li.
\newblock Learning to answer questions from image using convolutional neural network.
\newblock In \emph{Thirtieth AAAI Conference on Artificial Intelligence}, 2016.

\bibitem[Ma and Liu(2018)]{ma2018review}
Zhiliang Ma and Shilong Liu.
\newblock A review of 3d reconstruction techniques in civil engineering and their applications.
\newblock \emph{Advanced Engineering Informatics}, 37:\penalty0 163--174, 2018.

\bibitem[Malinowski and Fritz(2014)]{malinowski2014multi}
Mateusz Malinowski and Mario Fritz.
\newblock A multi-world approach to question answering about real-world scenes based on uncertain input.
\newblock \emph{Advances in neural information processing systems}, 27, 2014.

\bibitem[Malinowski et~al.(2015)Malinowski, Rohrbach, and Fritz]{malinowski2015ask}
Mateusz Malinowski, Marcus Rohrbach, and Mario Fritz.
\newblock Ask your neurons: A neural-based approach to answering questions about images.
\newblock In \emph{Proceedings of the IEEE international conference on computer vision}, pages 1--9, 2015.

\bibitem[Manning et~al.(2014)Manning, Surdeanu, Bauer, Finkel, Bethard, and McClosky]{manning2014stanford}
Christopher~D Manning, Mihai Surdeanu, John Bauer, Jenny~Rose Finkel, Steven Bethard, and David McClosky.
\newblock The stanford corenlp natural language processing toolkit.
\newblock In \emph{Proceedings of 52nd annual meeting of the association for computational linguistics: system demonstrations}, pages 55--60, 2014.

\bibitem[Marino et~al.(2019)Marino, Rastegari, Farhadi, and Mottaghi]{marino2019ok}
Kenneth Marino, Mohammad Rastegari, Ali Farhadi, and Roozbeh Mottaghi.
\newblock Ok-vqa: A visual question answering benchmark requiring external knowledge.
\newblock In \emph{Proceedings of the IEEE/cvf conference on computer vision and pattern recognition}, pages 3195--3204, 2019.

\bibitem[Marino et~al.(2021)Marino, Chen, Parikh, Gupta, and Rohrbach]{marino2021krisp}
Kenneth Marino, Xinlei Chen, Devi Parikh, Abhinav Gupta, and Marcus Rohrbach.
\newblock Krisp: Integrating implicit and symbolic knowledge for open-domain knowledge-based vqa.
\newblock In \emph{Proceedings of the IEEE/CVF Conference on Computer Vision and Pattern Recognition}, pages 14111--14121, 2021.

\bibitem[Mathew et~al.(2021)Mathew, Karatzas, and Jawahar]{mathew2021docvqa}
Minesh Mathew, Dimosthenis Karatzas, and CV~Jawahar.
\newblock Docvqa: A dataset for vqa on document images.
\newblock In \emph{Proceedings of the IEEE/CVF winter conference on applications of computer vision}, pages 2200--2209, 2021.

\bibitem[Mathew et~al.(2022)Mathew, Bagal, Tito, Karatzas, Valveny, and Jawahar]{mathew2022infographicvqa}
Minesh Mathew, Viraj Bagal, Rub{\`e}n Tito, Dimosthenis Karatzas, Ernest Valveny, and CV~Jawahar.
\newblock Infographicvqa.
\newblock In \emph{Proceedings of the IEEE/CVF Winter Conference on Applications of Computer Vision}, pages 1697--1706, 2022.

\bibitem[McCracken and Golden(1990)]{McCracken:1990:SSC:575315}
Daniel~D. McCracken and Donald~G. Golden.
\newblock \emph{Simplified Structured COBOL with Microsoft/MicroFocus COBOL}.
\newblock John Wiley \& Sons, Inc., New York, NY, USA, 1990.
\newblock ISBN 0471514071.

\bibitem[Medhat et~al.(2014)Medhat, Hassan, and Korashy]{medhat2014sentiment}
Walaa Medhat, Ahmed Hassan, and Hoda Korashy.
\newblock Sentiment analysis algorithms and applications: A survey.
\newblock \emph{Ain Shams engineering journal}, 5\penalty0 (4):\penalty0 1093--1113, 2014.

\bibitem[Mikolov et~al.(2013)Mikolov, Sutskever, Chen, Corrado, and Dean]{mikolov2013distributed}
Tomas Mikolov, Ilya Sutskever, Kai Chen, Greg~S Corrado, and Jeff Dean.
\newblock Distributed representations of words and phrases and their compositionality.
\newblock \emph{Advances in neural information processing systems}, 26, 2013.

\bibitem[Minaee et~al.(2021)Minaee, Boykov, Porikli, Plaza, Kehtarnavaz, and Terzopoulos]{minaee2021image}
Shervin Minaee, Yuri~Y Boykov, Fatih Porikli, Antonio~J Plaza, Nasser Kehtarnavaz, and Demetri Terzopoulos.
\newblock Image segmentation using deep learning: A survey.
\newblock \emph{IEEE transactions on pattern analysis and machine intelligence}, 2021.

\bibitem[Mokady et~al.(2021)Mokady, Hertz, and Bermano]{mokady2021clipcap}
Ron Mokady, Amir Hertz, and Amit~H Bermano.
\newblock Clipcap: Clip prefix for image captioning.
\newblock \emph{arXiv preprint arXiv:2111.09734}, 2021.

\bibitem[Mullender(1993)]{Mullender:1993:DS:302430}
Sape Mullender, editor.
\newblock \emph{Distributed systems (2nd Ed.)}.
\newblock ACM Press/Addison-Wesley Publishing Co., New York, NY, USA, 1993.
\newblock ISBN 0-201-62427-3.

\bibitem[Mumford(1987)]{Mumford:1987:MES:54905.54911}
E.~Mumford.
\newblock Managerial expert systems and organizational change: some critical research issues.
\newblock In \emph{Critical issues in information systems research (incoll)}, pages 135--155. John Wiley \& Sons, Inc., New York, NY, USA, 1987.
\newblock ISBN 0-471-91281-6.
\newblock URL \url{http://portal.acm.org/citation.cfm?id=54905.54911}.

\bibitem[Mun et~al.(2017)Mun, Hongsuck~Seo, Jung, and Han]{mun2017marioqa}
Jonghwan Mun, Paul Hongsuck~Seo, Ilchae Jung, and Bohyung Han.
\newblock Marioqa: Answering questions by watching gameplay videos.
\newblock In \emph{Proceedings of the IEEE International Conference on Computer Vision}, pages 2867--2875, 2017.

\bibitem[Na et~al.(2017)Na, Lee, Kim, and Kim]{na2017read}
Seil Na, Sangho Lee, Jisung Kim, and Gunhee Kim.
\newblock A read-write memory network for movie story understanding.
\newblock In \emph{Proceedings of the IEEE International Conference on Computer Vision}, pages 677--685, 2017.

\bibitem[Nair and Hinton(2010)]{nair2010rectified}
Vinod Nair and Geoffrey~E Hinton.
\newblock Rectified linear units improve restricted boltzmann machines.
\newblock In \emph{Icml}, 2010.

\bibitem[Nam et~al.(2017)Nam, Ha, and Kim]{nam2017dual}
Hyeonseob Nam, Jung-Woo Ha, and Jeonghee Kim.
\newblock Dual attention networks for multimodal reasoning and matching.
\newblock In \emph{Proceedings of the IEEE conference on computer vision and pattern recognition}, pages 299--307, 2017.

\bibitem[Narasimhan and Schwing(2018)]{narasimhan2018straight}
Medhini Narasimhan and Alexander~G Schwing.
\newblock Straight to the facts: Learning knowledge base retrieval for factual visual question answering.
\newblock In \emph{Proceedings of the European conference on computer vision (ECCV)}, pages 451--468, 2018.

\bibitem[Narasimhan et~al.(2018)Narasimhan, Lazebnik, and Schwing]{narasimhan2018out}
Medhini Narasimhan, Svetlana Lazebnik, and Alexander Schwing.
\newblock Out of the box: Reasoning with graph convolution nets for factual visual question answering.
\newblock \emph{Advances in neural information processing systems}, 31, 2018.

\bibitem[Natarajan et~al.(2007)Natarajan, Motani, de~Silva, Yap, and Chua]{Natarajan-01}
A.~Natarajan, M.~Motani, B.~de~Silva, K.~Yap, and K.~C. Chua.
\newblock Investigating network architectures for body sensor networks.
\newblock In G.~Whitcomb and P.~Neece, editors, \emph{Network Architectures}, pages 322--328, Dayton, OH, 2007. Keleuven Press.

\bibitem[Ngiam et~al.(2011)Ngiam, Khosla, Kim, Nam, Lee, and Ng]{ngiam2011multimodal}
Jiquan Ngiam, Aditya Khosla, Mingyu Kim, Juhan Nam, Honglak Lee, and Andrew~Y Ng.
\newblock Multimodal deep learning.
\newblock In \emph{ICML}, 2011.

\bibitem[Nguyen and Okatani(2018)]{nguyen2018improved}
Duy-Kien Nguyen and Takayuki Okatani.
\newblock Improved fusion of visual and language representations by dense symmetric co-attention for visual question answering.
\newblock In \emph{Proceedings of the IEEE conference on computer vision and pattern recognition}, pages 6087--6096, 2018.

\bibitem[Nielson(1985)]{7:3:359}
F.~Nielson.
\newblock Program transformations in a denotational setting.
\newblock \emph{ACM Trans. Program. Lang. Syst.}, 7\penalty0 (3):\penalty0 359--379, July 1985.

\bibitem[Noh and Han(2016)]{noh2016training}
Hyeonwoo Noh and Bohyung Han.
\newblock Training recurrent answering units with joint loss minimization for vqa.
\newblock \emph{arXiv preprint arXiv:1606.03647}, 2016.

\bibitem[Noh et~al.(2016)Noh, Seo, and Han]{noh2016image}
Hyeonwoo Noh, Paul~Hongsuck Seo, and Bohyung Han.
\newblock Image question answering using convolutional neural network with dynamic parameter prediction.
\newblock In \emph{Proceedings of the IEEE conference on computer vision and pattern recognition}, pages 30--38, 2016.

\bibitem[Norcliffe-Brown et~al.(2018)Norcliffe-Brown, Vafeias, and Parisot]{norcliffe2018learning}
Will Norcliffe-Brown, Stathis Vafeias, and Sarah Parisot.
\newblock Learning conditioned graph structures for interpretable visual question answering.
\newblock \emph{Advances in neural information processing systems}, 31, 2018.

\bibitem[Novak(2003)]{Novak03}
Dave Novak.
\newblock Solder man.
\newblock In \emph{ACM SIGGRAPH 2003 Video Review on Animation theater Program: Part I - Vol. 145 (July 27--27, 2003)}, page~4, New York, NY, March 21, 2008 2003. ACM Press.
\newblock \doi{99.9999/woot07-S422}.
\newblock URL \url{http://video.google.com/videoplay?docid=6528042696351994555}.

\bibitem[Nuthalapati et~al.(2021)Nuthalapati, Chandradevan, Giunchiglia, Li, Kayser, Lukasiewicz, and Yang]{nuthalapati2021lightweight}
Sai~Vidyaranya Nuthalapati, Ramraj Chandradevan, Eleonora Giunchiglia, Bowen Li, Maxime Kayser, Thomas Lukasiewicz, and Carl Yang.
\newblock Lightweight visual question answering using scene graphs.
\newblock In \emph{Proceedings of the 30th ACM International Conference on Information \& Knowledge Management}, pages 3353--3357, 2021.

\bibitem[Obama(2008)]{Obama08}
Barack Obama.
\newblock A more perfect union.
\newblock Video, March 2008.
\newblock URL \url{http://video.google.com/videoplay?docid=6528042696351994555}.

\bibitem[Oord et~al.(2018)Oord, Li, and Vinyals]{oord2018representation}
Aaron van~den Oord, Yazhe Li, and Oriol Vinyals.
\newblock Representation learning with contrastive predictive coding.
\newblock \emph{arXiv preprint arXiv:1807.03748}, 2018.

\bibitem[Pan et~al.(2022)Pan, Chen, Gong, Zhou, Wang, and Lin]{pan2022leveraging}
Xichen Pan, Peiyu Chen, Yichen Gong, Helong Zhou, Xinbing Wang, and Zhouhan Lin.
\newblock Leveraging uni-modal self-supervised learning for multimodal audio-visual speech recognition.
\newblock \emph{arXiv preprint arXiv:2203.07996}, 2022.

\bibitem[Panayotov et~al.(2015)Panayotov, Chen, Povey, and Khudanpur]{panayotov2015librispeech}
Vassil Panayotov, Guoguo Chen, Daniel Povey, and Sanjeev Khudanpur.
\newblock Librispeech: an asr corpus based on public domain audio books.
\newblock In \emph{2015 IEEE international conference on acoustics, speech and signal processing (ICASSP)}, pages 5206--5210. IEEE, 2015.

\bibitem[Papineni et~al.(2002)Papineni, Roukos, Ward, and Zhu]{papineni2002bleu}
Kishore Papineni, Salim Roukos, Todd Ward, and Wei-Jing Zhu.
\newblock Bleu: a method for automatic evaluation of machine translation.
\newblock In \emph{Proceedings of the 40th annual meeting of the Association for Computational Linguistics}, pages 311--318, 2002.

\bibitem[Parisi et~al.(2022)Parisi, Zhao, and Fiedel]{talm}
Aaron Parisi, Yao Zhao, and Noah Fiedel.
\newblock {TALM:} tool augmented language models.
\newblock \emph{CoRR}, abs/2205.12255, 2022.

\bibitem[Patel et~al.(2021)Patel, Parikh, and Shastri]{patel2021recent}
Devshree Patel, Ratnam Parikh, and Yesha Shastri.
\newblock Recent advances in video question answering: A review of datasets and methods.
\newblock In \emph{International Conference on Pattern Recognition}, pages 339--356. Springer, 2021.

\bibitem[Pennington et~al.(2014)Pennington, Socher, and Manning]{pennington2014glove}
Jeffrey Pennington, Richard Socher, and Christopher~D Manning.
\newblock Glove: Global vectors for word representation.
\newblock In \emph{Proceedings of the 2014 conference on empirical methods in natural language processing (EMNLP)}, pages 1532--1543, 2014.

\bibitem[Pereira~J{\'u}nior et~al.(2024)Pereira~J{\'u}nior, Rodrigues, Costa, Macario~Filho, and Mello]{pereira2024can}
Cleon Pereira~J{\'u}nior, Luiz Rodrigues, Newarney Costa, Valmir Macario~Filho, and Rafael Mello.
\newblock Can vlm understand children’s handwriting? an analysis on handwritten mathematical equation recognition.
\newblock In \emph{International Conference on Artificial Intelligence in Education}, pages 321--328. Springer, 2024.

\bibitem[Perozzi et~al.(2014)Perozzi, Al-Rfou, and Skiena]{perozzi2014deepwalk}
Bryan Perozzi, Rami Al-Rfou, and Steven Skiena.
\newblock Deepwalk: Online learning of social representations.
\newblock In \emph{Proceedings of the 20th ACM SIGKDD international conference on Knowledge discovery and data mining}, pages 701--710, 2014.

\bibitem[Perruchet and Vinter(1998)]{perruchet1998parser}
Pierre Perruchet and Annie Vinter.
\newblock Parser: A model for word segmentation.
\newblock \emph{Journal of memory and language}, 39\penalty0 (2):\penalty0 246--263, 1998.

\bibitem[Petrie(1986{\natexlab{a}})]{Petrie:1986:NAD:12345}
Charles~J. Petrie.
\newblock New algorithms for dependency-directed backtracking (master's thesis).
\newblock Master's thesis, University of Texas at Austin, Austin, TX, USA, 1986{\natexlab{a}}.

\bibitem[Petrie(1986{\natexlab{b}})]{Petrie:1986:NAD:899644}
Charles~J. Petrie.
\newblock New algorithms for dependency-directed backtracking (master's thesis).
\newblock Technical report, Austin, TX, USA, 1986{\natexlab{b}}.

\bibitem[Pezeshkpour et~al.(2018)Pezeshkpour, Chen, and Singh]{pezeshkpour2018embedding}
Pouya Pezeshkpour, Liyan Chen, and Sameer Singh.
\newblock Embedding multimodal relational data for knowledge base completion.
\newblock In \emph{Proceedings of the 2018 Conference on Empirical Methods in Natural Language Processing}, pages 3208--3218, Brussels, Belgium, October-November 2018. Association for Computational Linguistics.
\newblock \doi{10.18653/v1/D18-1359}.
\newblock URL \url{https://aclanthology.org/D18-1359}.

\bibitem[Plummer et~al.(2015)Plummer, Wang, Cervantes, Caicedo, Hockenmaier, and Lazebnik]{plummer2015flickr30k}
Bryan~A Plummer, Liwei Wang, Chris~M Cervantes, Juan~C Caicedo, Julia Hockenmaier, and Svetlana Lazebnik.
\newblock Flickr30k entities: Collecting region-to-phrase correspondences for richer image-to-sentence models.
\newblock In \emph{Proceedings of the IEEE international conference on computer vision}, pages 2641--2649, 2015.

\bibitem[Poker-Edge.Com(2006)]{Poker06}
Poker-Edge.Com.
\newblock Stats and analysis, March 2006.
\newblock URL \url{http://www.poker-edge.com/stats.php}.

\bibitem[Povey et~al.(2011)Povey, Ghoshal, Boulianne, Burget, Glembek, Goel, Hannemann, Motlicek, Qian, Schwarz, et~al.]{povey2011kaldi}
Daniel Povey, Arnab Ghoshal, Gilles Boulianne, Lukas Burget, Ondrej Glembek, Nagendra Goel, Mirko Hannemann, Petr Motlicek, Yanmin Qian, Petr Schwarz, et~al.
\newblock The kaldi speech recognition toolkit.
\newblock In \emph{IEEE 2011 workshop on automatic speech recognition and understanding}, number CONF. IEEE Signal Processing Society, 2011.

\bibitem[Pujara et~al.(2013)Pujara, Miao, Getoor, and Cohen]{pujara2013knowledge}
Jay Pujara, Hui Miao, Lise Getoor, and William Cohen.
\newblock Knowledge graph identification.
\newblock In \emph{International semantic web conference}, pages 542--557. Springer, 2013.

\bibitem[Qi et~al.(2020)Qi, Su, Song, Cui, Bharti, and Sacheti]{qi2020imagebert}
Di~Qi, Lin Su, Jia Song, Edward Cui, Taroon Bharti, and Arun Sacheti.
\newblock Imagebert: Cross-modal pre-training with large-scale weak-supervised image-text data.
\newblock \emph{arXiv preprint arXiv:2001.07966}, 2020.

\bibitem[Qian et~al.(2022)Qian, Hu, Wang, Feng, and Wang]{qian2022question}
Yuxi Qian, Yuncong Hu, Ruonan Wang, Fangxiang Feng, and Xiaojie Wang.
\newblock Question-driven graph fusion network for visual question answering.
\newblock \emph{arXiv preprint arXiv:2204.00975}, 2022.

\bibitem[Qu et~al.(2021)Qu, Zamani, Yang, Croft, and Learned-Miller]{qu2021passage}
Chen Qu, Hamed Zamani, Liu Yang, W~Bruce Croft, and Erik Learned-Miller.
\newblock Passage retrieval for outside-knowledge visual question answering.
\newblock In \emph{Proceedings of the 44th International ACM SIGIR Conference on Research and Development in Information Retrieval}, pages 1753--1757, 2021.

\bibitem[Quek et~al.(2002)Quek, McNeill, Bryll, Duncan, Ma, Kirbas, McCullough, and Ansari]{quek2002multimodal}
Francis Quek, David McNeill, Robert Bryll, Susan Duncan, Xin-Feng Ma, Cemil Kirbas, Karl~E McCullough, and Rashid Ansari.
\newblock Multimodal human discourse: gesture and speech.
\newblock \emph{ACM Transactions on Computer-Human Interaction (TOCHI)}, 9\penalty0 (3):\penalty0 171--193, 2002.

\bibitem[{R Core Team}(2019)]{R}
{R Core Team}.
\newblock R: A language and environment for statistical computing, 2019.
\newblock URL \url{https://www.R-project.org/}.

\bibitem[Radford et~al.(2021)Radford, Kim, Hallacy, Ramesh, Goh, Agarwal, Sastry, Askell, Mishkin, Clark, et~al.]{radford2021learning}
Alec Radford, Jong~Wook Kim, Chris Hallacy, Aditya Ramesh, Gabriel Goh, Sandhini Agarwal, Girish Sastry, Amanda Askell, Pamela Mishkin, Jack Clark, et~al.
\newblock Learning transferable visual models from natural language supervision.
\newblock In \emph{International Conference on Machine Learning}, pages 8748--8763. PMLR, 2021.

\bibitem[Ramnath and Hasegawa-Johnson(2020)]{ramnath2020seeing}
Kiran Ramnath and Mark Hasegawa-Johnson.
\newblock Seeing is knowing! fact-based visual question answering using knowledge graph embeddings.
\newblock \emph{arXiv preprint arXiv:2012.15484}, 2020.

\bibitem[Reid(1980)]{reid:scribe}
Brian~K. Reid.
\newblock A high-level approach to computer document formatting.
\newblock In \emph{Proceedings of the 7th Annual Symposium on Principles of Programming Languages}, pages 24--31, New York, January 1980. ACM.

\bibitem[Ren et~al.(2015{\natexlab{a}})Ren, Kiros, and Zemel]{ren2015exploring}
Mengye Ren, Ryan Kiros, and Richard Zemel.
\newblock Exploring models and data for image question answering.
\newblock \emph{Advances in neural information processing systems}, 28, 2015{\natexlab{a}}.

\bibitem[Ren et~al.(2015{\natexlab{b}})Ren, Kiros, and Zemel]{ren2015image}
Mengye Ren, Ryan Kiros, and Richard Zemel.
\newblock Image question answering: A visual semantic embedding model and a new dataset.
\newblock \emph{Proc. Advances in Neural Inf. Process. Syst}, 1\penalty0 (2):\penalty0 5, 2015{\natexlab{b}}.

\bibitem[Ren et~al.(2015{\natexlab{c}})Ren, He, Girshick, and Sun]{ren2015faster}
Shaoqing Ren, Kaiming He, Ross Girshick, and Jian Sun.
\newblock Faster r-cnn: Towards real-time object detection with region proposal networks.
\newblock \emph{Advances in neural information processing systems}, 28, 2015{\natexlab{c}}.

\bibitem[Reza(1994)]{reza1994introduction}
Fazlollah~M Reza.
\newblock \emph{An introduction to information theory}.
\newblock Courier Corporation, 1994.

\bibitem[Roberts et~al.(2023)Roberts, L{\"u}ddecke, Sheikh, Han, and Albanie]{roberts2023charting}
Jonathan Roberts, Timo L{\"u}ddecke, Rehan Sheikh, Kai Han, and Samuel Albanie.
\newblock Charting new territories: Exploring the geographic and geospatial capabilities of multimodal llms.
\newblock \emph{arXiv preprint arXiv:2311.14656}, 2023.

\bibitem[Robin and Lacroix(2016)]{robin2016multi}
Cyril Robin and Simon Lacroix.
\newblock Multi-robot target detection and tracking: taxonomy and survey.
\newblock \emph{Autonomous Robots}, 40\penalty0 (4):\penalty0 729--760, 2016.

\bibitem[Rose et~al.(2023)Rose, Himakunthala, Ouyang, He, Mei, Lu, Saxon, Sonar, Mirza, and Wang]{DBLP:journals/corr/abs-2305-02317}
Daniel Rose, Vaishnavi Himakunthala, Andy Ouyang, Ryan He, Alex Mei, Yujie Lu, Michael Saxon, Chinmay Sonar, Diba Mirza, and William~Yang Wang.
\newblock Visual chain of thought: Bridging logical gaps with multimodal infillings.
\newblock \emph{CoRR}, abs/2305.02317, 2023.

\bibitem[Ross(1989)]{ross1989signal}
Elliott~M Ross.
\newblock Signal sorting and amplification through g protein-coupled receptors.
\newblock \emph{Neuron}, 3\penalty0 (2):\penalty0 141--152, 1989.

\bibitem[Rosten et~al.(2008)Rosten, Porter, and Drummond]{rosten2008faster}
Edward Rosten, Reid Porter, and Tom Drummond.
\newblock Faster and better: A machine learning approach to corner detection.
\newblock \emph{IEEE transactions on pattern analysis and machine intelligence}, 32\penalty0 (1):\penalty0 105--119, 2008.

\bibitem[Rous(2008)]{rous08}
Bernard Rous.
\newblock The enabling of digital libraries.
\newblock \emph{Digital Libraries}, 12\penalty0 (3), July 2008.
\newblock To appear.

\bibitem[Rusu et~al.(2013)Rusu, Halcu, Grigoriu, Neculoiu, Sandulescu, Marinescu, and Marinescu]{rusu2013converting}
Octavian Rusu, Ionela Halcu, Oana Grigoriu, Giorgian Neculoiu, Virginia Sandulescu, Mariana Marinescu, and Viorel Marinescu.
\newblock Converting unstructured and semi-structured data into knowledge.
\newblock In \emph{2013 11th RoEduNet International Conference}, pages 1--4. IEEE, 2013.

\bibitem[Ruwa et~al.(2019)Ruwa, Mao, Wang, Gou, and Dong]{ruwa2019mood}
Nelson Ruwa, Qirong Mao, Liangjun Wang, Jianping Gou, and Ming Dong.
\newblock Mood-aware visual question answering.
\newblock \emph{Neurocomputing}, 330:\penalty0 305--316, 2019.

\bibitem[Saeedi et~al.(2010{\natexlab{a}})Saeedi, Zamani, and Sedighi]{SaeediMEJ10}
Mehdi Saeedi, Morteza~Saheb Zamani, and Mehdi Sedighi.
\newblock A library-based synthesis methodology for reversible logic.
\newblock \emph{Microelectron. J.}, 41\penalty0 (4):\penalty0 185--194, April 2010{\natexlab{a}}.

\bibitem[Saeedi et~al.(2010{\natexlab{b}})Saeedi, Zamani, Sedighi, and Sasanian]{SaeediJETC10}
Mehdi Saeedi, Morteza~Saheb Zamani, Mehdi Sedighi, and Zahra Sasanian.
\newblock Synthesis of reversible circuit using cycle-based approach.
\newblock \emph{J. Emerg. Technol. Comput. Syst.}, 6\penalty0 (4), December 2010{\natexlab{b}}.

\bibitem[Saito et~al.(2017)Saito, Shin, Ushiku, and Harada]{saito2017dualnet}
Kuniaki Saito, Andrew Shin, Yoshitaka Ushiku, and Tatsuya Harada.
\newblock Dualnet: Domain-invariant network for visual question answering.
\newblock In \emph{2017 IEEE International Conference on Multimedia and Expo (ICME)}, pages 829--834. IEEE, 2017.

\bibitem[Salas and Hille(1978)]{salas:calculus}
S.L. Salas and Einar Hille.
\newblock \emph{Calculus: One and Several Variable}.
\newblock John Wiley and Sons, New York, 1978.

\bibitem[Sanh et~al.(2022)Sanh, Webson, Raffel, Bach, Sutawika, Alyafeai, Chaffin, Stiegler, Le~Scao, Raja, et~al.]{sanh2022multitask}
Victor Sanh, Albert Webson, Colin Raffel, Stephen~H Bach, Lintang Sutawika, Zaid Alyafeai, Antoine Chaffin, Arnaud Stiegler, Teven Le~Scao, Arun Raja, et~al.
\newblock Multitask prompted training enables zero-shot task generalization.
\newblock In \emph{ICLR 2022-Tenth International Conference on Learning Representations}, 2022.

\bibitem[Santoro et~al.(2016)Santoro, Bartunov, Botvinick, Wierstra, and Lillicrap]{santoro2016meta}
Adam Santoro, Sergey Bartunov, Matthew Botvinick, Daan Wierstra, and Timothy Lillicrap.
\newblock Meta-learning with memory-augmented neural networks.
\newblock In \emph{International conference on machine learning}, pages 1842--1850. PMLR, 2016.

\bibitem[Schlichtkrull et~al.(2018)Schlichtkrull, Kipf, Bloem, Berg, Titov, and Welling]{schlichtkrull2018modeling}
Michael Schlichtkrull, Thomas~N Kipf, Peter Bloem, Rianne van~den Berg, Ivan Titov, and Max Welling.
\newblock Modeling relational data with graph convolutional networks.
\newblock In \emph{European semantic web conference}, pages 593--607. Springer, 2018.

\bibitem[Schwartz et~al.(2017)Schwartz, Schwing, and Hazan]{schwartz2017high}
Idan Schwartz, Alexander Schwing, and Tamir Hazan.
\newblock High-order attention models for visual question answering.
\newblock \emph{Advances in Neural Information Processing Systems}, 30, 2017.

\bibitem[Scientist(2009)]{JoeScientist001}
Joseph Scientist.
\newblock The fountain of youth, August 2009.
\newblock Patent No. 12345, Filed July 1st., 2008, Issued Aug. 9th., 2009.

\bibitem[Shah et~al.(2019{\natexlab{a}})Shah, Chen, Rohrbach, and Parikh]{shah2019cycle}
Meet Shah, Xinlei Chen, Marcus Rohrbach, and Devi Parikh.
\newblock Cycle-consistency for robust visual question answering.
\newblock In \emph{Proceedings of the IEEE/CVF Conference on Computer Vision and Pattern Recognition}, pages 6649--6658, 2019{\natexlab{a}}.

\bibitem[Shah et~al.(2019{\natexlab{b}})Shah, Mishra, Yadati, and Talukdar]{shah2019kvqa}
Sanket Shah, Anand Mishra, Naganand Yadati, and Partha~Pratim Talukdar.
\newblock Kvqa: Knowledge-aware visual question answering.
\newblock In \emph{Proceedings of the AAAI conference on artificial intelligence}, volume~33, pages 8876--8884, 2019{\natexlab{b}}.

\bibitem[Shao et~al.(2023)Shao, Yu, Wang, and Yu]{shao2023prompting}
Zhenwei Shao, Zhou Yu, Meng Wang, and Jun Yu.
\newblock Prompting large language models with answer heuristics for knowledge-based visual question answering.
\newblock In \emph{Proceedings of the IEEE/CVF Conference on Computer Vision and Pattern Recognition}, pages 14974--14983, 2023.

\bibitem[Shapiro et~al.(2001)Shapiro, Stockman, et~al.]{shapiro2001computer}
Linda~G Shapiro, George~C Stockman, et~al.
\newblock \emph{Computer vision}, volume~3.
\newblock Prentice Hall New Jersey, 2001.

\bibitem[Sharma et~al.(2018)Sharma, Ding, Goodman, and Soricut]{sharma2018conceptual}
Piyush Sharma, Nan Ding, Sebastian Goodman, and Radu Soricut.
\newblock Conceptual captions: A cleaned, hypernymed, image alt-text dataset for automatic image captioning.
\newblock In \emph{Proceedings of the 56th Annual Meeting of the Association for Computational Linguistics (Volume 1: Long Papers)}, pages 2556--2565, 2018.

\bibitem[Shen et~al.(2021{\natexlab{a}})Shen, Li, Tan, Bansal, Rohrbach, Chang, Yao, and Keutzer]{shen2021much}
Sheng Shen, Liunian~Harold Li, Hao Tan, Mohit Bansal, Anna Rohrbach, Kai-Wei Chang, Zhewei Yao, and Kurt Keutzer.
\newblock How much can clip benefit vision-and-language tasks?
\newblock In \emph{International Conference on Learning Representations}, 2021{\natexlab{a}}.

\bibitem[Shen et~al.(2020)Shen, Ding, Zheng, Li, and Yang]{shen2020modeling}
Ying Shen, Ning Ding, Hai-Tao Zheng, Yaliang Li, and Min Yang.
\newblock Modeling relation paths for knowledge graph completion.
\newblock \emph{IEEE Transactions on Knowledge and Data Engineering}, 33\penalty0 (11):\penalty0 3607--3617, 2020.

\bibitem[Shen et~al.(2021{\natexlab{b}})Shen, Yang, Li, Wang, Zheng, and Chen]{shen2021knowledge}
Ying Shen, Min Yang, Yaliang Li, Dong Wang, Haitao Zheng, and Daoyuan Chen.
\newblock Knowledge-based reasoning network for relation detection.
\newblock \emph{IEEE Transactions on Neural Networks and Learning Systems}, 2021{\natexlab{b}}.

\bibitem[Shen et~al.(2024)Shen, Song, Tan, Li, Lu, and Zhuang]{shen2024hugginggpt}
Yongliang Shen, Kaitao Song, Xu~Tan, Dongsheng Li, Weiming Lu, and Yueting Zhuang.
\newblock Hugginggpt: Solving ai tasks with chatgpt and its friends in hugging face.
\newblock \emph{Advances in Neural Information Processing Systems}, 36, 2024.

\bibitem[Shenoy and Sardana(2020)]{shenoy2020multilogue}
Aman Shenoy and Ashish Sardana.
\newblock Multilogue-net: A context-aware {RNN} for multi-modal emotion detection and sentiment analysis in conversation.
\newblock In \emph{Second Grand-Challenge and Workshop on Multimodal Language (Challenge-HML)}, pages 19--28, Seattle, USA, July 2020. Association for Computational Linguistics.
\newblock \doi{10.18653/v1/2020.challengehml-1.3}.
\newblock URL \url{https://aclanthology.org/2020.challengehml-1.3}.

\bibitem[Shi et~al.(2022)Shi, Hsu, and Mohamed]{shi2022robust}
Bowen Shi, Wei-Ning Hsu, and Abdelrahman Mohamed.
\newblock Robust self-supervised audio-visual speech recognition.
\newblock \emph{arXiv preprint arXiv:2201.01763}, 2022.

\bibitem[Shi et~al.(2019)Shi, Zhang, and Li]{shi2019explainable}
Jiaxin Shi, Hanwang Zhang, and Juanzi Li.
\newblock Explainable and explicit visual reasoning over scene graphs.
\newblock In \emph{Proceedings of the IEEE/CVF Conference on Computer Vision and Pattern Recognition}, pages 8376--8384, 2019.

\bibitem[Shi et~al.(2018)Shi, Furlanello, Zha, and Anandkumar]{shi2018question}
Yang Shi, Tommaso Furlanello, Sheng Zha, and Animashree Anandkumar.
\newblock Question type guided attention in visual question answering.
\newblock In \emph{Proceedings of the European Conference on Computer Vision (ECCV)}, pages 151--166, 2018.

\bibitem[Shibata et~al.(1999)Shibata, Kida, Fukamachi, Takeda, Shinohara, Shinohara, and Arikawa]{shibata1999byte}
Yusuxke Shibata, Takuya Kida, Shuichi Fukamachi, Masayuki Takeda, Ayumi Shinohara, Takeshi Shinohara, and Setsuo Arikawa.
\newblock Byte pair encoding: A text compression scheme that accelerates pattern matching.
\newblock 1999.

\bibitem[Shih et~al.(2016)Shih, Singh, and Hoiem]{shih2016look}
Kevin~J Shih, Saurabh Singh, and Derek Hoiem.
\newblock Where to look: Focus regions for visual question answering.
\newblock In \emph{Proceedings of the IEEE conference on computer vision and pattern recognition}, pages 4613--4621, 2016.

\bibitem[Shukor et~al.(2023)Shukor, Rame, Dancette, and Cord]{shukor2023beyond}
Mustafa Shukor, Alexandre Rame, Corentin Dancette, and Matthieu Cord.
\newblock Beyond task performance: Evaluating and reducing the flaws of large multimodal models with in-context learning.
\newblock \emph{arXiv preprint arXiv:2310.00647}, 2023.

\bibitem[Silberman et~al.(2012)Silberman, Hoiem, Kohli, and Fergus]{silberman2012indoor}
Nathan Silberman, Derek Hoiem, Pushmeet Kohli, and Rob Fergus.
\newblock Indoor segmentation and support inference from rgbd images.
\newblock In \emph{European conference on computer vision}, pages 746--760. Springer, 2012.

\bibitem[Simonyan and Zisserman(2014)]{simonyan2014very}
Karen Simonyan and Andrew Zisserman.
\newblock Very deep convolutional networks for large-scale image recognition.
\newblock \emph{arXiv preprint arXiv:1409.1556}, 2014.

\bibitem[Singh et~al.(2019)Singh, Natarajan, Shah, Jiang, Chen, Batra, Parikh, and Rohrbach]{singh2019towards}
Amanpreet Singh, Vivek Natarajan, Meet Shah, Yu~Jiang, Xinlei Chen, Dhruv Batra, Devi Parikh, and Marcus Rohrbach.
\newblock Towards vqa models that can read.
\newblock In \emph{Proceedings of the IEEE/CVF conference on computer vision and pattern recognition}, pages 8317--8326, 2019.

\bibitem[Singh et~al.(2021)Singh, Akrigg, Di~Maio, Fontana, Alitappeh, Saha, Jeddisaravi, Yousefi, Culley, Nicholson, et~al.]{singh2021road}
Gurkirt Singh, Stephen Akrigg, Manuele Di~Maio, Valentina Fontana, Reza~Javanmard Alitappeh, Suman Saha, Kossar Jeddisaravi, Farzad Yousefi, Jacob Culley, Tom Nicholson, et~al.
\newblock Road: The road event awareness dataset for autonomous driving.
\newblock \emph{arXiv preprint arXiv:2102.11585}, 2021.

\bibitem[Singh et~al.(2023)Singh, Bathla, Mehta, Chhabra, and Singh]{10104870}
Harsimran~Jit Singh, Gourav Bathla, Munish Mehta, Gunjan Chhabra, and Pardeep Singh.
\newblock Visual questions answering developments, applications, datasets and opportunities: A state-of-the-art survey.
\newblock In \emph{2023 International Conference on Sustainable Computing and Data Communication Systems (ICSCDS)}, pages 778--785, 2023.

\bibitem[Singh et~al.(2018)Singh, Ying, and Nutkiewicz]{singh2018attention}
Jasdeep Singh, Vincent Ying, and Alex Nutkiewicz.
\newblock Attention on attention: Architectures for visual question answering (vqa).
\newblock \emph{arXiv preprint arXiv:1803.07724}, 2018.

\bibitem[Singhal et~al.(2001)]{singhal2001modern}
Amit Singhal et~al.
\newblock Modern information retrieval: A brief overview.
\newblock \emph{IEEE Data Eng. Bull.}, 24\penalty0 (4):\penalty0 35--43, 2001.

\bibitem[Smith(2010)]{Smith10}
Stan~W. Smith.
\newblock An experiment in bibliographic mark-up: Parsing metadata for xml export.
\newblock In Reginald~N. Smythe and Alexander Noble, editors, \emph{Proceedings of the 3rd. annual workshop on Librarians and Computers}, volume~3 of \emph{LAC '10}, pages 422--431, Milan Italy, 2010. Paparazzi Press.
\newblock \doi{99.9999/woot07-S422}.
\newblock URL \url{http://dx.doi.org/99.0000/woot07-S422}.

\bibitem[Son~Chung et~al.(2017)Son~Chung, Senior, Vinyals, and Zisserman]{son2017lip}
Joon Son~Chung, Andrew Senior, Oriol Vinyals, and Andrew Zisserman.
\newblock Lip reading sentences in the wild.
\newblock In \emph{Proceedings of the IEEE conference on computer vision and pattern recognition}, pages 6447--6456, 2017.

\bibitem[Song et~al.(2018)Song, Zeng, Gao, and Shen]{song2022pixels}
J.~Song, P.~Zeng, L.~Gao, and H.~T. Shen.
\newblock From pixels to objects: Cubic visual attention for visual question answering.
\newblock In \emph{Twenty-Seventh International Joint Conference on Artificial Intelligence {IJCAI-18}}, 2018.

\bibitem[Spector(1990)]{Spector90}
Asad~Z. Spector.
\newblock Achieving application requirements.
\newblock In Sape Mullender, editor, \emph{Distributed Systems}, pages 19--33. ACM Press, New York, NY, 2nd. edition, 1990.
\newblock \doi{10.1145/90417.90738}.
\newblock URL \url{http://doi.acm.org/10.1145/90417.90738}.

\bibitem[Srivastava et~al.(2021)Srivastava, Murali, Dubey, and Mukherjee]{srivastava2021visual}
Yash Srivastava, Vaishnav Murali, Shiv~Ram Dubey, and Snehasis Mukherjee.
\newblock Visual question answering using deep learning: A survey and performance analysis.
\newblock In \emph{International Conference on Computer Vision and Image Processing}, pages 75--86. Springer, 2021.

\bibitem[Stefanini et~al.(2022)Stefanini, Cornia, Baraldi, Cascianelli, Fiameni, and Cucchiara]{stefanini2022show}
Matteo Stefanini, Marcella Cornia, Lorenzo Baraldi, Silvia Cascianelli, Giuseppe Fiameni, and Rita Cucchiara.
\newblock From show to tell: A survey on deep learning-based image captioning.
\newblock \emph{IEEE transactions on pattern analysis and machine intelligence}, 45\penalty0 (1):\penalty0 539--559, 2022.

\bibitem[Su et~al.(2020)Su, Zhu, Cao, Li, Lu, Wei, and Dai~J]{su2020pre}
W~Su, X~Zhu, Y~Cao, B~Li, L~Lu, F~Wei, and VL-BERT Dai~J.
\newblock Pre-training of generic visual-linguistic representations.
\newblock In \emph{Proceedings of the 8th International Conference on Learning Representations}, pages 1--14, 2020.

\bibitem[Su et~al.(2019)Su, Zhu, Cao, Li, Lu, Wei, and Dai]{su2019vl}
Weijie Su, Xizhou Zhu, Yue Cao, Bin Li, Lewei Lu, Furu Wei, and Jifeng Dai.
\newblock Vl-bert: Pre-training of generic visual-linguistic representations.
\newblock \emph{arXiv preprint arXiv:1908.08530}, 2019.

\bibitem[Su et~al.(2018)Su, Zhu, Dong, Cai, Chen, and Li]{su2018learning}
Zhou Su, Chen Zhu, Yinpeng Dong, Dongqi Cai, Yurong Chen, and Jianguo Li.
\newblock Learning visual knowledge memory networks for visual question answering.
\newblock In \emph{Proceedings of the IEEE conference on computer vision and pattern recognition}, pages 7736--7745, 2018.

\bibitem[Sukhbaatar et~al.(2015)Sukhbaatar, Weston, Fergus, et~al.]{sukhbaatar2015end}
Sainbayar Sukhbaatar, Jason Weston, Rob Fergus, et~al.
\newblock End-to-end memory networks.
\newblock \emph{Advances in neural information processing systems}, 28, 2015.

\bibitem[Sun et~al.(2020)Sun, Cao, Zhao, Wan, Zhou, Zhang, Wang, and Zheng]{sun2020multi}
Rui Sun, Xuezhi Cao, Yan Zhao, Junchen Wan, Kun Zhou, Fuzheng Zhang, Zhongyuan Wang, and Kai Zheng.
\newblock Multi-modal knowledge graphs for recommender systems.
\newblock In \emph{Proceedings of the 29th ACM international conference on information \& knowledge management}, pages 1405--1414, 2020.

\bibitem[Sutskever et~al.(2014)Sutskever, Vinyals, and Le]{sutskever2014sequence}
Ilya Sutskever, Oriol Vinyals, and Quoc~V Le.
\newblock Sequence to sequence learning with neural networks.
\newblock \emph{Advances in neural information processing systems}, 27, 2014.

\bibitem[Szegedy et~al.(2015)Szegedy, Liu, Jia, Sermanet, Reed, Anguelov, Erhan, Vanhoucke, and Rabinovich]{szegedy2015going}
Christian Szegedy, Wei Liu, Yangqing Jia, Pierre Sermanet, Scott Reed, Dragomir Anguelov, Dumitru Erhan, Vincent Vanhoucke, and Andrew Rabinovich.
\newblock Going deeper with convolutions.
\newblock In \emph{Proceedings of the IEEE conference on computer vision and pattern recognition}, pages 1--9, 2015.

\bibitem[Tan et~al.(2020)Tan, Liu, Wang, and Zha]{tan2020learning}
Ganchao Tan, Daqing Liu, Meng Wang, and Zheng-Jun Zha.
\newblock Learning to discretely compose reasoning module networks for video captioning.
\newblock \emph{arXiv preprint arXiv:2007.09049}, 2020.

\bibitem[Tan and Bansal(2019)]{tan2019lxmert}
Hao Tan and Mohit Bansal.
\newblock Lxmert: Learning cross-modality encoder representations from transformers.
\newblock \emph{arXiv preprint arXiv:1908.07490}, 2019.

\bibitem[Tang et~al.(2019)Tang, Zhang, Wu, Luo, and Liu]{tang2019learning}
Kaihua Tang, Hanwang Zhang, Baoyuan Wu, Wenhan Luo, and Wei Liu.
\newblock Learning to compose dynamic tree structures for visual contexts.
\newblock In \emph{Proceedings of the IEEE/CVF conference on computer vision and pattern recognition}, pages 6619--6628, 2019.

\bibitem[Tapaswi et~al.(2016)Tapaswi, Zhu, Stiefelhagen, Torralba, Urtasun, and Fidler]{tapaswi2016movieqa}
Makarand Tapaswi, Yukun Zhu, Rainer Stiefelhagen, Antonio Torralba, Raquel Urtasun, and Sanja Fidler.
\newblock Movieqa: Understanding stories in movies through question-answering.
\newblock In \emph{Proceedings of the IEEE conference on computer vision and pattern recognition}, pages 4631--4640, 2016.

\bibitem[Team et~al.(2023)Team, Anil, Borgeaud, Wu, Alayrac, Yu, Soricut, Schalkwyk, Dai, Hauth, et~al.]{team2023gemini}
Gemini Team, Rohan Anil, Sebastian Borgeaud, Yonghui Wu, Jean-Baptiste Alayrac, Jiahui Yu, Radu Soricut, Johan Schalkwyk, Andrew~M Dai, Anja Hauth, et~al.
\newblock Gemini: a family of highly capable multimodal models.
\newblock \emph{arXiv preprint arXiv:2312.11805}, 2023.

\bibitem[Teney et~al.(2017)Teney, Liu, and van Den~Hengel]{teney2017graph}
Damien Teney, Lingqiao Liu, and Anton van Den~Hengel.
\newblock Graph-structured representations for visual question answering.
\newblock In \emph{Proceedings of the IEEE conference on computer vision and pattern recognition}, pages 1--9, 2017.

\bibitem[Teney et~al.(2018)Teney, Anderson, He, and Van Den~Hengel]{teney2018tips}
Damien Teney, Peter Anderson, Xiaodong He, and Anton Van Den~Hengel.
\newblock Tips and tricks for visual question answering: Learnings from the 2017 challenge.
\newblock In \emph{Proceedings of the IEEE conference on computer vision and pattern recognition}, pages 4223--4232, 2018.

\bibitem[Thomas and Kovashka(2022)]{thomas2022emphasizing}
Christopher Thomas and Adriana Kovashka.
\newblock Emphasizing complementary samples for non-literal cross-modal retrieval.
\newblock In \emph{Proceedings of the IEEE/CVF Conference on Computer Vision and Pattern Recognition}, pages 4632--4641, 2022.

\bibitem[Thomee et~al.(2016)Thomee, Shamma, Friedland, Elizalde, Ni, Poland, Borth, and Li]{thomee2016yfcc100m}
Bart Thomee, David~A Shamma, Gerald Friedland, Benjamin Elizalde, Karl Ni, Douglas Poland, Damian Borth, and Li-Jia Li.
\newblock Yfcc100m: The new data in multimedia research.
\newblock \emph{Communications of the ACM}, 59\penalty0 (2):\penalty0 64--73, 2016.

\bibitem[Thornburg(2001)]{Thornburg01}
Harry Thornburg.
\newblock Introduction to bayesian statistics, March 2001.
\newblock URL \url{http://ccrma.stanford.edu/~jos/bayes/bayes.html}.

\bibitem[Tiong et~al.(2022)Tiong, Li, Li, Savarese, and Hoi]{DBLP:conf/emnlp/Tiong0LSH22}
Anthony Meng~Huat Tiong, Junnan Li, Boyang Li, Silvio Savarese, and Steven C.~H. Hoi.
\newblock Plug-and-play {VQA:} zero-shot {VQA} by conjoining large pretrained models with zero training.
\newblock In Yoav Goldberg, Zornitsa Kozareva, and Yue Zhang, editors, \emph{Findings of the Association for Computational Linguistics: {EMNLP} 2022, Abu Dhabi, United Arab Emirates, December 7-11, 2022}, pages 951--967. Association for Computational Linguistics, 2022.

\bibitem[Tran et~al.(2015)Tran, Bourdev, Fergus, Torresani, and Paluri]{tran2015learning}
Du~Tran, Lubomir Bourdev, Rob Fergus, Lorenzo Torresani, and Manohar Paluri.
\newblock Learning spatiotemporal features with 3d convolutional networks.
\newblock In \emph{Proceedings of the IEEE international conference on computer vision}, pages 4489--4497, 2015.

\bibitem[Tsanas et~al.(2012)Tsanas, Little, McSharry, Spielman, and Ramig]{tsanas2012novel}
Athanasios Tsanas, Max~A Little, Patrick~E McSharry, Jennifer Spielman, and Lorraine~O Ramig.
\newblock Novel speech signal processing algorithms for high-accuracy classification of parkinson's disease.
\newblock \emph{IEEE transactions on biomedical engineering}, 59\penalty0 (5):\penalty0 1264--1271, 2012.

\bibitem[Tsimpoukelli et~al.(2021)Tsimpoukelli, Menick, Cabi, Eslami, Vinyals, and Hill]{tsimpoukelli2021multimodal}
Maria Tsimpoukelli, Jacob~L Menick, Serkan Cabi, SM~Eslami, Oriol Vinyals, and Felix Hill.
\newblock Multimodal few-shot learning with frozen language models.
\newblock \emph{Advances in Neural Information Processing Systems}, 34:\penalty0 200--212, 2021.

\bibitem[TUG()]{TUGInstmem}
TUG.
\newblock Institutional members of the {\TeX} users group, 2017.
\newblock URL \url{http://wwtug.org/instmem.html}.

\bibitem[Turk(2014)]{turk2014multimodal}
Matthew Turk.
\newblock Multimodal interaction: A review.
\newblock \emph{Pattern recognition letters}, 36:\penalty0 189--195, 2014.

\bibitem[Tzamaloukas and Garcia-Luna-Aceves(2000)]{Tzamaloukas-01}
A.~Tzamaloukas and J.~J. Garcia-Luna-Aceves.
\newblock Channel-hopping multiple access.
\newblock Technical Report I-CA2301, Department of Computer Science, University of California, Berkeley, CA, 2000.

\bibitem[Uijlings et~al.(2013)Uijlings, Van De~Sande, Gevers, and Smeulders]{uijlings2013selective}
Jasper~RR Uijlings, Koen~EA Van De~Sande, Theo Gevers, and Arnold~WM Smeulders.
\newblock Selective search for object recognition.
\newblock \emph{International journal of computer vision}, 104\penalty0 (2):\penalty0 154--171, 2013.

\bibitem[Vaswani et~al.(2017)Vaswani, Shazeer, Parmar, Uszkoreit, Jones, Gomez, Kaiser, and Polosukhin]{vaswani2017attention}
Ashish Vaswani, Noam Shazeer, Niki Parmar, Jakob Uszkoreit, Llion Jones, Aidan~N Gomez, {\L}ukasz Kaiser, and Illia Polosukhin.
\newblock Attention is all you need.
\newblock \emph{Advances in neural information processing systems}, 30, 2017.

\bibitem[Veytsman(2017)]{CTANacmart}
Boris Veytsman.
\newblock acmart---{C}lass for typesetting publications of {ACM}, 2017.
\newblock URL \url{http://www.ctan.org/pkg/acmart}.

\bibitem[Vrande{\v{c}}i{\'c} and Kr{\"o}tzsch(2014)]{vrandevcic2014wikidata}
Denny Vrande{\v{c}}i{\'c} and Markus Kr{\"o}tzsch.
\newblock Wikidata: a free collaborative knowledgebase.
\newblock \emph{Communications of the ACM}, 57\penalty0 (10):\penalty0 78--85, 2014.

\bibitem[Wang et~al.(2018)Wang, Xu, Han, and Hong]{wang2018movie}
Bo~Wang, Youjiang Xu, Yahong Han, and Richang Hong.
\newblock Movie question answering: Remembering the textual cues for layered visual contents.
\newblock In \emph{Proceedings of the AAAI Conference on Artificial Intelligence}, volume~32, 2018.

\bibitem[Wang et~al.(2017{\natexlab{a}})Wang, Wu, Shen, Dick, and Van Den~Henge]{wang2015explicit}
Peng Wang, Qi~Wu, Chunhua Shen, Anthony Dick, and Anton Van Den~Henge.
\newblock Explicit knowledge-based reasoning for visual question answering.
\newblock In \emph{Proceedings of the 26th International Joint Conference on Artificial Intelligence}, page 1290–1296. AAAI Press, 2017{\natexlab{a}}.
\newblock ISBN 9780999241103.

\bibitem[Wang et~al.(2017{\natexlab{b}})Wang, Wu, Shen, Dick, and Van Den~Hengel]{wang2017fvqa}
Peng Wang, Qi~Wu, Chunhua Shen, Anthony Dick, and Anton Van Den~Hengel.
\newblock Fvqa: Fact-based visual question answering.
\newblock \emph{IEEE transactions on pattern analysis and machine intelligence}, 40\penalty0 (10):\penalty0 2413--2427, 2017{\natexlab{b}}.

\bibitem[Wang et~al.(2017{\natexlab{c}})Wang, Wu, Shen, Dick, and van~den Hengel]{wang2017explicit}
Peng Wang, Qi~Wu, Chunhua Shen, Anthony~R Dick, and Anton van~den Hengel.
\newblock Explicit knowledge-based reasoning for visual question answering.
\newblock In \emph{IJCAI}, 2017{\natexlab{c}}.

\bibitem[Wang et~al.(2024)Wang, Chen, Chen, Wu, Zhu, Zeng, Luo, Lu, Zhou, Qiao, et~al.]{wang2024visionllm}
Wenhai Wang, Zhe Chen, Xiaokang Chen, Jiannan Wu, Xizhou Zhu, Gang Zeng, Ping Luo, Tong Lu, Jie Zhou, Yu~Qiao, et~al.
\newblock Visionllm: Large language model is also an open-ended decoder for vision-centric tasks.
\newblock \emph{Advances in Neural Information Processing Systems}, 36, 2024.

\bibitem[Wang et~al.(2021{\natexlab{a}})Wang, Bao, Dong, and Wei]{wang2021vlmo}
Wenhui Wang, Hangbo Bao, Li~Dong, and Furu Wei.
\newblock Vlmo: Unified vision-language pre-training with mixture-of-modality-experts.
\newblock \emph{arXiv preprint arXiv:2111.02358}, 2021{\natexlab{a}}.

\bibitem[Wang et~al.(2022{\natexlab{a}})Wang, Bao, Dong, Bjorck, Peng, Liu, Aggarwal, Mohammed, Singhal, Som, et~al.]{wang2022image}
Wenhui Wang, Hangbo Bao, Li~Dong, Johan Bjorck, Zhiliang Peng, Qiang Liu, Kriti Aggarwal, Owais~Khan Mohammed, Saksham Singhal, Subhojit Som, et~al.
\newblock Image as a foreign language: Beit pretraining for all vision and vision-language tasks.
\newblock \emph{arXiv preprint arXiv:2208.10442}, 2022{\natexlab{a}}.

\bibitem[Wang et~al.(2020)Wang, Wang, Han, Jiang, Han, Liu, Li, Li, Lin, and Zhou]{wang2020maven}
Xiaozhi Wang, Ziqi Wang, Xu~Han, Wangyi Jiang, Rong Han, Zhiyuan Liu, Juanzi Li, Peng Li, Yankai Lin, and Jie Zhou.
\newblock Maven: A massive general domain event detection dataset.
\newblock \emph{arXiv preprint arXiv:2004.13590}, 2020.

\bibitem[Wang et~al.(2022{\natexlab{b}})Wang, Wu, Furumai, Wada, and Kurihara]{wang2022vae}
Yanan Wang, Jianming Wu, Kazuaki Furumai, Shinya Wada, and Satoshi Kurihara.
\newblock Vae-based adversarial multimodal domain transfer for video-level sentiment analysis.
\newblock \emph{IEEE Access}, 2022{\natexlab{b}}.

\bibitem[Wang et~al.(2021{\natexlab{b}})Wang, Zhang, Guo, Yin, Li, and Chen]{wang2021decoupling}
Yanling Wang, Jing Zhang, Shasha Guo, Hongzhi Yin, Cuiping Li, and Hong Chen.
\newblock Decoupling representation learning and classification for gnn-based anomaly detection.
\newblock In \emph{Proceedings of the 44th international ACM SIGIR conference on research and development in information retrieval}, pages 1239--1248, 2021{\natexlab{b}}.

\bibitem[Wang et~al.(2021{\natexlab{c}})Wang, Yu, Yu, Dai, Tsvetkov, and Cao]{wang2021simvlm}
Zirui Wang, Jiahui Yu, Adams~Wei Yu, Zihang Dai, Yulia Tsvetkov, and Yuan Cao.
\newblock Simvlm: Simple visual language model pretraining with weak supervision.
\newblock In \emph{International Conference on Learning Representations}, 2021{\natexlab{c}}.

\bibitem[Wei et~al.(2021)Wei, Bosma, Zhao, Guu, Yu, Lester, Du, Dai, and Le]{wei2021finetuned}
Jason Wei, Maarten Bosma, Vincent Zhao, Kelvin Guu, Adams~Wei Yu, Brian Lester, Nan Du, Andrew~M Dai, and Quoc~V Le.
\newblock Finetuned language models are zero-shot learners.
\newblock In \emph{International Conference on Learning Representations}, 2021.

\bibitem[Wei et~al.(2022)Wei, Wang, Schuurmans, Bosma, Xia, Chi, Le, Zhou, et~al.]{wei2022chain}
Jason Wei, Xuezhi Wang, Dale Schuurmans, Maarten Bosma, Fei Xia, Ed~Chi, Quoc~V Le, Denny Zhou, et~al.
\newblock Chain-of-thought prompting elicits reasoning in large language models.
\newblock \emph{Advances in Neural Information Processing Systems}, 35:\penalty0 24824--24837, 2022.

\bibitem[Wenzel(1992)]{Wenzel:1992:TVA:146022.146089}
Elizabeth~M. Wenzel.
\newblock Three-dimensional virtual acoustic displays.
\newblock In \emph{Multimedia interface design (incoll)}, pages 257--288. ACM, New York, NY, USA, 1992.
\newblock ISBN 0-201-54981-6.
\newblock \doi{10.1145/146022.146089}.
\newblock URL \url{http://portal.acm.org/citation.cfm?id=146022.146089}.

\bibitem[Werneck et~al.(2000{\natexlab{a}})Werneck, Setubal, and da~Conceic\~{a}o]{384253}
Renato Werneck, Jo\~{a}o Setubal, and Arlindo da~Conceic\~{a}o.
\newblock (old) finding minimum congestion spanning trees.
\newblock \emph{J. Exp. Algorithmics}, 5:\penalty0 11, 2000{\natexlab{a}}.
\newblock ISSN 1084-6654.
\newblock \doi{http://doi.acm.org/10.1145/351827.384253}.

\bibitem[Werneck et~al.(2000{\natexlab{b}})Werneck, Setubal, and da~Conceic\~{a}o]{Werneck:2000:FMC:351827.384253}
Renato Werneck, Jo\~{a}o Setubal, and Arlindo da~Conceic\~{a}o.
\newblock (new) finding minimum congestion spanning trees.
\newblock \emph{J. Exp. Algorithmics}, 5, December 2000{\natexlab{b}}.
\newblock ISSN 1084-6654.
\newblock \doi{10.1145/351827.384253}.
\newblock URL \url{http://portal.acm.org/citation.cfm?id=351827.384253}.

\bibitem[Wu et~al.(2018)Wu, Liu, Wang, and Dong]{wu2018chain}
Chenfei Wu, Jinlai Liu, Xiaojie Wang, and Xuan Dong.
\newblock Chain of reasoning for visual question answering.
\newblock \emph{Advances in Neural Information Processing Systems}, 31, 2018.

\bibitem[Wu et~al.(2019)Wu, Liu, Wang, and Li]{wu2019differential}
Chenfei Wu, Jinlai Liu, Xiaojie Wang, and Ruifan Li.
\newblock Differential networks for visual question answering.
\newblock In \emph{Proceedings of the AAAI Conference on Artificial Intelligence}, volume~33, pages 8997--9004, 2019.

\bibitem[Wu et~al.(2023{\natexlab{a}})Wu, Yin, Qi, Wang, Tang, and Duan]{visualgpt}
Chenfei Wu, Shengming Yin, Weizhen Qi, Xiaodong Wang, Zecheng Tang, and Nan Duan.
\newblock Visual chatgpt: Talking, drawing and editing with visual foundation models.
\newblock \emph{CoRR}, abs/2303.04671, 2023{\natexlab{a}}.

\bibitem[Wu et~al.(2023{\natexlab{b}})Wu, Yin, Qi, Wang, Tang, and Duan]{wu2023visual}
Chenfei Wu, Shengming Yin, Weizhen Qi, Xiaodong Wang, Zecheng Tang, and Nan Duan.
\newblock Visual chatgpt: Talking, drawing and editing with visual foundation models.
\newblock \emph{arXiv preprint arXiv:2303.04671}, 2023{\natexlab{b}}.

\bibitem[Wu et~al.(2022)Wu, Lu, Sabharwal, and Mottaghi]{wu2022multi}
Jialin Wu, Jiasen Lu, Ashish Sabharwal, and Roozbeh Mottaghi.
\newblock Multi-modal answer validation for knowledge-based vqa.
\newblock In \emph{Proceedings of the AAAI Conference on Artificial Intelligence}, volume~36, pages 2712--2721, 2022.

\bibitem[Wu et~al.(2016{\natexlab{a}})Wu, Shen, Liu, Dick, and Van Den~Hengel]{wu2016value}
Qi~Wu, Chunhua Shen, Lingqiao Liu, Anthony Dick, and Anton Van Den~Hengel.
\newblock What value do explicit high level concepts have in vision to language problems?
\newblock In \emph{Proceedings of the IEEE conference on computer vision and pattern recognition}, pages 203--212, 2016{\natexlab{a}}.

\bibitem[Wu et~al.(2016{\natexlab{b}})Wu, Wang, Shen, Dick, and Van Den~Hengel]{wu2016ask}
Qi~Wu, Peng Wang, Chunhua Shen, Anthony Dick, and Anton Van Den~Hengel.
\newblock Ask me anything: Free-form visual question answering based on knowledge from external sources.
\newblock In \emph{Proceedings of the IEEE conference on computer vision and pattern recognition}, pages 4622--4630, 2016{\natexlab{b}}.

\bibitem[Wu et~al.(2017)Wu, Teney, Wang, Shen, Dick, and Van Den~Hengel]{wu2017visual}
Qi~Wu, Damien Teney, Peng Wang, Chunhua Shen, Anthony Dick, and Anton Van Den~Hengel.
\newblock Visual question answering: A survey of methods and datasets.
\newblock \emph{Computer Vision and Image Understanding}, 163:\penalty0 21--40, 2017.

\bibitem[Wu et~al.(2024)Wu, Li, Sun, and Lu]{wu2024symbol}
Xiaoqian Wu, Yong-Lu Li, Jianhua Sun, and Cewu Lu.
\newblock Symbol-llm: leverage language models for symbolic system in visual human activity reasoning.
\newblock \emph{Advances in Neural Information Processing Systems}, 36, 2024.

\bibitem[Wu and Palmer(1994)]{wu1994verb}
Zhibiao Wu and Martha Palmer.
\newblock Verbs semantics and lexical selection.
\newblock In \emph{Proceedings of the 32nd Annual Meeting on Association for Computational Linguistics}, page 133–138, USA, 1994. Association for Computational Linguistics.
\newblock \doi{10.3115/981732.981751}.
\newblock URL \url{https://doi.org/10.3115/981732.981751}.

\bibitem[Xiao et~al.(2024)Xiao, Sun, Liu, and Wang]{xiao2024logicvista}
Yijia Xiao, Edward Sun, Tianyu Liu, and Wei Wang.
\newblock Logicvista: Multimodal llm logical reasoning benchmark in visual contexts.
\newblock \emph{arXiv preprint arXiv:2407.04973}, 2024.

\bibitem[Xiong et~al.(2016)Xiong, Merity, and Socher]{xiong2016dynamic}
Caiming Xiong, Stephen Merity, and Richard Socher.
\newblock Dynamic memory networks for visual and textual question answering.
\newblock In \emph{International conference on machine learning}, pages 2397--2406. PMLR, 2016.

\bibitem[Xu et~al.(2019)Xu, Ma, Zhang, Li, Kang, and Zhou]{xu2019adaptive}
BinChen Xu, Lu~Ma, Liang Zhang, HaoHai Li, Qi~Kang, and MengChu Zhou.
\newblock An adaptive wordpiece language model for learning chinese word embeddings.
\newblock In \emph{2019 IEEE 15th International Conference on Automation Science and Engineering (CASE)}, pages 812--817. IEEE, 2019.

\bibitem[Xu et~al.(2017{\natexlab{a}})Xu, Zhu, Choy, and Fei-Fei]{xu2017scene}
Danfei Xu, Yuke Zhu, Christopher~B Choy, and Li~Fei-Fei.
\newblock Scene graph generation by iterative message passing.
\newblock In \emph{Proceedings of the IEEE conference on computer vision and pattern recognition}, pages 5410--5419, 2017{\natexlab{a}}.

\bibitem[Xu et~al.(2017{\natexlab{b}})Xu, Zhao, Xiao, Wu, Zhang, He, and Zhuang]{xu2017video}
Dejing Xu, Zhou Zhao, Jun Xiao, Fei Wu, Hanwang Zhang, Xiangnan He, and Yueting Zhuang.
\newblock Video question answering via gradually refined attention over appearance and motion.
\newblock In \emph{Proceedings of the 25th ACM international conference on Multimedia}, pages 1645--1653, 2017{\natexlab{b}}.

\bibitem[Xu and Saenko(2016{\natexlab{a}})]{xu2016ask}
Huijuan Xu and Kate Saenko.
\newblock Ask, attend and answer: Exploring question-guided spatial attention for visual question answering.
\newblock In \emph{European conference on computer vision}, pages 451--466. Springer, 2016{\natexlab{a}}.

\bibitem[Xu and Saenko(2016{\natexlab{b}})]{xu2016dual}
Huijuan Xu and Kate Saenko.
\newblock Dual attention network for visual question answering.
\newblock In \emph{ECCV 2016 2nd Workshop on Storytelling with Images and Videos (VisStory)}, 2016{\natexlab{b}}.

\bibitem[Xue et~al.(2021)Xue, Huang, Liu, Peng, Fu, Li, and Luo]{xue2021probing}
Hongwei Xue, Yupan Huang, Bei Liu, Houwen Peng, Jianlong Fu, Houqiang Li, and Jiebo Luo.
\newblock Probing inter-modality: Visual parsing with self-attention for vision-and-language pre-training.
\newblock \emph{Advances in Neural Information Processing Systems}, 34:\penalty0 4514--4528, 2021.

\bibitem[Xue et~al.(2017)Xue, Zhao, and Cai]{xue2017unifying}
Hongyang Xue, Zhou Zhao, and Deng Cai.
\newblock Unifying the video and question attentions for open-ended video question answering.
\newblock \emph{IEEE Transactions on Image Processing}, 26\penalty0 (12):\penalty0 5656--5666, 2017.

\bibitem[Yang et~al.(2021{\natexlab{a}})Yang, Miech, Sivic, Laptev, and Schmid]{yang2021just}
Antoine Yang, Antoine Miech, Josef Sivic, Ivan Laptev, and Cordelia Schmid.
\newblock Just ask: Learning to answer questions from millions of narrated videos.
\newblock In \emph{Proceedings of the IEEE/CVF International Conference on Computer Vision}, pages 1686--1697, 2021{\natexlab{a}}.

\bibitem[Yang et~al.(2022{\natexlab{a}})Yang, Miech, Sivic, Laptev, and Schmid]{yang2022zero}
Antoine Yang, Antoine Miech, Josef Sivic, Ivan Laptev, and Cordelia Schmid.
\newblock Zero-shot video question answering via frozen bidirectional language models.
\newblock \emph{arXiv preprint arXiv:2206.08155}, 2022{\natexlab{a}}.

\bibitem[Yang et~al.(2019{\natexlab{a}})Yang, Jiang, Jiang, Zhou, and Li]{yang2019co}
Chao Yang, Mengqi Jiang, Bin Jiang, Weixin Zhou, and Keqin Li.
\newblock Co-attention network with question type for visual question answering.
\newblock \emph{IEEE Access}, 7:\penalty0 40771--40781, 2019{\natexlab{a}}.

\bibitem[Yang et~al.(2023{\natexlab{a}})Yang, Zhang, Li, Zou, Li, and Gao]{yang2023setofmark}
Jianwei Yang, Hao Zhang, Feng Li, Xueyan Zou, Chunyuan Li, and Jianfeng Gao.
\newblock Set-of-mark prompting unleashes extraordinary visual grounding in gpt-4v.
\newblock \emph{arXiv preprint arXiv:2310.11441}, 2023{\natexlab{a}}.

\bibitem[Yang et~al.(2020{\natexlab{a}})Yang, Xu, and Gao]{yang2020cm}
Kaicheng Yang, Hua Xu, and Kai Gao.
\newblock Cm-bert: Cross-modal bert for text-audio sentiment analysis.
\newblock In \emph{Proceedings of the 28th ACM international conference on multimedia}, pages 521--528, 2020{\natexlab{a}}.

\bibitem[Yang et~al.(2019{\natexlab{b}})Yang, Liu, Chen, Zhao, Chen, and Shen]{yang2019advanced}
Min Yang, Junhao Liu, Lei Chen, Zhou Zhao, Xiaojun Chen, and Ying Shen.
\newblock An advanced deep generative framework for temporal link prediction in dynamic networks.
\newblock \emph{IEEE transactions on cybernetics}, 50\penalty0 (12):\penalty0 4946--4957, 2019{\natexlab{b}}.

\bibitem[Yang et~al.(2020{\natexlab{b}})Yang, Li, Shen, Wu, Zhao, and Chen]{yang2020hierarchical}
Min Yang, Chengming Li, Ying Shen, Qingyao Wu, Zhou Zhao, and Xiaojun Chen.
\newblock Hierarchical human-like deep neural networks for abstractive text summarization.
\newblock \emph{IEEE Transactions on Neural Networks and Learning Systems}, 32\penalty0 (6):\penalty0 2744--2757, 2020{\natexlab{b}}.

\bibitem[Yang et~al.(2024)Yang, Song, Li, Zhao, Ge, Li, and Shan]{yang2024gpt4tools}
Rui Yang, Lin Song, Yanwei Li, Sijie Zhao, Yixiao Ge, Xiu Li, and Ying Shan.
\newblock Gpt4tools: Teaching large language model to use tools via self-instruction.
\newblock \emph{Advances in Neural Information Processing Systems}, 36, 2024.

\bibitem[Yang et~al.(2021{\natexlab{b}})Yang, Gao, Zhang, and Cai]{yang2021auto}
Xu~Yang, Chongyang Gao, Hanwang Zhang, and Jianfei Cai.
\newblock Auto-parsing network for image captioning and visual question answering.
\newblock In \emph{Proceedings of the IEEE/CVF International Conference on Computer Vision}, pages 2197--2207, 2021{\natexlab{b}}.

\bibitem[Yang et~al.(2020{\natexlab{c}})Yang, Garcia, Chu, Otani, Nakashima, and Takemura]{yang2020bert}
Zekun Yang, Noa Garcia, Chenhui Chu, Mayu Otani, Yuta Nakashima, and Haruo Takemura.
\newblock Bert representations for video question answering.
\newblock In \emph{Proceedings of the IEEE/CVF Winter Conference on Applications of Computer Vision}, pages 1556--1565, 2020{\natexlab{c}}.

\bibitem[Yang et~al.(2023{\natexlab{b}})Yang, Xiang, You, Li, and Liu]{yang2023event}
Zhenguo Yang, Jiale Xiang, Jiuxiang You, Qing Li, and Wenyin Liu.
\newblock Event-oriented visual question answering: The e-vqa dataset and benchmark.
\newblock \emph{IEEE Transactions on Knowledge and Data Engineering}, 35\penalty0 (10):\penalty0 10210--10223, 2023{\natexlab{b}}.

\bibitem[Yang et~al.(2022{\natexlab{b}})Yang, Gan, Wang, Hu, Lu, Liu, and Wang]{DBLP:conf/aaai/YangGW0L0W22}
Zhengyuan Yang, Zhe Gan, Jianfeng Wang, Xiaowei Hu, Yumao Lu, Zicheng Liu, and Lijuan Wang.
\newblock An empirical study of {GPT-3} for few-shot knowledge-based {VQA}.
\newblock In \emph{Thirty-Sixth {AAAI} Conference on Artificial Intelligence, {AAAI} 2022, Thirty-Fourth Conference on Innovative Applications of Artificial Intelligence, {IAAI} 2022, The Twelveth Symposium on Educational Advances in Artificial Intelligence, {EAAI} 2022 Virtual Event, February 22 - March 1, 2022}, pages 3081--3089. {AAAI} Press, 2022{\natexlab{b}}.

\bibitem[Yang et~al.(2022{\natexlab{c}})Yang, Gan, Wang, Hu, Lu, Liu, and Wang]{yang2022empirical}
Zhengyuan Yang, Zhe Gan, Jianfeng Wang, Xiaowei Hu, Yumao Lu, Zicheng Liu, and Lijuan Wang.
\newblock An empirical study of gpt-3 for few-shot knowledge-based vqa.
\newblock In \emph{Proceedings of the AAAI Conference on Artificial Intelligence}, volume~36, pages 3081--3089, 2022{\natexlab{c}}.

\bibitem[Yang et~al.(2023{\natexlab{c}})Yang, Li, Wang, Lin, Azarnasab, Ahmed, Liu, Liu, Zeng, and Wang]{mmreact}
Zhengyuan Yang, Linjie Li, Jianfeng Wang, Kevin Lin, Ehsan Azarnasab, Faisal Ahmed, Zicheng Liu, Ce~Liu, Michael Zeng, and Lijuan Wang.
\newblock {MM-REACT:} prompting chatgpt for multimodal reasoning and action.
\newblock \emph{CoRR}, abs/2303.11381, 2023{\natexlab{c}}.

\bibitem[Yang et~al.(2023{\natexlab{d}})Yang, Li, Wang, Lin, Azarnasab, Ahmed, Liu, Liu, Zeng, and Wang]{yang2023mm}
Zhengyuan Yang, Linjie Li, Jianfeng Wang, Kevin Lin, Ehsan Azarnasab, Faisal Ahmed, Zicheng Liu, Ce~Liu, Michael Zeng, and Lijuan Wang.
\newblock Mm-react: Prompting chatgpt for multimodal reasoning and action.
\newblock \emph{arXiv preprint arXiv:2303.11381}, 2023{\natexlab{d}}.

\bibitem[Yang et~al.(2019{\natexlab{c}})Yang, Dai, Yang, Carbonell, Salakhutdinov, and Le]{yang2019xlnet}
Zhilin Yang, Zihang Dai, Yiming Yang, Jaime Carbonell, Russ~R Salakhutdinov, and Quoc~V Le.
\newblock Xlnet: Generalized autoregressive pretraining for language understanding.
\newblock \emph{Advances in neural information processing systems}, 32, 2019{\natexlab{c}}.

\bibitem[Yang et~al.(2018)Yang, Qin, Yu, and Hu]{yang2018scene}
Zhuoqian Yang, Zengchang Qin, Jing Yu, and Yue Hu.
\newblock Scene graph reasoning with prior visual relationship for visual question answering.
\newblock \emph{arXiv preprint arXiv:1812.09681}, 2018.

\bibitem[Yang et~al.(2016)Yang, He, Gao, Deng, and Smola]{yang2016stacked}
Zichao Yang, Xiaodong He, Jianfeng Gao, Li~Deng, and Alex Smola.
\newblock Stacked attention networks for image question answering.
\newblock In \emph{Proceedings of the IEEE conference on computer vision and pattern recognition}, pages 21--29, 2016.

\bibitem[Ye et~al.(2023)Ye, Xu, Ye, Yan, Hu, Liu, Qian, Zhang, Huang, and Zhou]{mplug-owl2}
Qinghao Ye, Haiyang Xu, Jiabo Ye, Ming Yan, Anwen Hu, Haowei Liu, Qi~Qian, Ji~Zhang, Fei Huang, and Jingren Zhou.
\newblock mplug-owl2: Revolutionizing multi-modal large language model with modality collaboration.
\newblock \emph{CoRR}, abs/2311.04257, 2023.

\bibitem[Ye et~al.(2017)Ye, Zhao, Li, Chen, Xiao, and Zhuang]{ye2017video}
Yunan Ye, Zhou Zhao, Yimeng Li, Long Chen, Jun Xiao, and Yueting Zhuang.
\newblock Video question answering via attribute-augmented attention network learning.
\newblock In \emph{Proceedings of the 40th International ACM SIGIR conference on Research and Development in Information Retrieval}, pages 829--832, 2017.

\bibitem[Yi et~al.(2018)Yi, Wu, Gan, Torralba, Kohli, and Tenenbaum]{yi2018neural}
Kexin Yi, Jiajun Wu, Chuang Gan, Antonio Torralba, Pushmeet Kohli, and Josh Tenenbaum.
\newblock Neural-symbolic vqa: Disentangling reasoning from vision and language understanding.
\newblock \emph{Advances in neural information processing systems}, 31, 2018.

\bibitem[You et~al.(2023)You, Sun, Wang, Chen, Wang, Ayyubi, Chang, and Chang]{you2023idealgpt}
Haoxuan You, Rui Sun, Zhecan Wang, Long Chen, Gengyu Wang, Hammad~A Ayyubi, Kai-Wei Chang, and Shih-Fu Chang.
\newblock Idealgpt: Iteratively decomposing vision and language reasoning via large language models.
\newblock \emph{arXiv preprint arXiv:2305.14985}, 2023.

\bibitem[Yu et~al.(2021)Yu, Tang, Yin, Sun, Tian, Wu, and Wang]{yu2021ernie}
Fei Yu, Jiji Tang, Weichong Yin, Yu~Sun, Hao Tian, Hua Wu, and Haifeng Wang.
\newblock Ernie-vil: Knowledge enhanced vision-language representations through scene graphs.
\newblock In \emph{Proceedings of the AAAI Conference on Artificial Intelligence}, volume~35, pages 3208--3216, 2021.

\bibitem[Yu et~al.(2020)Yu, Zhu, Wang, Zhang, Hu, and Tan]{yu2020cross}
Jing Yu, Zihao Zhu, Yujing Wang, Weifeng Zhang, Yue Hu, and Jianlong Tan.
\newblock Cross-modal knowledge reasoning for knowledge-based visual question answering.
\newblock \emph{Pattern Recognition}, 108:\penalty0 107563, 2020.

\bibitem[Yu et~al.(2015)Yu, Park, Berg, and Berg]{yu2015visual}
Licheng Yu, Eunbyung Park, Alexander~C Berg, and Tamara~L Berg.
\newblock Visual madlibs: Fill in the blank image generation and question answering.
\newblock \emph{arXiv preprint arXiv:1506.00278}, 2015.

\bibitem[Yu et~al.(2017{\natexlab{a}})Yu, Ko, Choi, and Kim]{yu2017end}
Youngjae Yu, Hyungjin Ko, Jongwook Choi, and Gunhee Kim.
\newblock End-to-end concept word detection for video captioning, retrieval, and question answering.
\newblock In \emph{Proceedings of the IEEE Conference on Computer Vision and Pattern Recognition}, pages 3165--3173, 2017{\natexlab{a}}.

\bibitem[Yu et~al.(2018{\natexlab{a}})Yu, Kim, and Kim]{yu2018joint}
Youngjae Yu, Jongseok Kim, and Gunhee Kim.
\newblock A joint sequence fusion model for video question answering and retrieval.
\newblock In \emph{Proceedings of the European Conference on Computer Vision (ECCV)}, pages 471--487, 2018{\natexlab{a}}.

\bibitem[Yu et~al.(2017{\natexlab{b}})Yu, Yu, Fan, and Tao]{yu2017multi}
Zhou Yu, Jun Yu, Jianping Fan, and Dacheng Tao.
\newblock Multi-modal factorized bilinear pooling with co-attention learning for visual question answering.
\newblock In \emph{Proceedings of the IEEE international conference on computer vision}, pages 1821--1830, 2017{\natexlab{b}}.

\bibitem[Yu et~al.(2018{\natexlab{b}})Yu, Yu, Xiang, Fan, and Tao]{yu2018beyond}
Zhou Yu, Jun Yu, Chenchao Xiang, Jianping Fan, and Dacheng Tao.
\newblock Beyond bilinear: Generalized multimodal factorized high-order pooling for visual question answering.
\newblock \emph{IEEE transactions on neural networks and learning systems}, 29\penalty0 (12):\penalty0 5947--5959, 2018{\natexlab{b}}.

\bibitem[Yu et~al.(2019)Yu, Yu, Cui, Tao, and Tian]{yu2019deep}
Zhou Yu, Jun Yu, Yuhao Cui, Dacheng Tao, and Qi~Tian.
\newblock Deep modular co-attention networks for visual question answering.
\newblock In \emph{Proceedings of the IEEE/CVF conference on computer vision and pattern recognition}, pages 6281--6290, 2019.

\bibitem[Yuan et~al.(2020)Yuan, Sun, Duan, Li, Wu, and Xu]{yuan2020adversarial}
Zhaoquan Yuan, Siyuan Sun, Lixin Duan, Changsheng Li, Xiao Wu, and Changsheng Xu.
\newblock Adversarial multimodal network for movie story question answering.
\newblock \emph{IEEE Transactions on Multimedia}, 23:\penalty0 1744--1756, 2020.

\bibitem[Yue et~al.(2024)Yue, Ni, Zhang, Zheng, Liu, Zhang, Stevens, Jiang, Ren, Sun, Wei, Yu, Yuan, Sun, Yin, Zheng, Yang, Liu, Huang, Sun, Su, and Chen]{yue2023mmmu}
Xiang Yue, Yuansheng Ni, Kai Zhang, Tianyu Zheng, Ruoqi Liu, Ge~Zhang, Samuel Stevens, Dongfu Jiang, Weiming Ren, Yuxuan Sun, Cong Wei, Botao Yu, Ruibin Yuan, Renliang Sun, Ming Yin, Boyuan Zheng, Zhenzhu Yang, Yibo Liu, Wenhao Huang, Huan Sun, Yu~Su, and Wenhu Chen.
\newblock Mmmu: A massive multi-discipline multimodal understanding and reasoning benchmark for expert agi.
\newblock In \emph{Proceedings of CVPR}, 2024.

\bibitem[Zadeh et~al.(2016)Zadeh, Zellers, Pincus, and Morency]{zadeh2016mosi}
Amir Zadeh, Rowan Zellers, Eli Pincus, and Louis-Philippe Morency.
\newblock Mosi: multimodal corpus of sentiment intensity and subjectivity analysis in online opinion videos.
\newblock \emph{arXiv preprint arXiv:1606.06259}, 2016.

\bibitem[Zadeh et~al.(2020)Zadeh, Liang, and Morency]{zadeh2020foundations}
Amir Zadeh, Paul~Pu Liang, and Louis-Philippe Morency.
\newblock Foundations of multimodal co-learning.
\newblock \emph{Information Fusion}, 64:\penalty0 188--193, 2020.

\bibitem[Zadeh et~al.(2018)Zadeh, Liang, Poria, Cambria, and Morency]{zadeh2018multimodal}
AmirAli~Bagher Zadeh, Paul~Pu Liang, Soujanya Poria, Erik Cambria, and Louis-Philippe Morency.
\newblock Multimodal language analysis in the wild: Cmu-mosei dataset and interpretable dynamic fusion graph.
\newblock In \emph{Proceedings of the 56th Annual Meeting of the Association for Computational Linguistics (Volume 1: Long Papers)}, pages 2236--2246, 2018.

\bibitem[Zeng et~al.(2017)Zeng, Chen, Chuang, Liao, Niebles, and Sun]{zeng2017leveraging}
Kuo-Hao Zeng, Tseng-Hung Chen, Ching-Yao Chuang, Yuan-Hong Liao, Juan~Carlos Niebles, and Min Sun.
\newblock Leveraging video descriptions to learn video question answering.
\newblock In \emph{Thirty-First AAAI Conference on Artificial Intelligence}, 2017.

\bibitem[Zeng et~al.(2024)Zeng, Lin, Ye, and Zeng]{zeng2024advancing}
Xingchen Zeng, Haichuan Lin, Yilin Ye, and Wei Zeng.
\newblock Advancing multimodal large language models in chart question answering with visualization-referenced instruction tuning.
\newblock \emph{IEEE Transactions on Visualization and Computer Graphics}, 2024.

\bibitem[Zhai et~al.(2023)Zhai, Mustafa, Kolesnikov, and Beyer]{DBLP:conf/iccv/ZhaiM0B23}
Xiaohua Zhai, Basil Mustafa, Alexander Kolesnikov, and Lucas Beyer.
\newblock Sigmoid loss for language image pre-training.
\newblock In \emph{{IEEE/CVF} International Conference on Computer Vision, {ICCV} 2023, Paris, France, October 1-6, 2023}, pages 11941--11952. {IEEE}, 2023.

\bibitem[Zhang et~al.(2019)Zhang, Cao, and Wu]{zhang2019information}
Dongxiang Zhang, Rui Cao, and Sai Wu.
\newblock Information fusion in visual question answering: A survey.
\newblock \emph{Information Fusion}, 52:\penalty0 268--280, 2019.

\bibitem[Zhang et~al.(2024{\natexlab{a}})Zhang, Du, Chen, Liang, Luo, Zheng, Zhu, Cheng, Xu, Guo, et~al.]{zhang2024cmmmu}
Ge~Zhang, Xinrun Du, Bei Chen, Yiming Liang, Tongxu Luo, Tianyu Zheng, Kang Zhu, Yuyang Cheng, Chunpu Xu, Shuyue Guo, et~al.
\newblock Cmmmu: A chinese massive multi-discipline multimodal understanding benchmark.
\newblock \emph{arXiv preprint arXiv:2401.11944}, 2024{\natexlab{a}}.

\bibitem[Zhang et~al.(2022)Zhang, Zhang, Zhang, Wang, Li, jiang, wei, and Yang]{zhang2022Tempo}
Huibin Zhang, Zhengkun Zhang, Yao Zhang, Jun Wang, Yufan Li, Ning jiang, Xin wei, and Zhenglu Yang.
\newblock Modeling temporal-modal entity graph for procedural multimodal machine comprehension.
\newblock Proceedings of the 60th Annual Meeting of the Association for Computational Linguistics (Volume 1: Long Papers):\penalty0 1179--1189, 2022.

\bibitem[Zhang et~al.(2024{\natexlab{b}})Zhang, Pang, Du, Ren, Li, and Lin]{zhang2024benchmarking}
Jiawei Zhang, Tianyu Pang, Chao Du, Yi~Ren, Bo~Li, and Min Lin.
\newblock Benchmarking large multimodal models against common corruptions.
\newblock In \emph{NAACL}, 2024{\natexlab{b}}.

\bibitem[Zhang et~al.(2018)Zhang, Wang, and Liu]{zhang2018deep}
Lei Zhang, Shuai Wang, and Bing Liu.
\newblock Deep learning for sentiment analysis: A survey.
\newblock \emph{Wiley Interdisciplinary Reviews: Data Mining and Knowledge Discovery}, 8\penalty0 (4):\penalty0 e1253, 2018.

\bibitem[Zhang et~al.(2021)Zhang, Li, Hu, Yang, Zhang, Wang, Choi, and Gao]{zhang2021vinvl}
Pengchuan Zhang, Xiujun Li, Xiaowei Hu, Jianwei Yang, Lei Zhang, Lijuan Wang, Yejin Choi, and Jianfeng Gao.
\newblock Vinvl: Revisiting visual representations in vision-language models.
\newblock In \emph{Proceedings of the IEEE/CVF Conference on Computer Vision and Pattern Recognition}, pages 5579--5588, 2021.

\bibitem[Zhang et~al.(2020)Zhang, Lei, Zhang, and Li]{zhang2020context}
Qi~Zhang, Zhen Lei, Zhaoxiang Zhang, and Stan~Z Li.
\newblock Context-aware attention network for image-text retrieval.
\newblock In \emph{Proceedings of the IEEE/CVF conference on computer vision and pattern recognition}, pages 3536--3545, 2020.

\bibitem[Zhang et~al.(2024{\natexlab{c}})Zhang, Jiang, Zhang, Lin, Guo, Qiu, Zhou, Lu, Chang, Gao, et~al.]{zhang2024mathverse}
Renrui Zhang, Dongzhi Jiang, Yichi Zhang, Haokun Lin, Ziyu Guo, Pengshuo Qiu, Aojun Zhou, Pan Lu, Kai-Wei Chang, Peng Gao, et~al.
\newblock Mathverse: Does your multi-modal llm truly see the diagrams in visual math problems?
\newblock \emph{arXiv preprint arXiv:2403.14624}, 2024{\natexlab{c}}.

\bibitem[Zhang et~al.(2023)Zhang, Zhang, Li, Zhao, Karypis, and Smola]{DBLP:journals/corr/abs-2302-00923}
Zhuosheng Zhang, Aston Zhang, Mu~Li, Hai Zhao, George Karypis, and Alex Smola.
\newblock Multimodal chain-of-thought reasoning in language models.
\newblock \emph{CoRR}, abs/2302.00923, 2023.

\bibitem[Zhao et~al.(2023)Zhao, Cai, Si, Ma, An, Chen, Liu, Wang, Han, and Chang]{mmicl}
Haozhe Zhao, Zefan Cai, Shuzheng Si, Xiaojian Ma, Kaikai An, Liang Chen, Zixuan Liu, Sheng Wang, Wenjuan Han, and Baobao Chang.
\newblock {MMICL:} empowering vision-language model with multi-modal in-context learning.
\newblock \emph{CoRR}, abs/2309.07915, 2023.

\bibitem[Zhao et~al.(2020{\natexlab{a}})Zhao, Jia, Li, Jiang, and Song]{zhao2020multi}
Xiaojuan Zhao, Yan Jia, Aiping Li, Rong Jiang, and Yichen Song.
\newblock Multi-source knowledge fusion: a survey.
\newblock \emph{World Wide Web}, 23\penalty0 (4):\penalty0 2567--2592, 2020{\natexlab{a}}.

\bibitem[Zhao et~al.(2017)Zhao, Yang, Cai, He, Zhuang, Zhao, Yang, Cai, He, and Zhuang]{zhao2017video}
Zhou Zhao, Qifan Yang, Deng Cai, Xiaofei He, Yueting Zhuang, Zhou Zhao, Qifan Yang, Deng Cai, Xiaofei He, and Yueting Zhuang.
\newblock Video question answering via hierarchical spatio-temporal attention networks.
\newblock In \emph{IJCAI}, volume~2, page~8, 2017.

\bibitem[Zhao et~al.(2018)Zhao, Jiang, Cai, Xiao, He, and Pu]{zhao2018multi}
Zhou Zhao, Xinghua Jiang, Deng Cai, Jun Xiao, Xiaofei He, and Shiliang Pu.
\newblock Multi-turn video question answering via multi-stream hierarchical attention context network.
\newblock In \emph{IJCAI}, volume 2018, page 27th, 2018.

\bibitem[Zhao et~al.(2020{\natexlab{b}})Zhao, Xiao, Song, Lu, Xiao, and Zhuang]{zhao2020open}
Zhou Zhao, Shuwen Xiao, Zehan Song, Chujie Lu, Jun Xiao, and Yueting Zhuang.
\newblock Open-ended video question answering via multi-modal conditional adversarial networks.
\newblock \emph{IEEE Transactions on Image Processing}, 29:\penalty0 3859--3870, 2020{\natexlab{b}}.

\bibitem[Zheng et~al.(2020)Zheng, Guo, and Kordjamshidi]{zheng2020cross}
Chen Zheng, Quan Guo, and Parisa Kordjamshidi.
\newblock Cross-modality relevance for reasoning on language and vision.
\newblock \emph{arXiv preprint arXiv:2005.06035}, 2020.

\bibitem[Zheng et~al.(2019)Zheng, Wang, Qi, and Zhu]{zheng2019reasoning}
Zilong Zheng, Wenguan Wang, Siyuan Qi, and Song-Chun Zhu.
\newblock Reasoning visual dialogs with structural and partial observations.
\newblock In \emph{Proceedings of the IEEE/CVF Conference on Computer Vision and Pattern Recognition}, pages 6669--6678, 2019.

\bibitem[Zhong et~al.(2022)Zhong, Liu, Xu, Zhu, and Zeng]{zhong2022dialoglm}
Ming Zhong, Yang Liu, Yichong Xu, Chenguang Zhu, and Michael Zeng.
\newblock Dialoglm: Pre-trained model for long dialogue understanding and summarization.
\newblock In \emph{Proceedings of the AAAI Conference on Artificial Intelligence}, volume~36, pages 11765--11773, 2022.

\bibitem[Zhou et~al.(2015)Zhou, Tian, Sukhbaatar, Szlam, and Fergus]{zhou2015simple}
Bolei Zhou, Yuandong Tian, Sainbayar Sukhbaatar, Arthur Szlam, and Rob Fergus.
\newblock Simple baseline for visual question answering.
\newblock \emph{arXiv preprint arXiv:1512.02167}, 2015.

\bibitem[Zhou et~al.(2008)Zhou, Lu, Wan, Yarvis, and Stankovic]{Zhou-06}
G.~Zhou, J.~Lu, C.-Y. Wan, M.~D. Yarvis, and J.~A. Stankovic.
\newblock \emph{Body Sensor Networks}.
\newblock MIT Press, Cambridge, MA, 2008.

\bibitem[Zhou et~al.(2010)Zhou, Wu, Yan, He, Huang, Stankovic, and Abdelzaher]{Zhou:2010:MMS:1721695.1721705}
Gang Zhou, Yafeng Wu, Ting Yan, Tian He, Chengdu Huang, John~A. Stankovic, and Tarek~F. Abdelzaher.
\newblock A multifrequency mac specially designed for wireless sensor network applications.
\newblock \emph{ACM Trans. Embed. Comput. Syst.}, 9\penalty0 (4):\penalty0 39:1--39:41, April 2010.
\newblock ISSN 1539-9087.
\newblock \doi{10.1145/1721695.1721705}.
\newblock URL \url{http://doi.acm.org/10.1145/1721695.1721705}.

\bibitem[Zhou et~al.(2020)Zhou, Palangi, Zhang, Hu, Corso, and Gao]{zhou2020unified}
Luowei Zhou, Hamid Palangi, Lei Zhang, Houdong Hu, Jason Corso, and Jianfeng Gao.
\newblock Unified vision-language pre-training for image captioning and vqa.
\newblock In \emph{Proceedings of the AAAI Conference on Artificial Intelligence}, volume~34, pages 13041--13049, 2020.

\bibitem[Zhou and Mu(2021)]{zhou2021question}
Xinzhe Zhou and Yadong Mu.
\newblock Question-guided semantic dual-graph visual reasoning with novel answers.
\newblock In \emph{Proceedings of the 2021 International Conference on Multimedia Retrieval}, pages 411--419, 2021.

\bibitem[Zhou et~al.(2021)Zhou, Ren, Zhu, Sun, Liu, Ding, Xu, and Ji]{zhou2021trar}
Yiyi Zhou, Tianhe Ren, Chaoyang Zhu, Xiaoshuai Sun, Jianzhuang Liu, Xinghao Ding, Mingliang Xu, and Rongrong Ji.
\newblock Trar: Routing the attention spans in transformer for visual question answering.
\newblock In \emph{Proceedings of the IEEE/CVF International Conference on Computer Vision}, pages 2074--2084, 2021.

\bibitem[Zhu et~al.(2017{\natexlab{a}})Zhu, Zhao, Huang, Tu, and Ma]{zhu2017structured}
Chen Zhu, Yanpeng Zhao, Shuaiyi Huang, Kewei Tu, and Yi~Ma.
\newblock Structured attentions for visual question answering.
\newblock In \emph{Proceedings of the IEEE International Conference on Computer Vision}, pages 1291--1300, 2017{\natexlab{a}}.

\bibitem[Zhu et~al.(2023)Zhu, Chen, Shen, Li, and Elhoseiny]{zhu2023minigpt}
Deyao Zhu, Jun Chen, Xiaoqian Shen, Xiang Li, and Mohamed Elhoseiny.
\newblock Minigpt-4: Enhancing vision-language understanding with advanced large language models.
\newblock \emph{arXiv preprint arXiv:2304.10592}, 2023.

\bibitem[Zhu et~al.(2016)Zhu, Groth, Bernstein, and Fei-Fei]{zhu2016visual7w}
Yuke Zhu, Oliver Groth, Michael Bernstein, and Li~Fei-Fei.
\newblock Visual7w: Grounded question answering in images.
\newblock In \emph{Proceedings of the IEEE conference on computer vision and pattern recognition}, pages 4995--5004, 2016.

\bibitem[Zhu et~al.(2017{\natexlab{b}})Zhu, Lim, and Fei-Fei]{zhu2017knowledge}
Yuke Zhu, Joseph~J Lim, and Li~Fei-Fei.
\newblock Knowledge acquisition for visual question answering via iterative querying.
\newblock In \emph{Proceedings of the IEEE Conference on Computer Vision and Pattern Recognition}, pages 1154--1163, 2017{\natexlab{b}}.

\bibitem[Zhu et~al.(2021)Zhu, Yu, Wang, Sun, Hu, and Wu]{zhu2020mucko}
Zihao Zhu, Jing Yu, Yujing Wang, Yajing Sun, Yue Hu, and Qi~Wu.
\newblock Mucko: Multi-layer cross-modal knowledge reasoning for fact-based visual question answering.
\newblock In \emph{Proceedings of the Twenty-Ninth International Joint Conference on Artificial Intelligence}, IJCAI'20, 2021.
\newblock ISBN 9780999241165.

\bibitem[Zhuang et~al.(2022)Zhuang, Zhou, Li, Dempster, Budhwani, Al-Sharman, Rayside, and Melek]{zhuang2022radacs}
Alex Zhuang, Eddy Zhou, Quanquan Li, Rowan Dempster, Alikasim Budhwani, Mohammad Al-Sharman, Derek Rayside, and William Melek.
\newblock Radacs: Towards higher-order reasoning using action recognition in autonomous vehicles.
\newblock \emph{arXiv preprint arXiv:2209.14408}, 2022.

\bibitem[Zhuang et~al.(2020)Zhuang, Xu, Yan, Cheng, Zhao, Pu, and Xiao]{zhuang2020multichannel}
Yueting Zhuang, Dejing Xu, Xin Yan, Wenzhuo Cheng, Zhou Zhao, Shiliang Pu, and Jun Xiao.
\newblock Multichannel attention refinement for video question answering.
\newblock \emph{ACM Transactions on Multimedia Computing, Communications, and Applications (TOMM)}, 16\penalty0 (1s):\penalty0 1--23, 2020.

\bibitem[Ziaeefard and Lecue(2020)]{ziaeefard2020towards}
Maryam Ziaeefard and Freddy Lecue.
\newblock Towards knowledge-augmented visual question answering.
\newblock In \emph{Proceedings of the 28th International Conference on Computational Linguistics}, pages 1863--1873, 2020.

\bibitem[Zou and Xie(2020)]{zou2020survey}
Yeyun Zou and Qiyu Xie.
\newblock A survey on vqa: Datasets and approaches.
\newblock In \emph{2020 2nd International Conference on Information Technology and Computer Application (ITCA)}, pages 289--297. IEEE, 2020.

\end{thebibliography}

\end{document}